\documentclass[journal]{IEEEtran}

\ifCLASSOPTIONcompsoc
  \usepackage[nocompress]{cite}
\else
  \usepackage{cite}
\fi

\usepackage{times}
\usepackage{epsfig}
\usepackage{graphicx}
\usepackage{amsmath}
\usepackage{amssymb}
\usepackage{pdfpages}
\usepackage{caption}
\usepackage{subcaption}
\usepackage{xcolor}
\usepackage{hyperref}
\usepackage{amsfonts}
\usepackage{caption}
\usepackage{tabularx}
\usepackage{multirow}
\usepackage{gensymb}
\usepackage{bm}
\usepackage{color}

\DeclareMathOperator*{\argmin}{arg\,min}

\ifCLASSINFOpdf
\else
\fi

\begin{document}

\title{Deep Learning for Visual Localization and Mapping: A Survey}

\author{Changhao Chen*,
        Bing Wang*,
        Chris Xiaoxuan Lu,
        Niki Trigoni,
        Andrew Markham
\IEEEcompsocitemizethanks{
\IEEEcompsocthanksitem Changhao Chen is with the College of Intelligence Science and Technology, National University of Defense Technology, Changsha, 410073, China
\IEEEcompsocthanksitem Bing Wang is with the Department of Aeronautical and Aviation Engineering, The Hong Kong Polytechnic University, HKSAR, China
\IEEEcompsocthanksitem Chris Xiaoxuan Lu is with the School of Informatics, University of Edinburgh, Edinburgh, EH8 9AB, United Kingdom
\IEEEcompsocthanksitem Niki Trigoni and Andrew Markham are with the Department of Computer Science, University of Oxford, Oxford OX1 3QD, United Kingdom
\IEEEcompsocthanksitem *Changhao Chen and Bing Wang are co-first-authors.
\IEEEcompsocthanksitem Corresponding author: Changhao Chen (changhao.chen66@outlook.com)
}% <-this % stops an unwanted space
\thanks{This work was supported by National Natural Science Foundation of China (NFSC) under the Grant Number of 62103427 and 42301520, and EPSRC Program ``ACE-OPS: From Autonomy to Cognitive assistance in Emergency OPerationS" (Grant number: EP/S030832/1) . Changhao Chen is sponsored by the Young Elite Scientist Sponsorship Program by CAST (No. YESS20220181)}% <-this % stops a space
}

\markboth{IEEE Transactions on Neural Networks and Learning Systems, Preprint, Aug~2023}%
{Shell \MakeLowercase{\textit{et al.}}: Bare Demo of IEEEtran.cls for Computer Society Journals}

\maketitle

\begin{abstract}
Deep learning based localization and mapping approaches have recently emerged as a new research direction and receive significant attentions from both industry and academia. Instead of creating hand-designed algorithms based on physical models or geometric theories, deep learning solutions provide an alternative to solve the problem in a data-driven way. Benefiting from the ever-increasing volumes of data and computational power on devices, these learning methods are fast evolving into a new area that shows potentials to track self-motion and estimate environmental model accurately and robustly for mobile agents. In this work, we provide a comprehensive survey, and propose a taxonomy for the localization and mapping methods using deep learning. This survey aims to discuss two basic questions: whether deep learning is promising to localization and mapping; how deep learning should be applied to solve this problem. To this end, a series of localization and mapping topics are investigated, from the learning based visual odometry, global relocalization, to mapping, and simultaneous localization and mapping (SLAM). It is our hope that this survey organically weaves together the recent works in this vein from robotics, computer vision and machine learning communities, and serves as a guideline for future researchers to apply deep learning to tackle the problem of visual localization and mapping.
\end{abstract}

\begin{IEEEkeywords}
Deep Learning, Visual SLAM, Visual Odometry, Visual-inertial Odometry, Global Localization
\end{IEEEkeywords}

\maketitle

\section{Introduction}
Localization and mapping serve as essential requirements for both human beings and mobile agents. As a motivating example, humans possess the remarkable ability to perceive their own motion and the surrounding environment through multisensory perception. They heavily rely on this awareness to determine their location and navigate through intricate three-dimensional spaces.
In a similar vein, mobile agents, encompassing a diverse range of robots like self-driving vehicles, delivery drones, and home service robots, must possess the capability to perceive their environment and estimate positional states through onboard sensors. These agents actively engage in sensing their surroundings and autonomously make decisions \cite{Sunderhauf2018}. Equivalently, the integration of emerging technologies like Augmented Reality (AR) and Virtual Reality (VR) intertwines the virtual and physical realms, making it imperative for machines to possess perceptual awareness. This awareness forms the foundation for seamless interaction between humans and machines. Furthermore, the applications of these concepts extend to mobile and wearable devices such as smartphones, wristbands, and Internet-of-Things (IoT) devices. These devices offer a wide array of location-based services, ranging from pedestrian navigation and sports/activity monitoring to emergency response.

    \begin{figure}
   	\centering
        \includegraphics[width=0.5\textwidth]{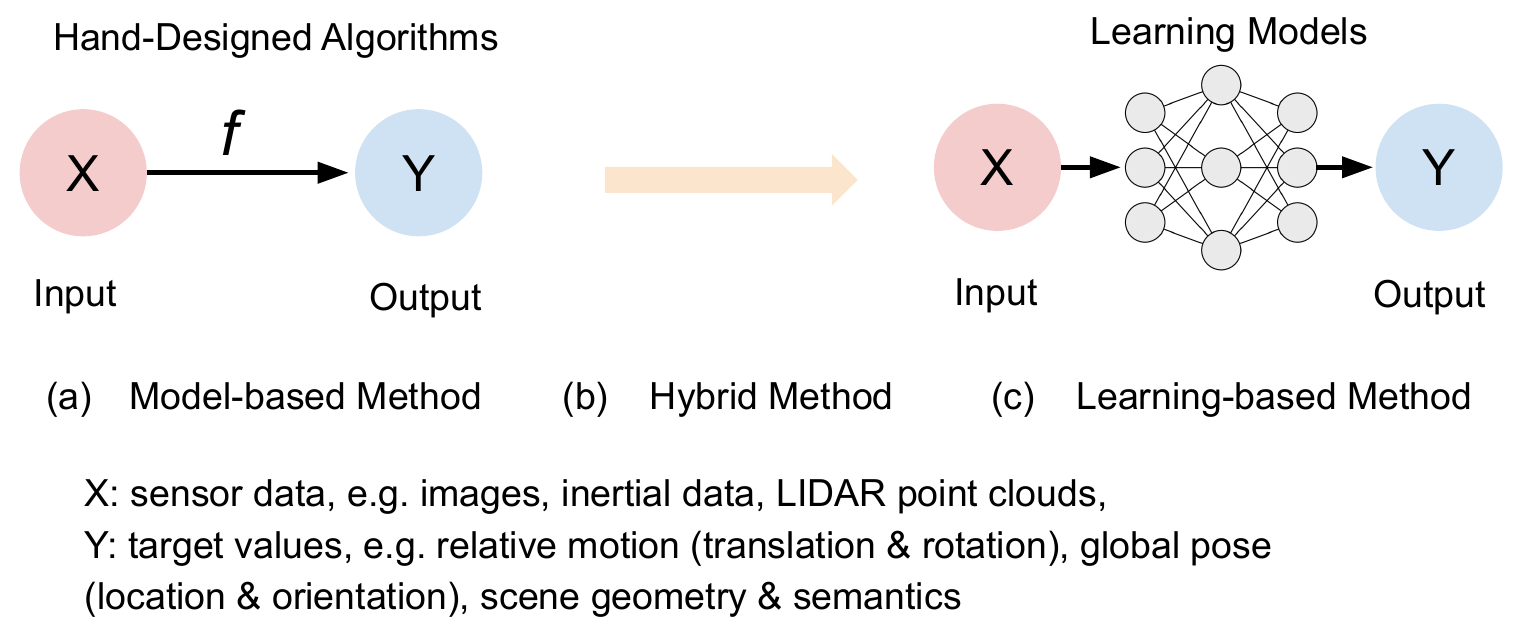}
        \caption{{A localization and mapping system exploits on-board sensors to perceive self-motion, global pose, scene geometry, and semantics. (a) Traditional model-based solutions employ hand-designed algorithms to transform input sensor data into desired output values. (c) Data-driven solutions, on the other hand, leverage learning models to construct this mapping function. (b) Hybrid approaches integrate both hand-designed algorithms and learning models for improved performance.}
        }
        \label{fig: overview}
    \end{figure} 

Enabling a high level of autonomy for these and other digital agents requires precise and robust localization, while incrementally building and maintaining a world model, with the capability to continuously process new information and adapt to various scenarios. In this work, \textit{localization} broadly refers to the ability to obtain internal system states of robot motion, including locations, orientations and velocities, whilst \textit{mapping} indicates the capacity to perceive external environmental states, including scene geometry, appearance and semantics. They can act individually to sense internal or external states respectively, or can operate jointly as a simultaneous localization and mapping (SLAM) system. 

The problem of localization and mapping has been studied for decades, {with a range of algorithms and systems being developed}, for example, visual odometry \cite{Forster2014}, visual-inertial odometry \cite{Qin2018}, image-based relocalization\cite{sattler2011fast}, place recognition\cite{lowry2015visual}, SLAM \cite{Montiel2015}.
These algorithms and systems have demonstrated their efficacy in supporting a wide range of real-world applications, such as delivery robots, self-driving vehicles, and VR devices.
However, the deployment of these systems is not without challenges. Factors such as imperfect sensor measurements, dynamic scenes, adverse lighting conditions, and real-world constraints somewhat hinder their practical implementation. In light of these limitations, recent advancements in machine learning, particularly deep learning, have prompted researchers to explore data-driven approaches as an alternative solution.
Unlike conventional model-based approaches that rely on concrete and explicit algorithms tailored to specific application domains, learning-based methods leverage the power of deep neural networks to extract features and construct implicit neural models, as shown in Figure \ref{fig: overview}.
By training these networks on large datasets, they learn to obtain ability to generate poses and describe scenes, even in challenging environments such as those characterized by high dynamics and poor lighting conditions. Consequently, deep learning-based localization and mapping methods exhibit good robustness and accuracy compared to their traditional counterparts.
Deep learning-based localization and mapping remain active areas of research, and further investigations are necessary to fully understand the strengths and limitations of different approaches.

In this article, we extensively review the existing deep learning based visual localization and mapping approaches, and try to explore the answers to the following two questions:

\begin{itemize}
    \item \textit{1) Is deep learning promising to visual localization and mapping?}
    \item \textit{2) How can deep learning be applied to solve the problem of visual localization and mapping?}
\end{itemize}

\begin{figure*}
    \centering
\includegraphics[width=0.95\textwidth]{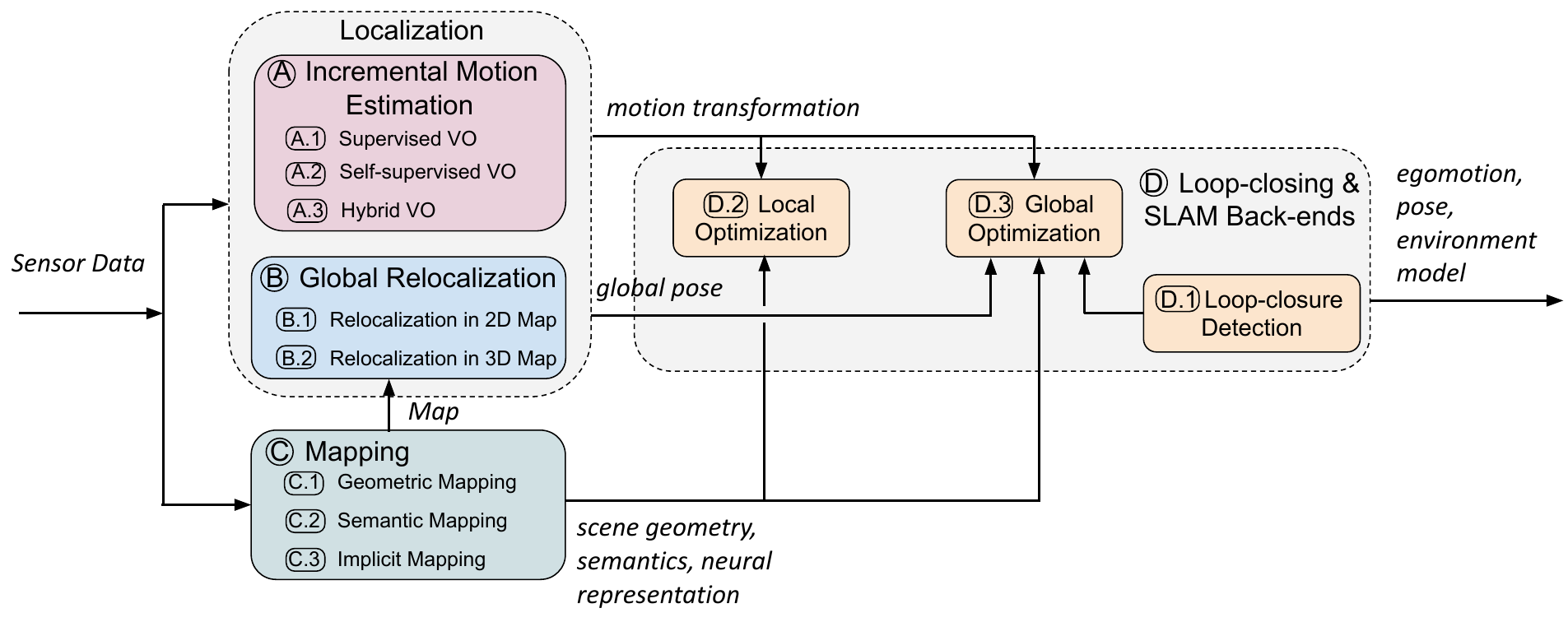}
    \caption{The taxonomy of deep learning  {based visual} localization and mapping. Individual modules can be integrated together into a complete deep learning based SLAM system. It is not mandatory to include all modules for the system to function effectively.  {In the diagram}, rounded rectangles represent function modules, and arrow lines depict the connections between these modules for data input and output.}
    \label{fig: concept figure}
\end{figure*}

The two questions will be revisited by the end of this survey.  {As vision is major information source for most mobile agents, this work will focus on vision based solutions.
The field of deep learning based localization and mapping is still relatively new, and there are a growing number of different approaches and techniques that have been proposed in recent years.}
Notably, although the problem of localization and mapping falls into the key notion of robotics, the incorporation of learning methods progresses in tandem with other research areas such as machine learning, computer vision and even natural language processing. This cross-disciplinary area thus imposes non-trivial difficulty when comprehensively summarizing related works into a survey paper.  {We hope that our survey
can help to promote collaboration and knowledge sharing within the research community, and foster new ideas and facilitate interdisciplinary research on deep learning based localization and mapping. In addition, this survey can help to identify key research challenges and open problems in the field, guide future research efforts, and provide guidance for researchers and practitioners who are interested in using deep learning solutions in their works.}
To the best of our knowledge, this is the first survey article that \textit{thoroughly and extensively covers existing work on deep learning for visual localization and mapping}. 

\begin{table}
 		\caption{ {A summary of relevant surveys and tutorials}} 
  		\label{tab: survey}
 		\small
  		\centering
  		\begin{tabular}{l l l}
  		\hline
    	 {Year} &  {Content} &  {Reference}  \\ \hline
             {2005} &  {probalistic SLAM}  &  {Thrun et al. \cite{thrun2005probabilistic}} \\
    	 {2006} &  {SLAM tutorial}  &  {Durrant-whyte et al. \cite{durrant2006simultaneous}} \\
             {2010} &  {pose-graph SLAM} &  {Grisetti et al. \cite{grisetti2010tutorial}}\\
             {2011} &  {visual odometry tutorial}  &  {Scaramuzza et al. \cite{scaramuzza2011visual}} \\
             {2015} &  {visual place recognition} &  {Lowry et al. \cite{lowry2015visual}} \\
             {2016} &  {SLAM in robust-perception age}   &  {Cadena et al. \cite{cadena2016past}} \\
             {2018} &  {dynamic SLAM} &  {Saputra et al.\cite{saputra2018visual}} \\
             {2018} &  {deep learning for robotics} &  {Sunderhauf et al.\cite{Sunderhauf2018}} \\
              {2022} &  {perception and navigation} &  {Tang et al. \cite{tang2022perception}}\\
             {2023} &  {deep learning based SLAM}  &  {This survey}\\
       \hline
  		\end{tabular}
	\end{table}

\subsection{Comparison with Other Surveys}
As an established field, the development of SLAM problem has been well summarized by several survey papers in literature \cite{durrant2006simultaneous,bailey2006simultaneous}, with their focus lying in the conventional model-based localization and mapping approaches. The seminal survey \cite{cadena2016past} provides a thorough discussion on existing SLAM works, reviews the history of development and charts several future directions. Although this paper contains a section which briefly discusses deep learning models, it does not overview this field comprehensively, especially due to the explosion of research in this area of the past five years. 
Other SLAM survey papers only focus on individual flavours of SLAM systems, including the probabilistic formulation of SLAM \cite{thrun2005probabilistic}, visual odometry \cite{scaramuzza2011visual}, pose-graph SLAM \cite{grisetti2010tutorial}, and SLAM in dynamic environments \cite{saputra2018visual}. 
We refer readers to these surveys for a better understanding of the conventional solutions to SLAM systems.
On the other hand, \cite{Sunderhauf2018} has a discussion on the applications of deep learning to robotics research; however, its main focus is not on localization and mapping specifically, but a more general perspective towards the potentials and limits of deep learning in a broad context of robotic policy learning, reasoning and planning.  {A recent survey \cite{tang2022perception} discusses deep learning based perception and navigation. Compared to \cite{tang2022perception} that throws a broader view on environment perception, motion estimation and reinforcement learning based control for autonomous systems, we provide a more comprehensive review and deep analysis on odometry estimation, relocalization, mapping and other aspects of visual SLAM.}

\subsection{Survey Organization}
The remainder of the paper is organized as follows: Section 2 offers an overview and presents a taxonomy of existing deep learning based localization and mapping; {Sections 3, 4, 5, 6 discuss the existing deep learning approaches on incremental motion (odometry) estimation, global relocalization, mapping, and SLAM back-ends respectively; Sections 7 and 8 review the learning based uncertainty estimation and sensor fusion methods; and finally Section 9 concludes the article, and discusses the limitations and future prospects}.

\section{Taxonomy of Existing Approaches}
From the perspective of learning approaches, we provide a taxonomy of existing deep learning based {visual} localization and mapping, to connect the fields of robotics, computer vision and machine learning.
{Based on their main technical contributions towards a complete SLAM system, related approaches can be broadly categorized into four main types in our context: \textit{incremental motion estimation (visual odometry)}, \textit{global relocalization}, \textit{mapping}, and \textit{loop-closing and SLAM Back-ends}}, as illustrated by the taxonomy shown in Figure \ref{fig: concept figure}:

\textit{A) Incremental Motion Estimation} concerns the calculation of the incremental change in pose, in terms of translation and rotation, between two or more frames of sensor data. It continuously tracks self-motion, and is followed by a process to integrate these pose changes with respect to an initial state to derive global pose. 
Incremental motion estimation,  {i.e. visual odometry (VO)}, can be used in providing pose information in a scenario without a pre-built map or as odometry motion model to assist the feedback loop of robot control.
Deep learning is applied to estimate motion transformations from various sensor measurements in an end-to-end fashion or extract useful features to support a hybrid system. 

\textit{B) Global Relocalization} retrieves the global pose of mobile agents in a known scene with prior knowledge. This is achieved by matching the inquiry input data with a pre-built map or other spatial references.
It can be leveraged to reduce the pose drift of a dead reckoning system or retrieve the absolute pose when motion tracking is lost \cite{thrun2005probabilistic}.  
Deep learning is used to tackle the tricky data association problem that is complicated by the changes in views, illumination, weather and scene dynamics, between the inquiry data and map.

\textit{C) Mapping} builds and reconstructs a consistent environmental model to describe the surroundings.  
Mapping can be used to provide environment information for human operators or high-level robot tasks, constrain the error drifts of self-motion tracking, and retrieve the inquiry observation for global localization \cite{cadena2016past}. 
Deep learning is leveraged as a useful tool to discover scene geometry and semantics from high-dimensional raw data for mapping. Deep learning based mapping methods are sub-divided into geometric, semantic, and implicit mapping, depending on whether the neural network learns the explicit geometry, or semantics of a scene, or encodes the scene into implicit neural representation.

\textit{D) Loop-closing and SLAM Back-ends} detect loop closures and optimize the aforementioned incremental motion estimation, global localization and mapping modules to boost the performance in a simultaneous localization and mapping (SLAM) system.
These modules perform to ensure the consistency of entire system as follows: \emph{local optimization} ensures the local consistency of camera motion and scene geometry; once a loop closure is detected by \emph{loop-closing module}, system error drifts can be mitigated by \emph{global optimization}. 

Besides the modules mentioned above, other modules that also contribute to a SLAM system include:

\textit{E) Uncertainty Estimation} provides a metric of belief in the learned poses and mapping, critical to probabilistic sensor fusion and back-end optimization in a SLAM system.

\textit{F) Sensor Fusion} exploits the complementary properties of each sensor modality, and aims to discover the suitable data fusion strategy such that more accurate and robust localization and mapping can be achieved.

In the following sections, we will discuss these components in details.

\section{{Incremental Motion Estimation}}
\label{sec: incremental motion}
We begin with incremental motion (odometry) estimation,  {i.e. visual odometry (VO)}, which continuously tracks camera egomotion and yields motion transformations.
Given an initial state, global trajectories are reconstructed by integrating these incremental poses. Thus, it is critical to keep the estimate of each motion transformation accurate enough to ensure high-prevision localization in a global scale. 
This section presents deep learning approaches to achieve  {visual odometry}.

Deep learning is capable of extracting high-level feature representations from raw images directly, and thereby provides an alternative to solve visual odometry (VO) problem, without requiring hand-crafted feature detectors. 
Existing deep learning based VO models can be categorized into \emph{end-to-end VO} and \emph{hybrid VO}, depending on whether they are purely DNN based or a combination of classical VO algorithms and DNNs. 
Depending on the availability of ground-truth labels in the training phase, end-to-end VO systems are further classified into \emph{supervised} VO and \emph{unsupervised} VO. Table \ref{tb: odometry estimation} lists and compares deep learning based visual odometry methods.

\begin{figure*}
    \centering
    \includegraphics[width=0.8\textwidth]{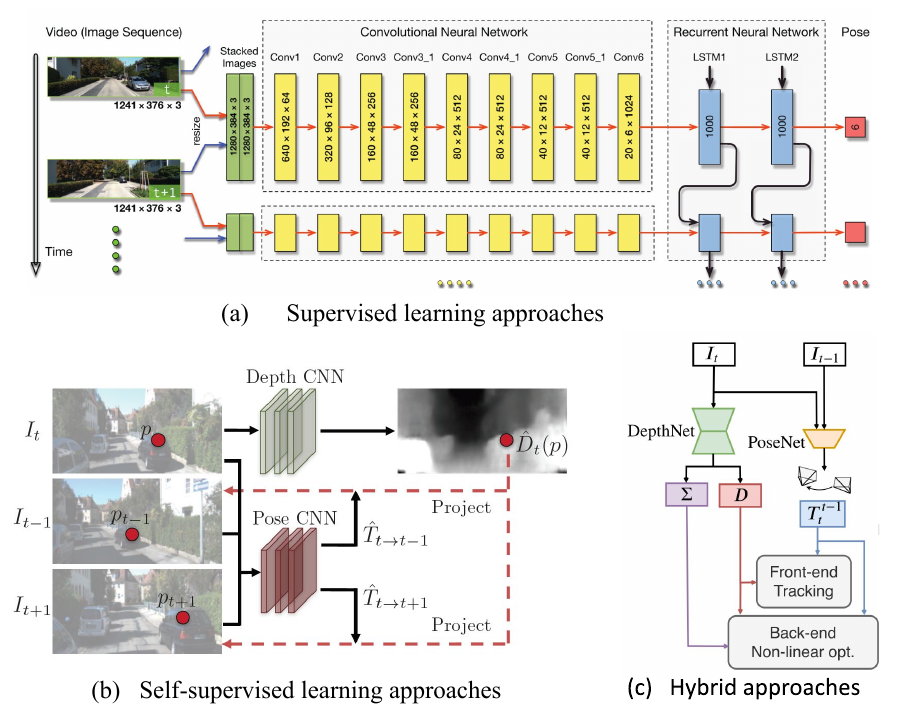}
    \caption{The typical structure of supervised learning of visual odometry (reprint from DeepVO \cite{Wang2017}), self-supervised learning of visual odometry, (reprint from SfmLearner \cite{Zhou2017}) and hybrid visual odometry, (reprint from D3VO \cite{yang2020d3vo}).}
    \label{fig: vo}
    %\vspace{-0.5cm}
\end{figure*}

\subsection{Supervised Learning of Visual Odometry}
Supervised learning based  {visual odometry (VO) methods} aim to train a deep neural network model on labelled datasets to construct a function from consecutive images to motion transformations, instead of exploiting the geometric structures of images as in conventional VO algorithms \cite{scaramuzza2011visual}. 
At its most basic, the input to the deep neural network (DNN) consists of a pair of consecutive images, while the output corresponds to the estimated translation and rotation between the two frames of images.

One of the early works in this area is Konda et al. \cite{Konda2015}. Their approach formulates visual odometry (VO) as a classification problem, and predicts the discrete changes of direction and velocity from input images using a convolutional neural network (ConvNet). However, this method is limited in its ability to estimate the full camera trajectory, and relies on a series of discrete motion estimates instead. Costante et al. \cite{costante2015exploring} propose a method that overcomes some of the limitations of the Konda et al. \cite{Konda2015} approach by using dense optical flow to extract visual features, and then using a ConvNet to estimate the frame-to-frame motion of the camera.  {This method shows performance improvements over the Konda et al. approach, and can generate smoother and more accurate camera trajectories.}
 {Despite the promising results of both approaches, they are not strictly an end-to-end learning model from images to motion estimates, and still fall short of traditional VO algorithms, e.g. VISO2 \cite{geiger2011stereoscan} in terms of accuracy and robustness. One limitation of both methods is that they do not fully exploit the rich geometric information contained in the input images, which is crucial for accurate motion estimation. Furthermore, the datasets used to train and evaluate these approaches are limited in their diversity and may not generalize well to different scenarios.}

 To enable end-to-end learning of visual odometry, DeepVO \cite{Wang2017} utilizes a combination of convolutional neural network (ConvNet) and recurrent neural network (RNN). 
Figure \ref{fig: vo} (a) shows the architecture of this typical RNN+ConvNet based VO model, which extracts visual features from pairs of images via a ConvNet, and passes features through RNNs to model the temporal correlation of features. Its ConvNet encoder is based on a FlowNet \cite{Fischer2015} structure to extract visual features suitable for optical flow and self-motion estimation. 
The recurrent model summarizes history information into its hidden states, so that the output is inferred from both past experience and current ConvNet features from sensor observations.
DeepVO is trained on datasets with groundtruthed poses as training labels. To recover the optimal parameters $\bm{\theta}^{*}$ of this framework, the optimization target is to minimize the Mean Square Error (MSE) of the estimated translations $\mathbf{\hat{p}} \in \mathbb{R}^3$ and Euler angle based rotations $\hat{\bm{\varphi}} \in \mathbb{R}^3$:
    \begin{equation}
        \label{eq: vo}
        \bm{\theta}^{*} = \argmin_{\bm{\theta}} \frac{1}{N} \displaystyle\sum_{i=1}^{N} \displaystyle\sum_{t=1}^{T} \| \hat{\mathbf{p}}_t - \mathbf{p}_t \|_2^2 + \| \hat{\bm{\varphi}}_t - \bm{\varphi}_t \|_2^2 ,
    \end{equation}
where $(\hat{\mathbf{p}}_t, \hat{\bm{\varphi}}_t)$ are the estimates of relative pose from DNN at the timestep $t$, $(\mathbf{p}, \bm{\varphi})$ are the corresponding groundtruth values, $\bm{\theta}$ are the parameters of the DNN framework, and $N$ is the number of samples.
This data-driven solution reports good results on estimating the pose of driving vehicles on several benchmarks. On the KITTI odometry dataset\cite{Geiger2013}, it shows competitive performance over conventional monocular VO, e.g. VISO2 \cite{geiger2011stereoscan} and ORB-SLAM (without loop closure) \cite{Montiel2015}. It is worth noting that supervised VO naturally produces trajectory with absolute scale from monocular camera, while classical monocular VO algorithm is scale-ambiguous. 
This is probably because DNN implicitly learns and maintains the global scale from large collections of images. Although DeepVO reports good results in experimental scenarios, its performance has still not been extensively evaluated by large-scale datasets (e.g., across cities) or real-world experiments/demonstrations in the wild. 

\begin{table*}[t]
\caption{A summary of deep learning  {based visual odometry} (incremental motion estimation) methods {(Section \ref{sec: incremental motion})}.}
\label{tb: odometry estimation}
    \small
    \begin{center}
\begin{tabular}{c l c c c c c l }
\hline
\multicolumn{2}{c}{\multirow{2}{*}{Model}}  & \multirow{2}{*}{Year}         & \multirow{2}{*}{Sensor}  & \multirow{2}{*}{Scale} & \multicolumn{2}{c}{Performance} & \multirow{2}{*}{Contributions} \\ 
\multicolumn{2}{c}{}                                                       &                         &                              &                        & Seq09                  & Seq10        &          \\ \hline
\multirow{8}{*}{Supervised}   
& Konda et al.\cite{Konda2015} & 2015 & MC & Yes & - & - & \footnotesize{formulate VO as a classification problem} \\
& Costante et al.\cite{costante2015exploring} & 2016 & MC & Yes  & \textbf{6.75} & 21.23 & \footnotesize{extract features from optical flow for VO estimates} \\
& DeepVO\cite{Wang2017} & 2017 & MC & Yes & - & 8.11 & \footnotesize{combine RNN and ConvNet for end-to-end learning}\\ 
& Zhao et al.\cite{zhao2018learning} & 2018 & MC & Yes & - & 4.38 & \footnotesize{generate dense 3D flow for VO and mapping}  \\ 
& Saputra et al.\cite{saputra2019learning} & 2019 & MC & Yes & - & 8.29 & \footnotesize{curriculum learning and geometric loss constraints} \\
& Xue et al.\cite{xue2019beyond} & 2019 & MC & Yes & - & \textbf{3.47} & \footnotesize{memory and refinement module} \\ 
& Saputra et al.\cite{saputra2019distilling} & 2019 & MC & Yes & - & - & \footnotesize{knowledge distilling to compress deep VO model}\\ 
& Koumis et al.\cite{koumis2019estimating} & 2019 & MC & Yes  & - & - & \footnotesize{3D convolutional networks}\\
& DAVO \cite{KuoLLLCL20} & 2020 & MC & Yes & - & 5.37 & \footnotesize{Use attention to weight semantics and optical flow}  \\ \hline
\multirow{10}{*}{Self-supervised}  
& SfmLearner\cite{Zhou2017} & 2017 & MC & No & 17.84 & 37.91 & \footnotesize{novel view synthesis for self-supervised learning} \\
& UnDeepVO\cite{li2018undeepvo} & 2018 & SC & Yes & 7.01 & 10.63 & \footnotesize{use fixed stereo line to recover scale metric} \\ 
& GeoNet\cite{Yin2018} & 2018 & MC & No & 43.76 & 35.6 & \footnotesize{geometric consistency loss and 2D flow generator}  \\ 
& Zhan et al.\cite{Zhan2018} & 2018 & SC & Yes & 11.92  & 12.45 & \footnotesize{use fixed stereo line for scale recovery} \\ 
& Struct2Depth\cite{casser2019depth} & 2019 & MC & No & 10.2 & 28.9 & \footnotesize{introduce 3D geometry structure during learning} \\  
& GANVO\cite{almalioglu2019ganvo} & 2019 & MC & No & - & - & \footnotesize{adversarial learning to generate depth}\\ 
& Wang et al.\cite{wang2019recurrent} & 2019 & MC & Yes & 9.30 & 7.21 & \footnotesize{integrate RNN and flow consistency constraint}\\ 
& Li et al.\cite{li2019pose} & 2019 & MC & No & - & - & \footnotesize{global optimization for pose graph}                   \\ 
& Gordon\cite{gordon2019depth} & 2019 & MC & No & {2.7} & {6.8} & \footnotesize{camera matrix learning}\\ 
& Bian et al.\cite{bian2019unsupervised} & 2019 & MC & No & 11.2 & 10.1 & \footnotesize{consistent scale from monocular images}  \\ 
& Li et al.\cite{Li0CXYZ20} & 2020 & MC & No & 5.89 & 4.79 & \footnotesize{meta learning to adapt into new environment}  \\
& Zou et al.\cite{ZouJTHC20} & 2020 & MC & No & 3.49 & 5.81 & \footnotesize{model the long-term dependency}  \\ 
&  {Zhao et al.\cite{zhao2020masked}} &  {2021} &  {MC} &  {No} &  {8.71} &  {9.63} &  {\footnotesize{introduce masked GAN to remove inconsistency}}  \\ 
& Chi et al.\cite{ChiWHGY21} & 2021 & MC & No & 2.02 & 1.81 & \footnotesize{collaborative learning of optical flow,
depth and motion}  \\ 
& Li et al.\cite{LiWCZ21} & 2021 & MC & No & 1.87 & 1.93 & \footnotesize{online adaptation}  \\
&  {Sun et al.\cite{zhao2020masked}} &  {2022} &  {MC} &  {No} &  {7.14} &  {7.72} &  {\footnotesize{introduce cover and filter masks}}  \\ 
& Dai et al. \cite{dai2022self} & 2022 & MC & No & 3.24 & \textbf{1.03} & \footnotesize{introduce attention and pose graph optimization} \\
& VRVO\cite{zhang2022towards} & 2022 & MC & Yes & \textbf{1.55} & 2.75 & \footnotesize{use virtual data to recover scale}  \\
\hline
\multirow{8}{*}{Hybrid}  
& Backprop KF\cite{haarnoja2016} & 2016 & MC & Yes & - & - & \footnotesize{a differentiable Kalman filter based VO}\\
& Yin et al.\cite{yin2017scale} & 2017 & MC & Yes & 4.14 & 1.70 & \footnotesize{introduce learned depth to recover scale metric} \\  
& Barnes et al.\cite{barnes2018driven} & 2018 & MC & Yes & -  & - & \footnotesize{integrate learned depth and ephemeral masks} \\  
& DPF\cite{Jonschkowski2018} & 2018 & MC & Yes & -  & - & \footnotesize{a differentiable particle filter based VO}\\ 
& Yang et al.\cite{yang2018deep} & 2018 & MC & Yes & 0.83 & 0.74 & \footnotesize{use learned depth into classical VO} \\
& CNN-SVO\cite{loo2019cnn} & 2019 & MC & Yes & 10.69 & 4.84 & \footnotesize{use learned depth to initialize SVO}  \\ 
& Zhan et al.\cite{zhan2020visual} & 2020 & MC & Yes  & 2.61 & 2.29 & \footnotesize{integrate learned optical flow and depth}              \\ 
& Wagstaff et al.\cite{WagstaffPK20} & 2020 & MC & Yes  & 2.82 & 3.81 & \footnotesize{integrate classical VO with learned pose corrections}              \\ 
& D3VO\cite{yang2020d3vo} & 2020 & MC & Yes & \textbf{0.78} & \textbf{0.62} & \footnotesize{integrate learned depth, uncertainty and pose} \\
& Sun et al.\cite{sun2022improving} & 2022 & MC & Yes & - & - & \footnotesize{integrate learned depth into DSO} 
\\  \hline
\end{tabular}
%\vspace{-0.3cm}
\end{center}
    \begin{itemize}
              \footnotesize{
                {\item \textit{Year} indicates the publication year (e.g. the date of conference) of each work.}
                \item \textit{Sensor:} MC and SC represent monocular camera and stereo camera respectively.
                \item \textit{Supervision} represents whether it is a supervised or unsupervised end-to-end model, or a hybrid model
                \item \textit{Scale} indicates whether a trajectory with a global scale can be produced.
                \item \textit{Performance} reports the localization error (a small number is better), i.e. the averaged translational RMSE drift (\%) on lengths of 100m-800m  on the KITTI odometry dataset\cite{Geiger2013}. Most works were evaluated on the Sequence 09 and 10, and thus we took the results on these two sequences from their original papers for a performance comparison. Note that the training sets may be different in each work. 
                %\item \textit{Contributions} summarize the main contributions of each work compared with previous research.
            }
    \end{itemize}
    %\vspace{-0.3cm}
\end{table*}

 {Enhancing the generalization capability of supervised Visual Odometry (VO) models and improving their efficacy for operating in real-time on devices with limited resources are still formidable challenges. While supervised learning-based VO is trained on extensive datasets of image sequences with ground-truth poses, not all sequences are equally informative or challenging for the model to learn. Curriculum learning is a technique that gradually elevates the complexity of the training data by initially presenting simple sequences and progressively introducing more challenging ones. In \cite{saputra2019learning}, curriculum learning is integrated into the supervised VO model by increasing the amount of motion and rotation in the training sequences, enabling the model to learn to estimate camera motion more robustly and generalize better to new data. Knowledge distillation is another approach that can be introduced to improve the efficiency of supervised VO models by compressing a large model through teaching a smaller one. This method is applied in \cite{saputra2019distilling}, reducing the number of network parameters and making the model more suitable for real-time operation on mobile devices. Compared to pure supervised VO without knowledge distillation, this method significantly reduces network parameters by 92.95\% and enhances computation speed by 2.12 times.}

 {Furthermore, to enhance the localization performance, a memory module that stores global information about the scene and camera motion is introduced in \cite{xue2019beyond}. The background information is then utilized by a refining module that enhances the accuracy of the predicted camera poses. Additionally, attention mechanisms have been implemented to weigh the inputs from different sources and enhance the efficacy of supervised VO models. For example, DAVO \cite{KuoLLLCL20} integrates an attention module to weigh the inputs from semantic segmentation, optical flow, and RGB images, leading to improved odometry estimation performance. Despite the promising end-to-end learning performance achieved on publicly available datasets by these supervised VO frameworks, their deployment performance in real-world scenarios remains to be further verified as of the writing of this survey.}

 {Overall, supervised learning-based visual odometry models primarily rely on ConvNet or RNN to learn pose transformations automatically from raw images. Recent advancements in machine learning, including attention mechanisms, GANs, and knowledge distillation, have allowed these models to extract more expressive visual features and accurately model motion. However, these learning methods often require a vast amount of training data with precise poses as labels to optimize model parameters and improve robustness.
While supervised learning-based VO models have demonstrated promising end-to-end learning performance on publicly available datasets, their deployment performance in real-world scenarios requires further validation.} Additionally, obtaining labeled data is often time-consuming and costly, and inaccurate labels can occur. In the following section, we will discuss recent efforts to address the issue of label scarcity through self-supervised learning techniques.

\subsection{Self-supervised Learning of Visual Odometry}
\label{sec: unsupervised vo}
There are growing interests in exploring self-supervised learning of visual odometry (VO).
Self-supervised solutions are capable of exploiting unlabelled sensor data, and thus it saves human efforts. Compared with supervised approaches, they normally show better adaptation ability in new scenarios, where no labelled data are available. 
This has been achieved in a self-supervised framework that jointly learns camera ego-motion and depth from video sequences, by utilizing view synthesis as a self-supervisory signal \cite{Zhou2017}.

As shown in Figure \ref{fig: vo} (b), a typical self-supervised VO framework \cite{Zhou2017} consists of a depth network to predict depth maps, and a pose network to produce motion transformations between images. 
The entire framework takes consecutive images as input, and the supervision signal is based on novel view synthesis -
given a source image $\mathbf{I}_s$, the view synthesis task is to generate a synthetic target image $\mathbf{I}_t$. A pixel of source image $\mathbf{I}_s(p_s)$ is projected onto a target view $\mathbf{I}_t(p_t)$ via: 
    \begin{equation}
        \label{eq: un vo}
        p_s \sim \mathbf{K} \mathbf{T}_{t \to s} \mathbf{D}_t(p_t) \mathbf{K}^{-1} p_t
    \end{equation}
where $\mathbf{K}$ is the camera's intrinsic matrix, $\mathbf{T}_{t \to s}$ denotes the camera motion matrix from target frame to source frame, and $\mathbf{D}_t(p_t)$ denotes the per-pixel depth maps in the target frame.
The training objective is to ensure the consistency of the scene geometry by optimizing the photometric reconstruction loss between the real target image and the synthetic one: 
    \begin{equation}
        \mathcal{L}_{\text{photo}} = \sum_{<\mathbf{I}_1, ..., \mathbf{I}_N>\in S} \sum_{p} | \mathbf{I}_t(p) - \hat{\mathbf{I}}_s(p) |,
    \end{equation}
where $p$ denotes pixel coordinates, $\mathbf{I}_t$ is the target image, and $\hat{\mathbf{I}}_s$ is the synthetic target image generated from the source image $\mathbf{I}_s$.

However, there are basically two main problems that remain unsolved in the original work \cite{Zhou2017}: 
1) this monocular image based approach is not able to provide pose estimates in a consistent global scale. Due to the scale ambiguity, no physically meaningful global trajectory can be reconstructed, limiting its real usage;
2) the photometric loss assumes that the scene is static and without camera occlusions. Although the authors propose the use of an explainability mask to remove scene dynamics, the influence of these environmental factors is still not addressed completely, which violates the assumption.

To solve global scale problem, \cite{li2018undeepvo,Zhan2018} propose to utilize stereo image pairs to recover the absolute scale of pose estimation. They introduce an additional spatial photometric loss between the left and right pairs of images, as the stereo baseline (i.e. motion transformation between the left and right images) is fixed and known throughout the dataset. 
Once the training is complete, the network produces pose predictions using only monocular images.  {Compared with \cite{Zhou2017}, they are able to produce  camera poses with global metric scale and higher accuracy.} 
 {Another approach is to use virtual stereo data from simulator to recover the absolute scale of pose estimation in VRVO\cite{zhang2022towards}.
It utilizes a generative adversarial network (GAN) to generate virtual stereo data that is similar to real-world data. By bridging the gap between virtual and real data using adversarial learning, the pose network is then trained using the virtual data to recover the absolute scale of pose estimation.}
\cite{bian2019unsupervised} tackles the scale issue by introducing a geometric consistency loss, that enforces the consistency between predicted depth maps and reconstructed depth maps. The framework transforms the predicted depth maps into a 3D space, and projects them back to produce reconstructed depth maps. By doing so, the depth predictions can remain scale-consistent over consecutive frames, enabling pose estimates to be scale-consistent as well.  {Different from previous works that either use stereo images \cite{li2018undeepvo,Zhan2018} or virtual data \cite{zhang2022towards}, this work successfully produces scale-consistent camera poses and depth estimates only using monocular images.}

The photometric consistency constraint is based on the assumption that the entire scene consists only of rigid static structures such as buildings and lanes. However, in real-world applications, the presence of environmental dynamics such as pedestrians and vehicles can cause distortion in the photometric projection, leading to reduced accuracy in pose estimation. To address this concern, GeoNet \cite{Yin2018} divides its learning process into two sub-tasks by estimating static scene structures and motion dynamics separately through a rigid structure reconstructor and a non-rigid motion localizer. In addition, GeoNet enforces a geometric consistency loss to mitigate the issues caused by camera occlusions and non-Lambertian surfaces. \cite{zhao2018learning} adds a 2D flow generator along with a depth network to generate 3D flow. Benefiting from better 3D understanding of environment, this framework is able to produce more accurate camera poses, along with a point cloud map. GANVO \cite{almalioglu2019ganvo} employs a generative adversarial learning paradigm for depth generation and introduces a temporal recurrent module for pose regression.  {This method improves accuracy in depth map and pose estimation, as well as tolerating environmental dynamics.}
\cite{li2019sequential} also utilizes a generative adversarial network (GAN) to generate more realistic depth maps and poses, and further encourages more accurate synthetic images in the target frame.  {Unlike hand-crafted metrics}, a discriminator is used to evaluate the quality of synthetic image generation. In doing so, the generative adversarial setup facilitates the generated depth maps to be more texture-rich and crisper. In this way, high-level scene perception and representation are accurately captured and environmental dynamics are implicitly tolerated. 
 {\cite{zhao2020masked} introduces a masked GAN into joint learning of depth and visual odometry (VO) estimation, addressing influences from light-condition changes and occlusions. By incorporating MaskNet and a Boolean mask scheme, it mitigates the impacts of occlusions and visual field changes, improving adversarial loss and image reconstruction. A scale-consistency loss ensures accurate pose estimation in long monocular sequences. Similarly,  \cite{sun2021unsupervised} introduces hybrid masks to mitigate the negative impact of dynamic environments. Cover masks and filter masks alleviate adverse effects on VO estimation and view reconstruction processes. Both approaches demonstrate competitive depth prediction and globally consistent VO estimation in car-driving scenarios.}

 Recent attempts \cite{Li0CXYZ20,LiWCZ21} design online learning strategies that enable the learned model to adapt into new environments.  {These approaches allow the learning model to automatically update its parameters and learn from new data without forgetting the previously learned knowledge. }  {Collaborative learning of multiple learning tasks, such as optical flow, depth, and camera motion estimation, has also been shown to improve the performance of self-supervised VO \cite{ChiWHGY21}. By jointly optimizing the different learning targets, it exploits the complementary information between them so that learns more robust representations for pose estimation. To further improve visual odometry,
 \cite{dai2022self} proposes a self-supervised VO with an attention mechanism and pose graph optimization. The introduced attention mechanism is sensitive to geometrical structure and helps to accurately regress the rotation matrix.
 }

 {Overall, self-supervised learning-based visual odometry methods have emerged as a promising approach for estimating camera poses and scene depths without the need for labeled data during training. They normally consist of two ConvNet based neural networks- one for depth estimation and the other for pose estimation. Compared to supervised learning-based methods, self-supervised approaches offer several advantages, including the ability to handle non-rigid dynamics and adapt to new environments in real-time. However, despite these benefits, self-supervised VO methods still underperform compared to their supervised counterparts, and there remain challenges with scaling and scene dynamics.}

 As demonstrated in Table \ref{tb: odometry estimation}, self-supervised VO still cannot compete with supervised VO in performance, its concerns of scale metric and scene dynamics problem have been largely resolved with the efforts of many researchers. With the benefits of self-supervised learning and ever-increasing improvement on performance, self-supervised VO would be a promising solution to deep learning based SLAM. Currently, end-to-end learning based VOs have not been proved to surpass the state-of-the-art model-based VOs in performance. Next section will show how to combine the benefits from both sides to construct hybrid approaches.

\subsection{Hybrid Visual Odometry}
Unlike end-to-end approaches that rely solely on a deep neural network to interpret pose from data, hybrid approaches combine classical geometric models with a deep learning framework. The deep neural network is used to replace part of a geometry model, which allows for more expressive representations.

One of the key challenges in traditional monocular visual odometry (VO) is the scale-ambiguity problem, where monocular VOs can only estimate relative scale. This poses a problem in scenarios where absolute scale is required. One way to solve this issue is to integrate learned depth estimates into a classical visual odometry algorithm, which helps to recover the absolute scale metric of poses. Depth estimation is a well-established research area in computer vision, and various methods have been proposed to tackle this problem. For instance, Godard et al. \cite{godard2017unsupervised} proposed a deep neural model that predicts per-pixel depths in an absolute scale. The details of depth learning are discussed in Section \ref{sec: depth}.

In \cite{yin2017scale}, a ConvNet produces coarse depth values from raw images, which are then refined by conditional random fields. The scale factor is calculated by comparing the estimated depth predictions with the observed point positions.
Once the scale factor is obtained, the ego-motions with absolute scale are obtained by multiplying the scale-factor and estimated translations from a monocular VO algorithm. This approach mitigates the scale problem by incorporating depth information. 
Additionally, \cite{barnes2018driven} proposes the integration of predicted ephemeral masks (i.e., the area of moving objects) with depth maps in a traditional VO system to enhance its robustness to moving objects.  {This method enables the system to produce metric-scale pose estimates using a single camera, even when a significant portion of the image is obscured by dynamic objects.}
\cite{WagstaffPK20} proposes to combine a classical VO with learned pose corrections, that largely reduces the error drifts of classical VOs.  {Compared with pure learning based VOs, instead of directly regressing inter-frame pose changes, this approach regresses pose corrections from data, without the need of pose ground truth as training data.}
 {Similarly, \cite{sun2022improving} proposes to improve classical monocular VO with learned depth estimates. This framework consists of a monocular depth estimation module with two separate working modes to assist localization and mapping, and it demonstrates strong generalization ability to diverse scenes, compared with existing learning based VOs.}
Furthermore, \cite{zhan2020visual} integrates learned depth and optical flow predictions into a conventional VO model.  {Specifically, this framework uses optical flow and single-view depth predictions from deep ConvNets as intermediate outputs to establish 2D-2D/3D-2D correspondences, and the depth estimates with consistent scale can mitigate the scale drift issue in monocular VO/SLAM systems. By integrating deep predictions with geometry-based methods, the study shows that deep VO models can complement standard VO/SLAM systems.}

D3VO \cite{yang2020d3vo} is proposed to incorporate the predictions of depth, pose, and photometric uncertainty from deep neural networks into direct visual odometry (DVO) \cite{engel2017direct}. In D3VO, a self-supervised framework is employed to learn depth and ego-motion jointly, similar to the approaches discussed in Section \ref{sec: unsupervised vo}. D3VO employs the uncertainty estimation method proposed by \cite{kendall2017uncertainties} to  {generate a photometric uncertainty map that indicates which parts of the visual observations can be trusted.}
As illustrated in Fig. \ref{fig: vo} (c), the learned depth and pose estimates are integrated into the front-end of a VO algorithm, and the uncertainties are used in the system back-end. This method shows impressive results on the KITTI \cite{Geiger2013} and EuroC \cite{Burri2016} benchmarks, surpassing several popular conventional VO/VIO systems, e.g. DSO \cite{vi-dso}, ORB-SLAM \cite{Montiel2015} and VINS-Mono \cite{Qin2018}. This indicates the promise of integrating learning methods with geometric models.

In addition to geometric models, there have been studies that combine physical motion models with deep neural networks, such as through a differentiable Kalman filter  {\cite{haarnoja2016,chen2021dynanet}} or a differentiable particle filter \cite{Jonschkowski2018}. In \cite{haarnoja2016}, Kalman filter is transformed into a differentiable module that is combined with deep neural networks for an end-to-end training. \cite{chen2021dynanet} proposes DynaNet, a hybrid model integrating deep neural networks (DNNs) and state-space models (SSMs) to leverage their strengths. DynaNet enhances interpretability and robustness in car-driving scenarios by combining powerful feature representation from DNNs with explicit modeling of physical processes from SSMs. The incorporation of a recursive Kalman filter enables optimal filtering on the feature state space, facilitating accurate positioning estimation, and showcasing its ability to detect failures through internal filtering model parameters such as the rate of innovation (Kalman gain). Instead of Kalman filter, \cite{Jonschkowski2018} presents a differentiable particle filter with learnable motion and measurement models. The proposed differentiable particle filter can approximate complex nonlinear functions, allowing for efficient training of motion models by optimizing state estimation performance. Both two works incorporate the physical motion model of visual odometry into the state update process of filtering. Thus, the physical model serves as an algorithmic prior in the learning process.   Compared with ConvNet or LSTM based models, differentiable filters improve the data efficiency and generalization ability of the learning based motion estimation.

In summary, hybrid models that combine geometric or physical priors with deep learning techniques are generally more accurate than end-to-end VO/SLAM systems and can even outperform conventional monocular VO systems on common benchmarks. Geometry-based models integrate deep neural networks into VO/SLAM pipelines to improve depth and egomotion estimation, as well as increase robustness to dynamic objects. Physical motion-based models combine deep neural networks with physical motion models, such as the Kalman filter or particle filter, to integrate the physical motion model of VO/SLAM systems into the learning process. Combining the benefits from combining geometric or physical priors with deep learning, hybrid models are normally more accurate than end-to-end VO at this stage, as shown in Table \ref{tb: odometry estimation}. It is notable that recent hybrid models even outperform some representative conventional monocular VO systems on common benchmarks \cite{yang2020d3vo}. This demonstrates the rapid rate of progress in this area.

\subsection{Performance Comparison of Deep Learning based Visual Odometry (VO) Methods} 
Table \ref{tb: odometry estimation}  {presents a comprehensive comparison of existing works focusing on deep learning-based visual odometry (VO). The table includes information regarding the sensor type utilized, the model employed, whether the method produces a trajectory with an absolute scale, and the performance evaluation conducted on the KITTI dataset. A concise overview of the contribution made by each model is also provided. The KITTI dataset \cite{Geiger2013} serves as a widely recognized benchmark for visual odometry estimation and comprises a collection of sensor data captured during car-driving scenarios.} As most deep learning based approaches use the trajectory 09 and 10 of the KITTI dataset to test a trained model, we compared them according to the averaged Root Mean Square Error (RMSE) of the translation for all the subsequences of lengths (100, 200, .., 800) meters, which is provided by the official KITTI VO/SLAM evaluation metrics. 

Hybrid  {VO}  {models demonstrate superior performance compared to both supervised and unsupervised VO approaches. This is attributed to the hybrid model's ability to leverage the well-established geometry models of traditional VO algorithms alongside the powerful feature extraction capabilities offered by deep learning methods.} Although supervised VO models still outperform unsupervised approaches, the performance gap between them is diminishing as the limitations of self-supervised VO methods are gradually addressed. Notably,  {recent advancements have shown that self-supervised VO can now recover scale-consistent poses from monocular images} \cite{bian2019unsupervised}. Overall, data-driven visual odometry shows a remarkable increase in model performance, indicating the potentials of deep learning approaches in achieving more accurate  {visual odometry} estimation in the future. However, it is worth noting that this upward trend is not always consistent, as several published papers focus on addressing intrinsic issues within learning frameworks rather than solely aiming to achieve the best performance.

\begin{table*}[t]
\caption{A summary on existing methods on deep learning for relocalization in 2D map {(Section \ref{sec: 2d localization})}}
\label{tb: learning for 2d-2d localization}
    \small
    \begin{center}
\begin{tabular}{c c c l c c c l }
\hline
\multicolumn{3}{c}{\multirow{2}{*}{Model}} & \multirow{2}{*}{Year} & \multirow{2}{*}{Agnostic} & \multicolumn{2}{c}{Performance (m/degree)} & \multicolumn{1}{c}{\multirow{2}{*}{Contributions}} \\
\multicolumn{3}{c}{} & & & 7Scenes & Cambridge & \multicolumn{1}{c}{}\\ \hline
\multirow{26}{*}{\rotatebox{90}{Relocalization in 2D Map}}
& \multirow{6}{*}{\rotatebox{90}{Explicit Map}} 
& NN-Net  \cite{laskar2017camera} & 2017 & Yes & 0.21/9.30  & - & \footnotesize{combine retrieval and relative pose estimation} \\
&& DeLS-3D \cite{wang2018dels} & 2018 & No  & -  & - & \footnotesize{jointly learn with semantics} \\ 
&& AnchorNet \cite{saha2018improved} & 2018 & Yes & 0.09/6.74  & 0.84/2.10 & \footnotesize{anchor point allocation} \\
&& RelocNet \cite{balntas2018relocnet} & 2018 & Yes & 0.21/6.73 & - & \footnotesize{camera frustum overlap loss} \\
&& CamNet \cite{ding2019camnet} & 2019 & Yes & 0.04/1.69  & - & \footnotesize{multi-stage image retrieval} \\
&& PixLoc \cite{sarlin2021back} & 2021 & Yes & \textbf{0.03}/\textbf{0.98}  & \textbf{0.15}/\textbf{0.25} & \footnotesize{cast camera localization as metric learning} \\
\cline{2-7}
& \multirow{20}{*}{\rotatebox{90}{Implicit Map}}                              
& PoseNet \cite{kendall2015posenet} & 2015 & No  & 0.44/10.44 & 2.09/6.84 & \footnotesize{first neural network in global pose regression} \\ 
&& Bayesian PoseNet \cite{kendall2016modelling} & 2016 & No  & 0.47/9.81  & 1.92/6.28 & \footnotesize{estimate Bayesian uncertainty for global pose}\\ 
&& BranchNet \cite{wu2017delving} & 2017 & No  & 0.29/8.30  & - & \footnotesize{multi-task learning for orientation and translation}\\ 
&& VidLoc \cite{Clark2017} & 2017 & No  & 0.25/- & - & \footnotesize{efficient localization from image sequences}\\ 
&& Geometric PoseNet \cite{kendall2017geometric} & 2017 & No  & 0.23/8.12  & 1.63/2.86 & \footnotesize{geometry-aware loss} \\
&& SVS-Pose \cite{naseer2017deep} & 2017 & No  & - & 1.33/5.17 & \footnotesize{data augmentation in 3D space} \\  
&& LSTM PoseNet \cite{walch2017image} & 2017 & No  & 0.31/9.85  & 1.30/5.52 & \footnotesize{spatial correlation} \\ 
&& Hourglass PoseNet \cite{melekhov2017image} & 2017 & No  & 0.23/9.53  & - & \footnotesize{hourglass-shaped architecture} \\
%&& VLocNet \cite{valada2018deep} & 2018 & No  & 0.05/3.80  & \textbf{0.78}/2.82 & \footnotesize{jointly learn global localization and odometry}\\ 
&& MapNet \cite{Brahmbhatt2018} & 2018 & No  & 0.21/7.77  & 1.63/3.64 & \footnotesize{impose spatial and temporal constraints}\\ 
&& SPP-Net \cite{purkait2018synthetic} & 2018 & No  & \textbf{0.18}/6.20  & 1.24/2.68 & \footnotesize{synthetic data augmentation} \\ 
&& GPoseNet \cite{cai2018hybrid} & 2018 & No  & 0.30/9.90  & 2.00/4.60 & \footnotesize{hybrid model with Gaussian Process Regressor} \\
%&& VLocNet++ \cite{radwan2018vlocnet++} & 2018 & No  & 0.02/\textbf{1.39}  & - & \footnotesize{jointly learn with odometry and semantics} \\ 
&& LSG \cite{xue2019local} & 2019 & No  & 0.19/7.47  & -  & \footnotesize{odometry-aided localization}\\  
&& PVL \cite{huang2019prior} & 2019 & No & - & 1.60/4.21 & \footnotesize{prior-guided dropout mask to improve robustness}\\
&& AdPR \cite{bui2019adversarial} & 2019 & No  & 0.22/8.8   & -  & \footnotesize{adversarial architecture}\\ 
&& AtLoc \cite{wang2019atloc} & 2019 & No  & 0.20/7.56  & -  & \footnotesize{attention-guided spatial correlation}\\ 
&& GR-Net \cite{xue2020learning} & 2020 & No  & 0.19/\textbf{6.33}  & 1.12/\textbf{2.40}  & \footnotesize{construct a view graph}\\
%&& MSPN \cite{blanton2020extending} & 2020 & Yes  & 0.20/8.41  & 2.47/5.34  & \footnotesize{extend to multiple scenes}\\
&& MS-Transformer~\cite{shavit2021learning} & 2021 & Yes  & \textbf{0.18}/ 7.28  & 1.28/2.73  & \footnotesize{extend to multiple scenes with transformers}
\\ \hline
\end{tabular}
\end{center}
\begin{itemize}
              \footnotesize{
              {\item \textit{Year} indicates the publication year (e.g. the date of conference) of each work.}
                \item \textit{Agnostic} indicates whether it can generalize to new scenarios.
                \item \textit{Performance} reports the position (m) and orientation (degree) error (a small number is better) on the 7-Scenes (Indoor)\cite{shotton2013scene} and Cambridge (Outdoor) dataset\cite{kendall2015posenet}. Both datasets are split into training and testing set. We report the averaged error on the testing set.
                %\item \textit{Contributions} summarize the main contributions of each work compared with previous research.
            }
    \end{itemize}
    %\vspace{-0.5cm}
\end{table*}

\section{Global Relocalization}
Global relocalization is the process of determining the absolute camera pose within a known scene. Different from incremental motion estimation  {(visual odometry)} that can  {perform in unfamiliar environments}, global relocalization  {relies on prior knowledge of the scene and utilizes} a 2D or 3D scene model. Basically, it establishes the relation between sensor observations and the map by matching a query image or view against a pre-built model, followed by returning an estimate of the global pose. According to the type of map  {used}, deep learning-based methods for global relocalization can be categorized into two categories: \emph{Relocalization in a 2D Map}, where input 2D images are matched against a database of geo-referenced images or an implicit neural map; \emph{Relocalization in a 3D Map}, where correspondences are established between 2D image pixels and 3D points from an explicit or implicit scene model. Table \ref{tb: learning for 2d-2d localization} and \ref{tb: learning for 2d-3d localization} summarize the existing approaches in deep learning based global relocalization within a 2D map or a 3D map, respectively.

\subsection{{Relocalization in a 2D Map}}
\label{sec: 2d localization}
Relocalization in a 2D map involves estimating the image pose relative to a 2D map. This type of map can be created explicitly using a geo-referenced database or implicitly encoded within a neural network.

\begin{figure}
    \centering
    \includegraphics[width=0.48\textwidth]{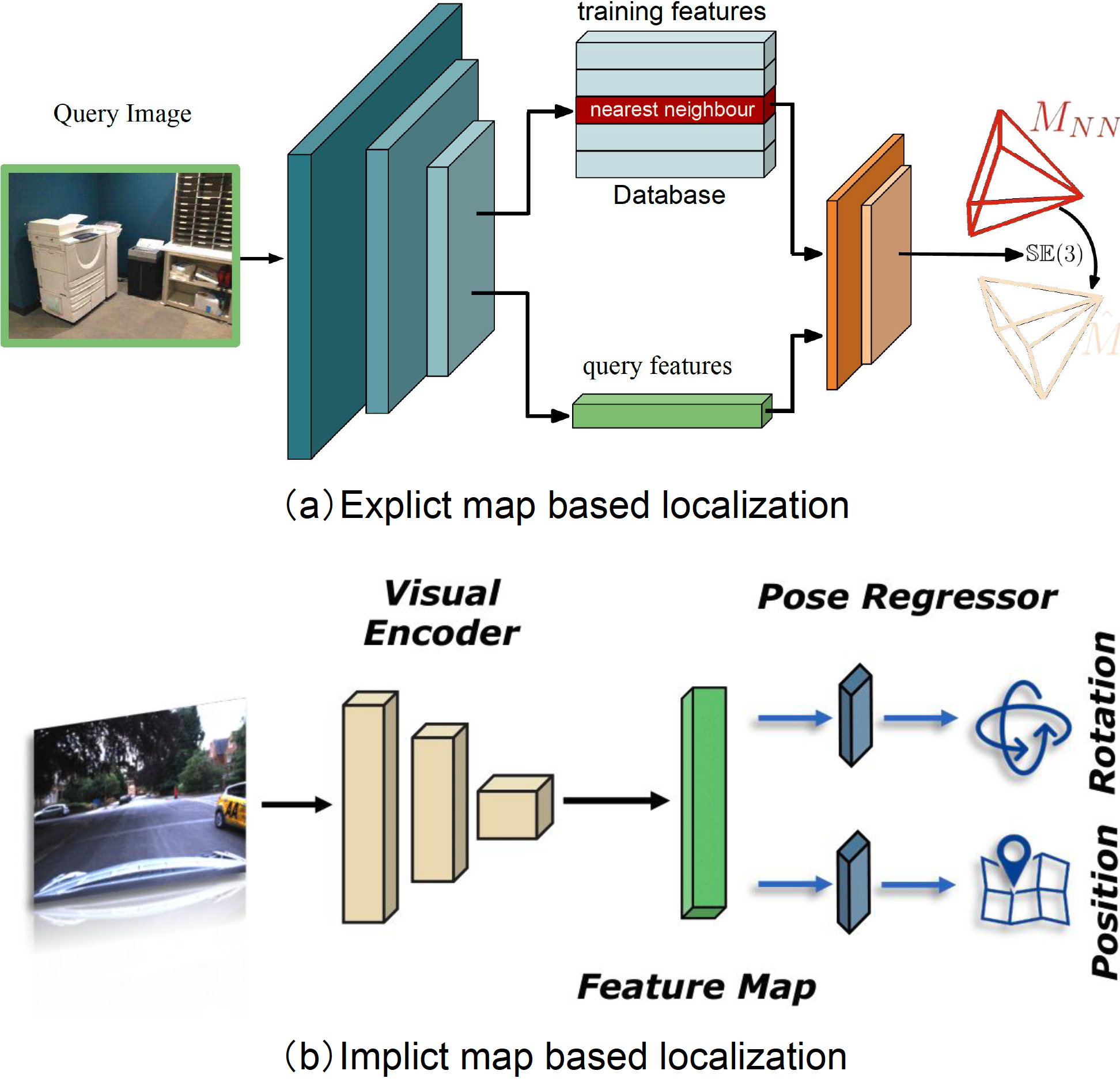}
    %\vspace{-0.2cm}
    \caption{The typical architectures of relocalization in 2D map through (a) explicit map, i.e. RelocNet \cite{balntas2018relocnet} and (b) implicit map, i.e. e.g. PoseNet \cite{kendall2015posenet}}
    \label{fig: 2d2d}
    %\vspace{-0.3cm}
\end{figure}

\subsubsection{Explicit 2D Map Based Relocalization}
Explicit 2D map based relocalization typically represents a scene by a database of geo-tagged images (references) \cite{torii201524, ge2020self, thoma2020soft}. Figure \ref{fig: 2d2d} (a) illustrates the two stages of this relocalization process with 2D references: image retrieval and pose regression. 

 {In the first stage, the goal is to determine the most relevant part of the scene represented by reference images to the visual query. This is achieved by finding suitable image descriptors for image retrieval, which is a challenging task.} Deep learning-based approaches \cite{chen2014convolutional,sunderhauf2015performance} use pre-trained convolutional neural networks (ConvNets) to extract image-level features that are invariant to changes in viewpoint, lighting, and other factors that can affect image appearance. In challenging situations, local descriptors are extracted and aggregated to obtain robust global descriptors. For instance, NetVLAD \cite{arandjelovic2016netvlad} uses a trainable generalized VLAD layer (Vector of Locally Aggregated Descriptors \cite{jegou2010aggregating}, a descriptor vector used in image retrieval), while CamNet \cite{ding2019camnet} applies a two-stage retrieval approach that combines image-based coarse retrieval and pose-based fine retrieval to select the most similar reference frames for the final precise pose estimation.

 The second stage of explicit 2D map-based relocalization aims to obtain more precise poses of the queries by performing additional relative pose estimation with respect to the retrieved images. Traditionally, this is tackled by epipolar geometry, relying on the 2D-2D correspondences determined by local descriptors \cite{zhou2019learn, melekhov2019dgc, zhuang2021fusing}. In contrast, deep learning-based approaches regress the relative poses directly from pairwise images. For example, NN-Net \cite{laskar2017camera} uses a neural network to estimate the pairwise relative poses between the query and the top N ranked references, followed by a triangulation-based fusion algorithm that coalesces the predicted N relative poses and the ground truth of 3D geometry poses to obtain the absolute query pose. Alternatively, RelocNet \cite{balntas2018relocnet} introduces a frustum overlap loss to assist global descriptors learning that are suitable for camera localization.
 
 Explicit 2D map-based relocalization is scalable and flexible, as it does not require training on specific scenarios. However, maintaining a database of geo-tagged images and accurate image retrieval can be challenging, making it difficult to scale to large-scale scenarios. Moreover, explicit 2D map-based relocalization is normally time-consuming compared to implicit-map-based counterparts, which will be discussed in the next section.

\subsubsection{Implicit 2D Map Based Relocalization}
Implicit 2D map based relocalization directly regresses camera pose from single images, by implicitly representing a 2D map inside a deep neural network. The common pipeline is illustrated in Figure \ref{fig: 2d2d} (b) - the input to a neural network is single images, while the output is the global position and orientation of query images.

PoseNet \cite{kendall2015posenet} is the first approach to tackle the camera relocalization problem by training a ConvNet to predict camera pose from single RGB images in an end-to-end manner. It leverages the main structure of GoogleNet \cite{szegedy2015going} to extract visual features and removes the last softmax layers. Instead, a fully connected layer is introduced to output a 7 dimensional global pose, which consists of position and orientation vectors in 3 and 4 dimensions, respectively. However, PoseNet has some limitations. It is designed with a naive regression loss function that does not take into account the underlying geometry of the problem. This leads to hyper-parameters requiring expensive hand-engineering to be tuned, and it may not generalize well to new scenes. Additionally, due to the high dimensionality of the feature embedding and limited training data, PoseNet suffers from overfitting problems.

Various extensions are proposed to enhance the original pipeline, for example, by exploiting LSTM units to reduce the dimensionality \cite{walch2017image}, applying synthetic generation to augment training data \cite{naseer2017deep, wu2017delving, purkait2018synthetic, zhu2021learning}, replacing the backbone \cite{melekhov2017image}, modelling pose uncertainty \cite{kendall2016modelling, cai2018hybrid, bui20206d}, introducing geometry-aware loss function \cite{kendall2017geometric} and associating features via an attention mechanism \cite{wang2019atloc}. A prior guided dropout mask is additionally adopted in RVL \cite{huang2019prior} to further eliminate the uncertainty caused by dynamic objects. VidLoc \cite{Clark2017} incorporates temporal constraints of image sequences to model the temporal connections of input images for visual localization. Moreover, additional motion constraints, including spatial constraints and other sensor constraints from GPS or SLAM systems are exploited in MapNet \cite{Brahmbhatt2018}, to enforce the motion consistency between predicted poses. Similar motion constraints are also introduced by jointly optimizing a relocalization network and visual odometry network \cite{valada2018deep, xue2019local, tian20203d}. However, being application-specific, scene representations learned from localization tasks may ignore some useful features they are not designed for. To this end, VLocNet++ \cite{radwan2018vlocnet++} additionally exploits the inter-task relationship between learning semantics and regressing poses, achieving impressive results. More recently, Graph Neural Networks (GNNs) are introduced to tackle the multi-view camera relocalization task in GR-Net~\cite{xue2020learning} and PoGO-Net~\cite{li2021pogo}, enabling the messages of different frames to be transferred beyond temporal connections. MS-Transformer~\cite{shavit2021learning} 
extends the absolute pose regression paradigm for learning a single model on multiple scenes.

 {Both explicit and implicit 2D map-based relocalization methods exploit the benefits of deep learning in automatically extracting crucial features for global relocalization in environments lacking distinctive features. Implicit map-based learning approaches directly regress the absolute pose of a camera through a DNN, making them easier to implement and more efficient than explicit map-based learning approaches. However, current implicit map-based approaches exhibit performance limitations, and their dependence on scene-specific training prevents them from generalizing to unfamiliar scenes without necessitating retraining.} In the next section, we will introduce the concept of learning to match images against a 3D model for global relocalization.

\subsection{{Relocalization in a 3D Map}}
\label{sec: 3d relocalization}
\begin{figure}
    \centering
    \includegraphics[width=0.48\textwidth]{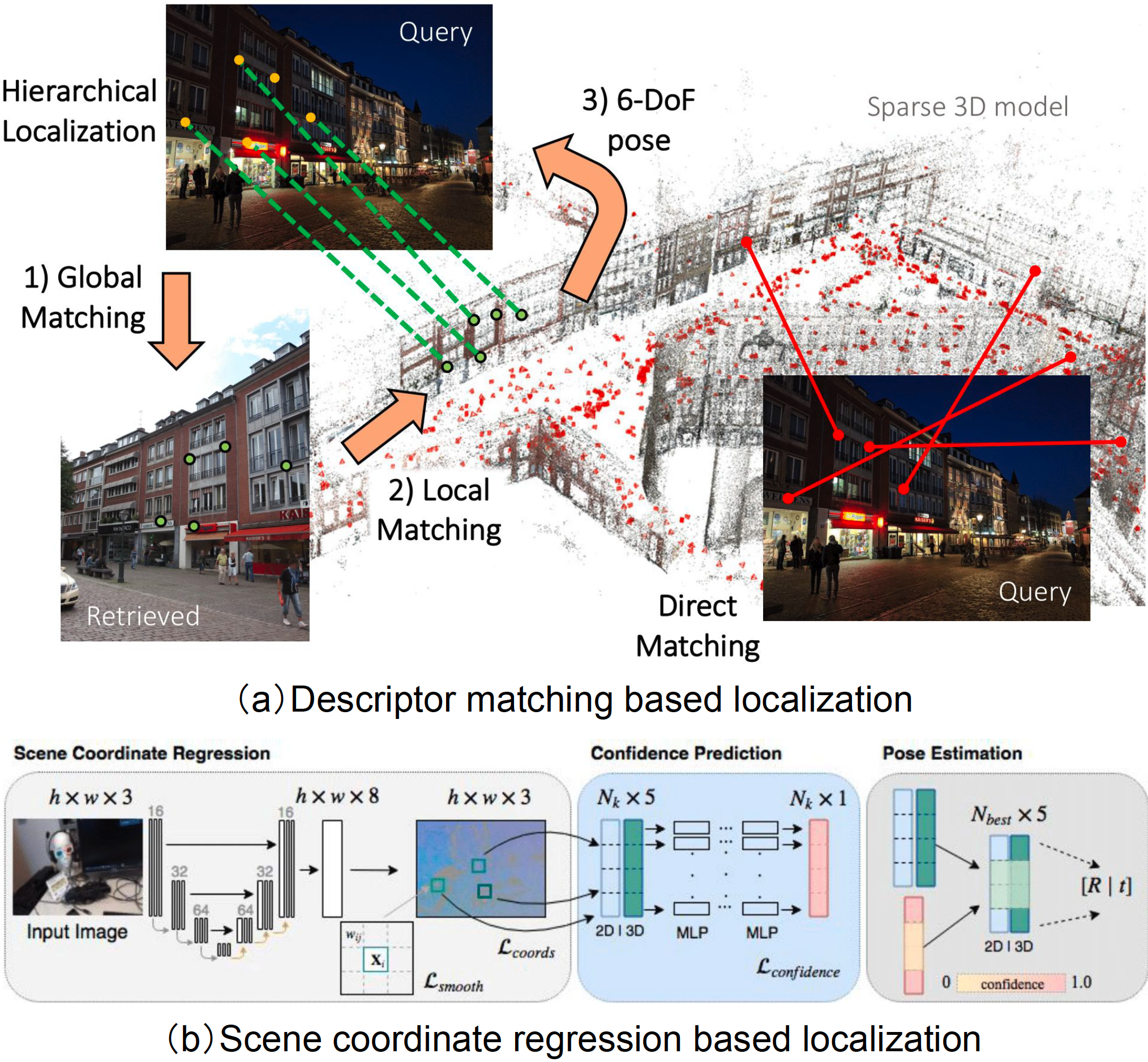}
    \caption{The typical architectures of 3D Map based relocalization through (a) descriptor matching, i.e. HF-Net \cite{sarlin2019coarse} and (b) scene coordinate regression, i.e. Confidence SCR \cite{bui2018scene}.}
    \label{fig: structure}
\end{figure}

% image map
Relocalization in a 3D map involves recovering the camera pose of a 2D image with respect to a pre-built 3D scene model. This 3D map is constructed from color images using approaches such as structure-from-motion (SfM) \cite{saputra2018visual} or range images using approaches such as truncated-signed-distance-function (TSDF) \cite{Tzeng2017}. As depicted in Figure \ref{fig: structure}, 3D map based methods establish 2D-3D correspondences between the 2D pixels of a query image and the 3D points using local descriptors \cite{li2010location, li2012worldwide, zeisl2015camera, germain2021neural} or scene coordinate regression \cite{cavallari2017fly, guzman2014multi, shotton2013scene, massiceti2017random}. These 2D-3D matches are then used to compute the camera pose by applying a Perspective-n-Point (PnP) solver \cite{gao2003complete} within a RANSAC loop \cite{ wald2020beyond}.

\subsubsection{Local Descriptor Based Relocalization}
\label{section:Descriptor Based}
Local descriptor based relocalization relies on establishing correspondences between 2D map inputs and the given explicit 3D model using feature descriptors. As the learning of feature descriptor is typically coupled with keypoint detection, existing learning methods can be divided into three types: \textit{detect-then-describe}, \textit{detect-and-describe}, and \textit{describe-then-detect}, according to the role of detector and descriptor in the learning process.

\begin{table*}[t]
\caption{A summary on existing methods on deep learning based relocalization in a 3D map {(Section \ref{sec: 3d relocalization})}}
\label{tb: learning for 2d-3d localization}
    \small
    \begin{center}
\begin{tabular}{c c c l c c c l}
\hline
\multicolumn{3}{c}{\multirow{2}{*}{Model}} & \multirow{2}{*}{Year} & \multirow{2}{*}{Agnostic} & \multicolumn{2}{c}{Performance (m/degree)} & \multicolumn{1}{c}{\multirow{2}{*}{Contributions}} \\
\multicolumn{3}{c}{} & & & 7Scenes & Cambridge & \multicolumn{1}{c}{}\\ \hline
\multirow{32}{*}{\rotatebox{90}{Relocalization in 3D Map}}
& \multirow{20}{*}{\rotatebox{90}{Descriptor Based}}                            
& NetVLAD \cite{arandjelovic2016netvlad} & 2016 & Yes & -  & -  & \footnotesize{differentiable VLAD layer}\\
&& DELF \cite{noh2017large} & 2017 & Yes & - & - & \footnotesize{attentive local feature descriptor}\\
&& InLoc \cite{taira2018inloc} & 2018 & Yes & 0.04/1.38  & 0.31/0.73 & \footnotesize{dense data association} \\
&& SVL \cite{schonberger2018semantic} & 2018 & No  & - & - & \footnotesize{leverage a generative model for descriptor learning} \\ 
&& SuperPoint \cite{detone2018superpoint} & 2018 & Yes & - & - & \footnotesize{jointly extract interest points and descriptors} \\ 
&& Sarlin et al. \cite{sarlin2018leveraging} & 2018 & Yes & - & - & \footnotesize{hierarchical localization} \\ 
&& NC-Net \cite{rocco2018neighbourhood} & 2018 & Yes & - & - & \footnotesize{neighbourhood consensus constraints} \\
&& 2D3D-MatchNet \cite{feng20192d3d} & 2019 & Yes & - & - & \footnotesize{jointly learn the descriptors for 2D and 3D keypoints}\\
%&& Unsuperpoint \cite{christiansen2019unsuperpoint} & 2019 & Yes & - & - & \footnotesize{unsupervised detector and descriptor learning}\\
&& HF-Net \cite{sarlin2019coarse} & 2019 & Yes & 0.042/1.3 & 0.356/0.31 & \footnotesize{coarse-to-fine localization} \\ 
&& D2-Net \cite{dusmanu2019d2} & 2019 & Yes & - & - & \footnotesize{jointly learn keypoints and descriptors} \\
&& Speciale et al \cite{speciale2019privacy} & 2019 & No  & - & - & \footnotesize{privacy preserving localization} \\
&& OOI-Net \cite{weinzaepfel2019visual} & 2019 & No  & - & - & \footnotesize{objects-of-interest annotations} \\
&& Camposeco et al. \cite{camposeco2019hybrid} & 2019 & Yes & - & 0.56/0.66 & \footnotesize{hybrid scene compression for localization}\\
&& Cheng et al. \cite{cheng2019cascaded} & 2019 & Yes & - & - & \footnotesize{cascaded parallel filtering} \\
&& Taira et al. \cite{taira2019right} & 2019 & Yes & - & - & \footnotesize{comprehensive analysis of pose verification} \\
&& R2D2 \cite{revaud2019r2d2} & 2019 & Yes & - & - & \footnotesize{learn a predictor of the descriptor discriminativeness} \\
&& ASLFeat \cite{luo2020aslfeat} & 2020 & Yes & - & - & \footnotesize{leverage deformable convolutional networks}\\
&& CD-VLM \cite{dusmanu2021cross} & 2021 & Yes & - & - & \footnotesize{cross-descriptor matching}\\
&& VS-Net \cite{huang2021vs} & 2021 & No & \textbf{0.024}/\textbf{0.8} & \textbf{0.136}/\textbf{0.24} & \footnotesize{vote by segmentation}\\
\cline{2-7}
& \multirow{12}{*}{\rotatebox{90}{Scene Coordinate Regression}}                 
& DSAC \cite{brachmann2017dsac} & 2017 & No  & 0.20/6.3   & 0.32/0.78 & \footnotesize{differentiable RANSAC} \\
&& DSAC++ \cite{brachmann2018learning} & 2018 & No  & 0.08/2.40  & 0.19/0.50 & \footnotesize{without using a 3D model of the scene}\\
&& Angle DSAC++ \cite{li2018scene} & 2018 & No  & 0.06/1.47  & 0.17/0.50 & \footnotesize{angle-based reprojection loss} \\
&& Dense SCR \cite{li2018full} & 2018 & No  & 0.04/1.4 & -  & \footnotesize{full frame scene coordinate regression}\\
&& Confidence SCR \cite{bui2018scene} & 2018 & No  & 0.06/3.1 & - & \footnotesize{model uncertainty of correspondences}\\
&& ESAC \cite{brachmann2019expert} & 2019 & No  & 0.034/1.50 & - & \footnotesize{integrates DSAC in a Mixture of Experts}\\
&& NG-RANSAC \cite{brachmann2019neural} & 2019 & No & - & 0.24/\textbf{0.30} & \footnotesize{prior-guided model hypothesis search}\\
&& SANet \cite{yang2019sanet} & 2019 & Yes & 0.05/1.68  & 0.23/0.53 & \footnotesize{scene agnostic architecture for camera localization} \\
&& MV-SCR \cite{cai2019camera} & 2019 & No  & 0.05/1.63  & 0.17/0.40 & \footnotesize{multi-view constraints}\\
&& HSC-Net \cite{li2020hscnet} & 2020 & No  & 0.03/0.90  & \textbf{0.13}/\textbf{0.30} & \footnotesize{hierarchical scene coordinate network}\\
&& KFNet \cite{zhou2020kfnet} & 2020 & No  & 0.03/\textbf{0.88}  & \textbf{0.13}/\textbf{0.30} & \footnotesize{extends the problem to the time domain}
\\
&& DSM \cite{tang2021learning} & 2021 & Yes  & \textbf{0.027}/0.92  & 0.27/0.52 & \footnotesize{dense coordinates prediction}
\\
\hline
\end{tabular}
\end{center}
\begin{itemize}
              \footnotesize{
              {\item \textit{Year} indicates the publication year (e.g. the date of conference) of each work.}
                \item \textit{Agnostic} indicates whether it can generalize to new scenarios.
                \item \textit{Performance} reports the position (m) and orientation (degree) error (a small number is better) on the 7-Scenes (Indoor)\cite{shotton2013scene} and Cambridge (Outdoor) dataset\cite{kendall2015posenet}. Both datasets are split into training and testing set. We report the averaged error on the testing set. 
                \item \textit{Contributions} summarize the main contributions of each work compared with previous research.
            }
    \end{itemize}
\end{table*}

 {\textit{Detect-then-describe} is a common pipeline for local descriptor-based relocalization.} This approach first performs feature detection and then extracts a feature descriptor from a patch centered around each keypoint \cite{mikolajczyk2004scale, leutenegger2011brisk}. The keypoint detector is responsible for providing robustness or invariance against possible real issues such as scale transformation, rotation, or viewpoint changes by normalizing the patch accordingly. However, some of these responsibilities might also be delegated to the descriptor. The common pipeline varies from using hand-crafted detectors \cite{bay2006surf, lowe2004distinctive} and descriptors \cite{calonder2010brief, rublee2011orb}, replacing either the descriptor \cite{balntas2016learning, simo2015discriminative, simonyan2014learning, rocco2018neighbourhood, moo2018learning, ebel2019beyond, larsson2019fine, pautrat2020online, wang2020learning, tian2020hynet} or detector \cite{savinov2017quad, zhang2018learning, laguna2019key} with a learned alternative, or learning both the detector and descriptor \cite{ono2018lf, yi2016lift, zhou2020da4ad, lu2020rskdd}. For efficiency, the feature detector often considers only small image regions and typically focuses on low-level structures such as corners or blobs \cite{harris1988combined}, while the descriptor often captures higher level information in a larger patch around the keypoint. 

 In contrast, \textit{detect-and-describe} approaches advance the description stage. By sharing a representation from deep neural network, SuperPoint \cite{detone2018superpoint} and R2D2 \cite{revaud2019r2d2} attempt to learn a dense feature descriptor and a feature detector. However, they rely on different decoder branches which are trained independently with specific losses. On the contrary, D2-net \cite{dusmanu2019d2} and ASLFeat \cite{luo2020aslfeat} share all parameters between detection and description and use a joint formulation that simultaneously optimizes for both tasks. Different from these works, which purely rely on image features, P2-Net \cite{wang2021p2} proposes a unified descriptor between 2D and 3D representations for pixel and point matching. 

Alternatively, the \textit{describe-then-detect} approach, e.g. D2D \cite{tian2020d2d}, postpones the detection to a later stage but applies such detector on pre-learned dense descriptors to extract a sparse set of keypoints and corresponding descriptors. 

In practice, descriptors are commonly used to perform sparse feature extraction and matching for the requirement of efficiency with keypoint detector. Moreover, by disabling the function of keypoint detector, dense feature extraction and matching \cite{choy2016universal, fathy2018hierarchical, savinov2017matching, schonberger2018semantic, taira2018inloc, hyeon2021pose, berton2021viewpoint}, show better matching results than sparse feature matching, particularly under strong variations in illumination \cite{sattler2018benchmarking}.
More recently, new approaches have been proposed to establish correspondence for visual localization. For example, CD-VLM~\cite{dusmanu2021cross} uses cross-descriptor matching to overcome challenges in cross-seasonal and cross-domain visual localization. VS-Net~\cite{huang2021vs} proposes a scene-specific landmark-based approach, which uses a set of keyframe-based landmarks to establish correspondences in visual localization. These new approaches offer promising alternatives for robust and accurate visual localization.

\subsubsection{Scene Coordinate Regression Based Localization}

Different from local descriptor-based relocalization which relies on matching descriptors between images and an explicit 3D map to establish 2D-3D correspondences, scene coordinate regression approaches eliminate the need for explicit 3D map construction and descriptor extraction, making it relatively more efficient. Instead of relying on explicit 3D maps, these methods learn an implicit transformation from 2D pixel coordinates to 3D point coordinates. By estimating the 3D coordinates of each pixel in the query image within the world coordinate system (i.e., the scene coordinates \cite{shotton2013scene, dong2021robust}), these approaches allow for more flexibility in dealing with different environments and scene structures. This makes scene coordinate regression a promising alternative for relocalization tasks, especially in scenarios where explicit 3D maps may not be available or accurate enough.

DSAC \cite{brachmann2017dsac} is a relocalization pipeline that leverages a ConvNet to regress scene coordinates and incorporates a novel differentiable RANSAC algorithm to allow for end-to-end training of the pipeline. This approach has been extended in several ways to improve its performance and applicability. For example, reprojection loss \cite{brachmann2018learning, brachmann2020visual, li2018scene} and multi-view geometric constraints \cite{cai2019camera} have been introduced to enable unsupervised learning and joint learning of observation confidences \cite{bui2018scene, brachmann2019neural} to enhance sampling efficiency and accuracy. Other strategies, such as Mixture of Experts (MoE) \cite{brachmann2019expert} and hierarchical coarse-to-fine \cite{li2020hscnet, wang2021continual}, have been integrated to eliminate environment ambiguities. Different from these, KFNet \cite{zhou2020kfnet} extends the scene coordinate regression problem to the time domain, effectively bridging the performance gap between temporal and one-shot relocalization approaches. However, these methods are still limited to a specific scene and cannot be generalized to unseen scenes without retraining. To address this limitation, SANet \cite{yang2019sanet} regresses the scene coordinate map of the query by interpolating the 3D points associated with the retrieved scene images, making it a scene-agnostic method. Unlike aforementioned methods which are trained in a sparse manner, Dense SCR and DSM \cite{li2018full, tang2021learning} perform scene coordinate regression in a dense manner, making the computation more efficient during testing. Moreover, they incorporate global context into the regression process to improve robustness. Overall, these advances in scene coordinate regression and relocalization techniques offer promising avenues for improving localization accuracy in diverse scenarios.

Scene coordinate regression-based methods can be more efficient than local descriptor-based methods as they eliminate the need for descriptor extraction and matching. These methods can directly regress the corresponding 3D point for a given 2D pixel, thus generating 2D-3D correspondences efficiently. Additionally, implicit 3D map-based relocalization methods have shown promising results, exhibiting robust and accurate performance in small indoor environments and achieving comparable, if not better, performance than explicit 3D map-based methods. It is worth noting, however, that the effectiveness of these implicit methods in large-scale outdoor scenes has not been demonstrated. This is due to their dependence on learning a regression function that maps 2D image coordinates to 3D scene coordinates, which may not generalize well to outdoor scenes with diverse illumination, weather conditions, and scene layouts.

\section{Mapping}
\label{sec: mapping}
Mapping refers to the ability of a mobile agent to perceive and build a consistent environmental model to describe surroundings. 
Deep learning has fostered a set of tools for scene perception and understanding, with applications ranging from depth prediction, object detection, to semantic labelling and 3D geometry reconstruction. This section provides an overview of existing works relevant to deep learning based mapping (scene perception) methods. We categorize them into geometric mapping, semantic mapping, and implicit mapping. 

\subsection{Geometric Mapping}
\begin{figure}
    \centering
    \includegraphics[width=0.48\textwidth]{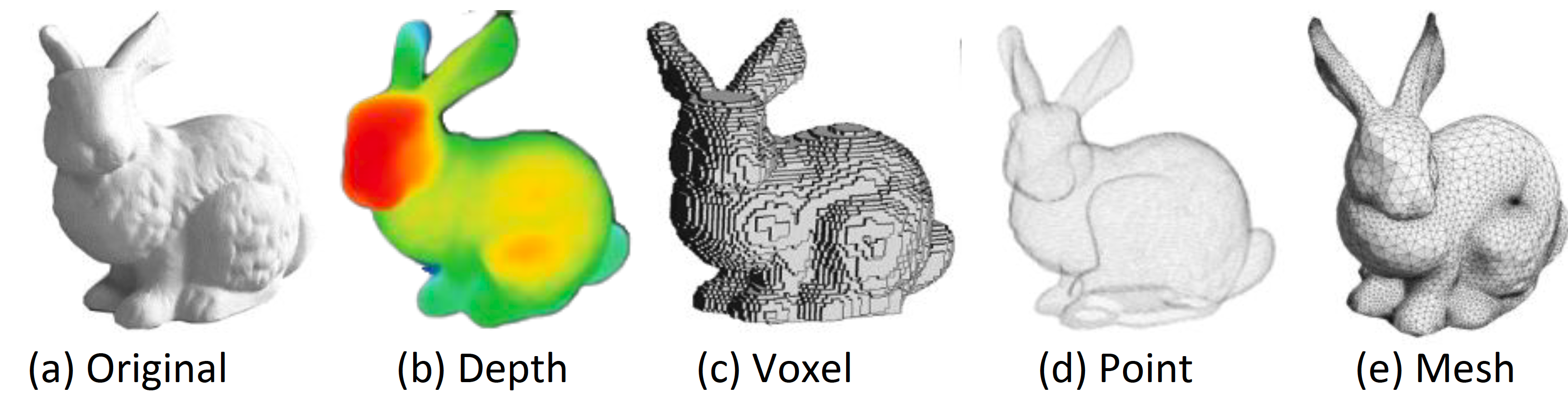}
    \caption{An illustrations of scene representations on the Stanford Bunny benchmark: (a) original model, (b) depth , (c) voxel  (d) point and (e) mesh representation.}
    \label{fig: scene represenation}
    %\vspace{-0.3cm}
\end{figure}

Broadly, geometric mapping captures the shape and structural description of a scene. The classical mapping algorithms can be categorized into sparse features or dense methods. As deep learning based approaches mostly represent scene with dense representations, this section focuses on introducing relevant works in this area. Typical choices of dense scene representations include depth, point, boundary, mesh and voxel. Figure \ref{fig: scene represenation} visualizes these representative geometric representations on the Stanford Bunny benchmark. Inspired by \cite{cadena2016past}, we further divide the learning approaches into two parts: raw dense representations and boundary dense representations.

\subsubsection{{Raw Dense Representations}}
\label{sec: depth}

Conditioned on input images, deep learning approaches are able to generate 2.5D depth maps or 3D points as raw dense representations that express scene geometry in high resolution. Such raw representations  {serve as fundamental components} to constitute a scene that is well-suited to robotic tasks, such as obstacle avoidance.
In SLAM (Simultaneous Localization and Mapping) systems,  {these raw dense mapping methods} are jointly used with motion tracking. For example,
dense scene reconstruction can be achieved by fusing per-pixel depth and RGB images, such as DTAM \cite{Newcombe2011} and \cite{kerl2013dense,whelan2015real}. 

1) 2.5D depth representation:
Learning depth from raw images is a fast evolving area in computer vision community. There are generally three main categories: supervised learning based, self-supervised learning with spatial consistency based and self-supervised learning with temporal consistency based depth estimation.

One of the earliest approaches is \cite{eigen2014depth} that takes a single image as input, and processes to output per-pixel depths.  {It uses two deep neural networks, i.e. one for coarse global prediction and the other for local refinement, and applies scale-invariant error to measure depth relations. This method achieves new state-of-the-art performance on NYU Depth and KITTI datasets.} More accurate depth prediction is achieved by jointly optimizing the depth and self-motion estimation \cite{ummenhofer2017demon}.
 {This work learns to produce depth and camera motion from unconstrained image pairs via ConvNet based encoder-decoder structure and an iterative network that improves predictions. The network estimates surface normals, optical flow, and matching confidence, with a training loss based on spatial relative differences. Compared to traditional depth estimation methods, this approach achieves higher accuracy and robustness, and outperforms single-image-based depth learning network \cite{ummenhofer2017demon} by better generalizing to unseen structures.}
 {\cite{liu2015learning} proposes a ConvNet based neural model to estimate depth from monocular images by using continuous conditional random field (CRF) learning and a structured learning scheme that learns the unary and pairwise potentials of continuous CRF in a unified deep CNN framework. This model improves upon supervised learning based depth estimation and is relatively more efficient.}
While these supervised learning methods have shown superior performance compared to traditional structure-based methods, such as \cite{karsch2014depth}, their effectiveness is limited by the availability of labeled data during model training, making generalization to new scenarios difficult.

On the other side, recent advances in this field focus on unsupervised solutions, by reformulating depth prediction as a novel view synthesis problem. \cite{garg2016unsupervised}
 utilizes photometric consistency loss as a self-supervision signal for training neural models. With stereo images and a known camera baseline, it synthesizes the left view from the right image, and the predicted depth maps of the left view. By minimizing the distance between synthesized images and real images, i.e. the spatial consistency, the parameters of the networks are recovered via this self-supervision in an end-to-end manner.  {Similarly, \cite{godard2017unsupervised} proposes a single image depth estimation model that uses binocular stereo footage instead of ground truth depth data. Their approach utilizes an image reconstruction loss to generate disparity images and enforces consistency between disparities produced relative to both the left and right images to improve performance and robustness, outperforming \cite{liu2015learning} and \cite{garg2016unsupervised}.} 

In addition to spatial consistency, temporal consistency can also be used as a self-supervised signal\cite{Zhou2017}. These approaches synthesize the image in the target time frame from the source time frame, while simultaneously recovering egomotion and depth estimation. Importantly, this framework only requires monocular images to learn both depth maps and egomotion. As we have discussed this part in Section \ref{sec: unsupervised vo}, we refer the readers to Section \ref{sec: unsupervised vo} for more details.

The learned depth information can be integrated into SLAM systems to address some limitations of classical monocular solution. For example, CNN-SLAM \cite{tateno2017cnn} utilizes the learned depths from single images into a monocular SLAM framework (i.e. LSD-SLAM \cite{engel2014lsd}). It shows how learned depth maps contribute to mitigating the absolute scale recovery problem in pose estimates and scene reconstruction. {With the dense depth maps predicted by ConvNets,} CNN-SLAM provides dense scene predictions in texture-less areas, which is normally hard for a conventional SLAM system.

2) 3D Points Representation: 
Deep learning  {techniques have also been} introduced to generate 3D points from raw images.  {The} point-based formulation  {represents} the 3-dimensional coordinates (x, y, z) of points in 3D space. While this formulation is straightforward and easily manipulated, it encounters the challenge of ambiguity, wherein different configurations of point clouds can represent the same underlying geometry.

The pioneer work in this domain is PointNet \cite{qi2017pointnet} that directly operates on point clouds, without the need for unnecessary conversion to regular 3D voxel grids or image collections. PointNet is specifically designed to  {handle} the permutation invariance of points in the input, and its applications span various tasks, such as object classification, part segmentation, and scene semantic parsing.
Furthermore, \cite{fan2017point} develops a deep generative model that can generate 3D geometry in point-based formulation from single images. In their work, a loss function based on Earth Mover's distance is introduced to tackle the problem of data ambiguity. However, their method  {has only} been validated on the reconstruction task of single objects.  {As of now, no research} on point generation for scene reconstruction has been found, primarily due to the large computational burden  {associated with such endeavors}.

\subsubsection{{Boundary and Spatial-Partitioning Representations}}

Beyond unstructured raw dense representations (i.e. 2.5D depth maps and 3D points), boundary representations express the 3D scene with explicit surfaces and spatial-partitioning (i.e. boundaries).

{1) Surface mesh representation:} Mesh-based formulation naturally captures the surface of 3D shape. It encodes the underlying surface structure of 3D models, such as edges, vertices and faces.
Several works consider the problem of learning mesh generation from images \cite{groueix2018papier,wang2018pixel2mesh} or point clouds data \cite{ladicky2017point,dai2019scan2mesh,peng2021shape}. However, these approaches are only able to reconstruct single objects, and limited to generating models with simple structures or from familiar classes. 
To tackle the problem of scene reconstruction in mesh representation, \cite{mukasa20173d} integrates the sparse features from monocular SLAM with the dense depth maps from ConvNets to the update 3D mesh representation. In this work, SLAM-measured sparse features and CNN-predicted dense depth maps are fused to obtain a more accurate 3D reconstruction,  a 3D mesh representation is updated by integrating accurately tracked sparse features points. The proposed work shows a reduction in the mean residual error of 38\% compared to ConvNet-based depth map prediction alone in 3D reconstruction.
To allow efficient computation and flexible information fusion, \cite{bloesch2019learning} utilizes 2.5D mesh to represent scene geometry. In this approach, the image plane coordinates of mesh vertices are learned by deep neural networks, while depth maps are optimized as free variables. A factor graph is utilized to integrate information in a flexible and continuous manner through the use of learnable residuals. Experimental evaluation on synthetic and real data shows the effectiveness and practicability of the proposed approach.

2) Surface function representation: This representation describes the surface as the zero-crossing of an implicit function. A popular choice is signed distance function, a continuous volumetric field, in which the magnitude of a point is the distance to the surface boundary and the sign determines whether it is inside or outside. DeepSDF is proposed learn to generate such a continuous field by a classifier, indicating which boundary is the shape surface \cite{park2019deepsdf}. 
Specifically, DeepSDF is a learned continuous Signed Distance Function (SDF) representation of a class of shapes that enables high quality shape representation, interpolation and completion from partial and noisy 3D input data. It represents a shape's surface by a continuous volumetric field and explicitly represents the classification of space as being part of the shapes interior or not. DeepSDF can represent an entire class of shapes and has impressive performance on learning 3D shape representation and completion while reducing the model size by an order of magnitude compared with previous works.
Another approach, Occupancy Networks generate a continuous 3D occupancy function with deep neural networks, representing the decision boundary with neural classifier\cite{mescheder2019occupancy},  {a description of the 3D output at infinite resolution without excessive memory footprint. The effectiveness of this approach has been validated for 3D reconstruction from single images, noisy point clouds, and coarse discrete voxel grids, and demonstrate competitive results over baselines. To further improve Occupancy Networks, Convolutional Occupancy Networks \cite{peng2020convolutional} combines Convolutional encoders with implicit occupancy decoders. This method is empirically validated through experiments reconstructing complex geometry from noisy point clouds and low-resolution voxel representations.} In addition,
\cite{mildenhall2020nerf} leverages deep fully-connected neural network to optimize a radiance field function to represent a scene. Their experiments demonstrate good performance in novel view synthesis task.
Compared with raw representations, surface function representation reduces storage memory significantly. Different from aforementioned methods that are limited to closed surfaces, NDF~\cite{chibane2020neural} is proposed to predict unsigned distance fields for arbitrary 3D shapes, which is more flexible in practical usages.

3) Voxel representation:
Similar to the usage of pixel (i.e. 2D element) in images, voxel is a volume element in a three-dimensional space. 
Previous works explore to use multiple input views, to reconstruct the volumetric representation of a scene \cite{ji2017surfacenet,paschalidou2018raynet} and objects \cite{kar2017learning}. For example, SurfaceNet \cite{ji2017surfacenet} learns to predict the confidence of a voxel to determine whether it is on surface or not, and reconstruct the 2D surface of a scene. 
SurfaceNet is based on a 3D convolutional network that encodes the camera parameters together with the images in a 3D voxel representation, allowing for the direct learning of both photo-consistency and geometric relations of the surface structure. This framework is evaluated on the large-scale scene reconstruction dataset, demonstrating its effectiveness for multiview stereopsis.
RayNet \cite{paschalidou2018raynet} reconstructs the scene geometry by extracting view-invariant features while imposing geometric constraints. 
It encodes the physics of perspective projection and occlusion via Markov Random Fields while utilizing a ConvNet to learn view-invariant feature representations.
Some works focus on generating high-resolution 3D volumetric models \cite{hane2017hierarchical,tatarchenko2017octree}. For example, \cite{tatarchenko2017octree} designes a convolutional decoder based on octree-based formulation to enable scene reconstruction in much higher resolution. 
This network predicts the structure of the octree and the occupancy values of individual cells, making it valuable for generating complex 3D shapes. Unlike standard decoders with cubic complexity, this architecture allows for higher resolution outputs with limited memory budget.
Others can be found on scene completion from RGB-D data \cite{dai2017shape,riegler2017octnetfusion}.
One limitation of voxel representation is its high computational requirement, especially when attempting to reconstruct a scene in high resolution.

Choosing optimal representation for mapping is still an open question. The choice of scene representation for SLAM depends on a range of factors, including the sensor modality, the level of detail required, the computational resources available, and the size and complexity of the environment. In general, dense representations, such as depth maps or point clouds, offer a comprehensive and detailed view of the scene but incur a high computational and memory cost. This renders them more suitable for small-scale scenes. On the other hand, boundary representations, such as mesh and surface function-based formulations, are preferred for large-scale outdoor environments due to their ability to capture the scene's structure and geometry while keeping memory and computational requirements within feasible limits.

\subsection{Semantic Map}

\begin{figure}
    \centering
    \includegraphics[width=0.48\textwidth]{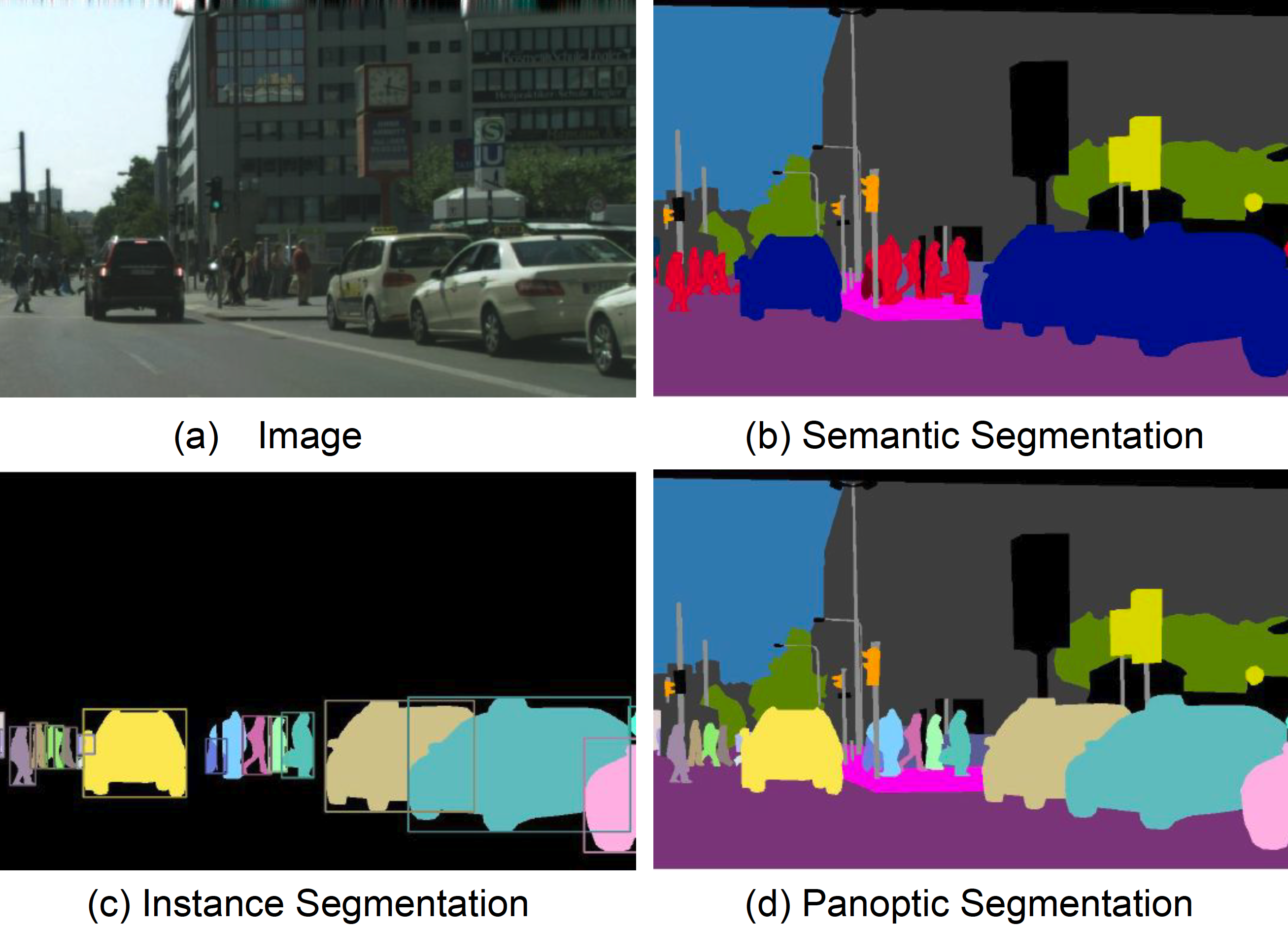}
    \caption{(a) Raw image (b) semantic segmentation, (c) instance segmentation and (d) panoptic segmentation for semantic mapping \cite{kirillov2019panoptic}.}
    \label{fig: semantic mapping}
    %\vspace{-0.3cm}
\end{figure}

Semantic mapping connects semantic concepts (i.e. object classification, material composition etc) with environment geometry. The advances in deep learning greatly foster the developments of object recognition and semantic segmentation. Maps with semantic meanings enable mobile agents to have a high-level understanding of their environments beyond pure geometry, and allow for a greater range of functionality and autonomy. 

SemanticFusion \cite{mccormac2017semanticfusion} is one of the early contributions that combines semantic segmentation labels  {obtained} from deep ConvNet with dense scene geometry  {derived} from a SLAM system. This integration is achieved by probabilistically associating 2D frames with a 3D map, thereby incrementally incorporating per-frame semantic segmentation predictions into the dense 3D map.  {The combined framework} not only generates a map  {enriched} with useful semantic information, but also shows that the integration with a SLAM system  enhances single-frame segmentation. However, in SemanticFusion, the two modules, i.e. semantic segmentation and SLAM, are loosely coupled.
\cite{ma2017multi} proposes a self-supervised network that predicts consistent semantic labels for a map, by imposing constraints on the coherence of semantic predictions  {across across different viewpoints}.
DA-RNN \cite{xiang2017rnn} introduces recurrent models into the semantic segmentation framework, enabling the learning of temporal connections across multiple view frames, producing more accurate and consistent semantic labelling for volumetric maps.  {Another recent work \cite{qin2021light} proposes a framework that builds a compact semantic map using crowd-sourced visual data. Localization is achieved by matching current feature points against the built semantic map via the iterative closest point (ICP) method. Unlike previous approaches that are evaluated on a room level, this work provides a lightweight semantic mapping and localization that performs well in large-scale city scenes.}
Yet it is worth noting that these semantic segmentation-based methods do not provide information about object instances. Therefore, they are unable to distinguish between different objects belonging to the same category.

With the advances in instance segmentation, semantic mapping  has evolved operate at the instance level. A notable example is
\cite{sunderhauf2017meaningful} that offers object-level semantic mapping by employing a bounding box detection module and an unsupervised geometric segmentation module to identify individual objects.
\cite{grinvald2019volumetric} presents a framework that achieves instance-aware semantic mapping, and enables novel object discovery   {within the mapped environment}.
Unlike other dense semantic mapping approaches, Fusion++ \cite{mccormac2018fusion++} builds a semantic graph-based map that specifically predicts object instances and maintains a consistent map via loop closure detection, pose-graph optimization and further refinement. 
 In order to leverage learned object information  {more effectively}, \cite{doherty2019multimodal} presents a probabilistic framework  {within the context of SLAM}. It introduces object detectors as semantic landmarks into a factor graph, enabling the joint optimization of pose estimation, landmark positions/classes and data association. This integration helps address ambiguous data association challenges encountered in the mapping process.

Recently, panoptic segmentation \cite{kirillov2019panoptic} attracts attentions. PanopticFusion \cite{narita2019panopticfusion}  {represents an advancement in semantic mapping that extends to the level of stuff and things classification.} In this context, stuff classes encompass static objects such as walls, doors, and lanes, while things classes include accountable objects like moving vehicles, humans, and tables. Figure \ref{fig: semantic mapping}  {provides a visual comparison} between semantic segmentation, instance segmentation, and panoptic segmentation.

In summary, semantic mapping generates high-level representations of the environment by incorporating information about the semantic meaning, material composition, and geometry of objects, along with the specifics of their instance-level characteristics. These methods employs probabilistic models to gradually fuse per-frame semantic segmentation predictions into a dense 3D map or employs object detectors as semantic landmarks to optimize pose and scene estimation together. They provide a comprehensive understanding of the environment, enabling mobile agents to achieve higher levels of functionality and autonomy.

\subsection{Implicit Map}
In addition to \textit{explicit} geometric and semantic map representations, deep learning models are able to encode the entire scene into an \textit{implicit} representation, known as a neural map. This neural map representation captures the underlying scene geometry and appearance in an implicit manner.

1) Autoencoder based Scene Representation:
Deep autoencoders offer the capability to automatically discover high-level compact representations of high-dimensional image data. A notable example is CodeSLAM \cite{Bloesch2018} that encodes observed images into a compact and optimizable representation to contain the essential information of a dense scene. The learned implicit representation  {is then utilized within} a keyframe-based SLAM system to infer both  {camera poses} and depth maps. The reduced size of learned representation in CodeSLAM  {enables} efficient optimization of camera motion tracking and scene geometry,  {facilitating global consistency in visual localization and mapping}.

2) Neural Rendering based Scene Representation:
Neural rendering models  {form a distinct category of research that leverages view synthesis as a self-supervision signal to implicitly learn and model the 3D structure of a scene}. These models aim to reconstruct a new scene from an unknown viewpoint.

A notable example is the Generative Query Network (GQN) \cite{eslami2018neural} that learns to capture a neural implicit representation and utilizes it to render new scenes. GQN consists of a representation network and a generation network. The representation network encodes observations from reference views into a scene representation, while the generation network, based on a recurrent model, reconstructs the scene from a new view conditioned on the scene representation and a stochastic latent variable.  By taking observed images from multiple viewpoints and the camera pose of a new view as inputs, GQN predicts the physical scene of the new view. Through end-to-end training, the representation network can capture the necessary and important factors of 3D environment for the scene reconstruction task via the generation network. GQN  {has been} extended to incorporate a geometric-aware attention mechanism to allow more complex environment modelling \cite{tobin2019geometry}. Furthermore, the integration of multimodal data for scene inference has been explored to enhance the capabilities of GQN \cite{lim2019neural}.

Recently, NeRF \cite{mildenhall2020nerf} is proposed to explicitly encode the radiance fields of complicated 3D scenes into the weights of MLPs. It delivers an impressive realism for demanding 3D situations by utilizing volume rendering to generate new views for 2D supervision. However, there are three main limitations: 1) because each 3D scene is stored into all MLP weights, the trained network (i.e., a learned radiance field) can only represent a single scene and  is hard to generalize to novel circumstances; 2) as a single camera ray requires tens or even hundreds of the evaluations of the 3D neural scene representation, NeRF-based approaches are highly computational, leading to slow rendering time; 3) due to the fact that each spatial 3D location along a light ray is only optimized by the available pixel RGBs, the learned implicit representations of that site lack the general geometric patterns, resulting in less photo-realistic synthetic images. 
To address these limitations, several works have been proposed, including those that focus on generalization \cite{trevithick2020grf, schwarz2020graf}, efficiency \cite{yu2021plenoctrees, Lindell20arxiv_AutoInt}, and geometry \cite{neff2021donerf, Wang2021d}. NeRF can also be combined with a semantic map, as seen in Semantic-NeRF~\cite{zhi2021place}, which jointly encodes semantics with appearance and geometry, exploiting the intrinsic multi-view consistency and smoothness of NeRF to benefit semantics. 

Additionally, NeRF is also introduced to build SLAM systems, such as iMAP \cite{sucar2021imap} and NICE-SLAM \cite{zhu2022nice}. Specifically, iMAP \cite{sucar2021imap} employs a multilayer perceptron (MLP) as the sole scene representation in a SLAM system, which is trained in live operation without prior data. iMAP designs a keyframe structure, multi-processing computation flow, and dynamic information-guided pixel sampling for speed, achieving tracking at 10 Hz and global map updating at 2 Hz. Compared to standard dense SLAMs, iMAP has efficient geometry representation with automatic detail control and smooth filling-in of unobserved regions. To overcome the limitations of over-smoothed scene reconstructions and difficulty in scaling up to large scenes in SLAM, NICE-SLAM \cite{zhu2022nice} has been proposed as an efficient and robust dense SLAM system. It incorporates multi-level local information through a hierarchical scene representation and is optimized with pre-trained geometric priors, resulting in more detailed reconstruction on large indoor scenes.

3) Reinforcement Learning based Scene Representation:
Last but not least, in the quest of `map-less' navigation, task-driven maps emerge as a novel map representation. This representation is jointly modelled by deep neural networks with respect to the task at hand. Generally those tasks leverage location information, such as navigation or path planning, requiring mobile agents to understand the geometry and semantics of environment. Navigation in unstructured environments (even in a city scale) is formulated as a policy learning problem in these works  \cite{mirowski2016learning,zhu2017target,mirowski2018learning,li2019deep}, and solved by deep reinforcement learning. 
Different from traditional solutions that follow a procedure of building an explicit map, planning path and making decisions, these learning based techniques predict control signals directly from sensor observations in an end-to-end manner, without explicitly modelling the environment. The model parameters are optimized via sparse reward signals, for example, whenever agents reach a destination, a positive reward will be given to tune the neural network.
Once a model is trained, the actions of agents can be determined conditioned on the current observations of environment, i.e. images. In this case, all of environmental factors, such as the geometry, appearance and semantics of a scene, are embedded inside the neurons of a deep neural network and suitable to solving the task at hand.
Interestingly, the visualization of the neurons inside a neural model that is trained on the navigation task via reinforcement learning, has similar patterns as the grid and place cells inside human brain \cite{banino2018vector}. This provides cognitive cues to support the effectiveness of neural map representation. 

\section{Loop-closing and SLAM Back-ends}
\label{sec: back-ends}
Simultaneously tracking self-motion and building environmental structures construct a simultaneous localization and mapping (SLAM) system. The localization and mapping methods discussed in the preceding sections can be  {considered as individual modules within a comprehensive SLAM frameworks}. This section overviews deep learning based loop closure detection and SLAM back-ends.

\subsection{Loop-closure Detection}
Loop-closing (or place recognition) module determines whether a particular location has been visited previously. Upon detecting a loop-closure, global optimization is performed to ensure the overall consistency of motion tracking and the map. For a more comprehensive discussion on this topic, readers are referred to the survey \cite{lowry2015visual}.

Conventional works  {typically} rely on the bag-of-words (BoW) to store and use visual features extracted from hand-designed detectors. However, real-world scenarios often introduce complications such as changes in illumination, weather conditions, viewpoints, and the presence of moving objects. 
To address these challenges, researchers have proposed to use the ConvNet features, that are from pre-trained neural models on large-scale generic image processing dataset. In \cite{sunderhauf2015place}, by adapting object proposal techniques and utilizing convolutional neural network features, potential landmarks within an image can be identified for place recognition. This method does not require any form of training, and the system's components are generic enough to be used off-the-shelf, resulting in performance improvement over current state-of-the-art techniques.
Other representative works, e.g. \cite{gao2017unsupervised,merrill2018lightweight,memon2020loop} are built on a deep auto-encoder structure to extract a compact representation, that compresses scene in an unsupervised manner. 
Specifically, \cite{gao2017unsupervised} utilizes a stacked denoising auto-encoder (SDA) that learns a compressed representation from raw input data in an unsupervised manner, allowing for complex inner structures in image data to be learned without the need for manual visual feature design. 
\cite{merrill2018lightweight} leverages an unsupervised autoencoder architecture, trained with randomized projective transformations to emulate natural viewpoint changes and histogram of oriented gradients (HOG) descriptors for illumination invariance. It is without the need for labeled training data or environment-specific training, and is capable of closing loops in real time with no dimensionality reduction.
\cite{memon2020loop} is based on a super dictionary, which is more memory-efficient than traditional BoW dictionaries. The proposed model uses two deep neural networks to speed up the loop closure detection and to ignore the effect of mobile objects. Experimental results show that it performs robustly and is significantly faster.

Deep learning-based loop closure methods offer more robust and effective visual features by leveraging high-level representations learned from deep neural networks. These approaches have demonstrated improved performance in place recognition and loop-closure detection and can be integrated into SLAM systems.

\subsection{Local Optimization}
When jointly optimizing estimated camera motion and scene geometry, SLAM systems enforce them to satisfy a certain constraint. It is done by minimizing a geometric or photometric loss to ensure their consistency in the local area - the surroundings of camera poses. This is bundle adjustment (BA) problem \cite{triggs1999bundle}. 
Learning based approaches predict depth maps and ego-motion through two individual networks trained above large datasets \cite{Zhou2017}. During the testing procedure when deployed online, there is a requirement that enforces the predictions to satisfy some local constraints. To enable local optimization, traditionally, the second-order solvers, e.g. Gauss-Newton (GN) method or Levenberg-Marquadt (LM) algorithm \cite{nocedal2006numerical}, are applied to optimize motion transformations and per-pixel depth maps.

To this end, LS-Net \cite{clark2018learning} tackles this problem via a learning based optimizer by integrating analytical solvers into its learning process. It learns a data-driven prior, followed by refining neural network predictions with an analytical optimizer to ensure photometric consistency.
 {It can optimize sum-of-squares objective functions in SLAM algorithms, which are often difficult to optimize due to violated assumptions and ill-posed problems.}
BA-Net \cite{tang2019ba} integrates a differentiable second-order optimizer (LM algorithm) into a deep neural network for an end-to-end learning. Instead of minimizing geometric or photometric error, BA-Net is performed on feature space to optimize the consistency loss of features from multiview images extracted by ConvNets. 
This feature-level optimizer can mitigate the fundamental problems of geometric or photometric solution (e.g. some information may be lost in the geometric optimization, while environmental dynamics and lighting changes may impact the photometric optimization).  {This work combines domain knowledge of SLAM with deep learning and achieves successful results on large-scale real data, outperforming conventional SLAM with geometric or photometric BA and
 deep learning based methods, e.g. Zhou et al.\cite{Zhou2017}.}

These learning based optimizers provide an alternative to solve local bundle adjustment problem.  {By integrating analytical solvers and differentiable second-order optimizers into their learning processes, these methods have demonstrated the potential to improve SLAM performance by mitigating challenges such as violated assumptions and ill-posed problems or information loss during optimization. Consequently, they are able to offer promising results for enhancing the accuracy and robustness of local optimization in SLAM systems.}

\subsection{Global Optimization}
Incremental motion estimation  {(visual odometry)} suffers from accumulative error drifts during long-term operation.  {This issue stems from the inherent problem} of path integration, where the system's errors progressively accumulate without effective constraints. To address this challenge, graph-SLAM \cite{grisetti2010tutorial} constructs a topological graph to represent camera poses or scene features as graph nodes, which are connected by edges (measured by sensors) to constrain the poses. This graph-based formulation can be optimized to ensure the global consistency of graph nodes and edges, mitigating the possible errors on pose estimates and the inherent sensor measurement noise. A popular solver for global optimization is through Levenberg-Marquardt (LM) algorithm.

In the era of deep learning, deep neural networks excel at extracting features, and constructing functions from observations to poses and scene representations. A global optimization upon the DNN predictions is necessary to reducing the drifts of global trajectories and supporting large-scale mapping. Compared with a variety of well-researched solutions in classical SLAM, optimizing deep predictions globally is underexplored.

Various studies have explored the integration of learning modules into classical SLAM systems at different levels. At the front-end, deep neural networks (DNNs) generate predictions, which are then incorporated into the back-end for optimization and refinement. One good example is CNN-SLAM \cite{tateno2017cnn}, which uses learned per-pixel depths to support loop closing and graph optimization in LSD-SLAM, a complete SLAM system \cite{engel2014lsd}. The joint optimization of camera poses, scene representations, and depth maps in CNN-SLAM produces consistent scale metrics. This method has been evaluated for estimating the absolute scale of the reconstruction and fusing semantic labels, which results in semantically coherent scene reconstruction from a single view. CNN-SLAM is capable of producing pose and depth estimates consistently in low-textured areas where traditional SLAM systems tend to fail by utilizing depth predictions from neural networks.
In DeepTAM \cite{zhou2020deeptam}, the depth and pose predictions from deep neural networks are integrated into a classical DTAM system\cite{Newcombe2011}, where the system estimates small pose increments and accumulates information in a cost volume to update the depth prediction. Depth measurements and image-based priors are combined for optimization, which results in more accurate scene reconstruction and camera motion tracking. Few images are required, and the system is robust to noisy camera poses. Similarly, in \cite{li2019pose}, unsupervised learning-based VO is combined with a graph optimization back-end. This method generates a windowed pose graph consisting of multi-view constraints and uses a novel pose cycle consistency loss to improve performance and robustness.
Conversely, DeepFactors \cite{czarnowski2020deepfactors} integrates the learned optimizable scene representation (their so-called code representation) into a probabilistic factor graph-based back-end for global optimization. The advantage of the factor-graph-based formulation is its flexibility to include sensor measurements, state estimates, and constraints. It is comparably easy and convenient to add new sensor modalities, pairwise constraints, and system states into the graph for optimization.

In summary, these methods integrate deep neural networks with SLAM back-ends, resulting in numerous benefits, such as improved accuracy and robustness in scene reconstruction and camera motion tracking, handling of low-textured areas, and flexibility in the factor-graph-based formulation for adding new sensor modalities, pairwise constraints, and system states for optimization. It is worth noting, however, that the back-end optimizers employed in these methods are not entirely differentiable at present.

\section{Uncertainty Estimation}
Safety and interpretability are an critical step towards the practical deployment of mobile agents in real-world applications: the former enables robots to live and act with human reliably, while the latter allows users to have better understanding over model behaviours. Although deep learning models achieve impressive performance {in} many visual regression and classification tasks, once failure cases occur, errors from one component inevitably propagate to other downstream modules, causing catastrophic consequences. 
To this end, there is an emerging need to estimate the uncertainty of DNN predictions. This section introduces deep learning approaches to estimating uncertainty for localization and mapping, i.e. {to} capture the uncertainty with the purpose of motion tracking or scene understanding. The estimated uncertainty plays a vital role in probabilistic sensor fusion or the back-end optimization of SLAM systems. 

Deep learning models normally produce the mean values of target predictions, for example, the output of a DNN-based visual odometry model is a 6-dimensional pose vector. 
In order to capture the uncertainty of deep predictions, deep learning models can be augmented into a Bayesian model \cite{gal2016dropout,kendall2017uncertainties}. The {Bayesian uncertainties} are broadly categorized into Aleatoric and epistemic uncertainty: Aleatoric uncertainty reflects observation noises, e.g. sensor measurement or motion noises; epistemic uncertainty captures the belief in model parameters \cite{kendall2017uncertainties}. Bayesian models have been applied to solve global localization problem. As illustrated in \cite{kendall2016modelling,Clark2017}, the uncertainty from deep models are able to reflect the global location errors, in which the unreliable pose estimates are avoided with this belief metric. Estimating the uncertainty of DNN-based incremental motion estimation has been explored by \cite{wang2018end}.
{It} adopts a strategy to convert target predictions into a Gaussian distribution, conditioned on the mean value of pose estimates and its covariance. The parameters inside the framework are optimized via a loss function with a combination of mean and covariance. By minimizing the error function to find the best combination, the uncertainty of motion transformation is automatically recovered in an unsupervised fashion.
\cite{wang2018end} integrates the learned uncertainty into a graph based SLAM as the covariances of odometry edges. It validates that the learned uncertainty further improves the performance of a SLAM system over the baseline with a fixed predefined value of covariance. 

The uncertainty for scene understanding also contributes to SLAM systems. The scene uncertainty offers a belief metric in to what extent the environmental perception and scene structure should be trusted. For example, in the semantic segmentation and depth estimation tasks, uncertainty estimation provides per-pixel uncertainties for the DNN predictions of semantics and depth maps \cite{kendall2017bayesian,kendall2017uncertainties,klodt2018supervising}. 
Further more, scene uncertainty is applicable to building a hybrid SLAM system. For example, photometric uncertainty can be learned to capture the variance of intensity on each image pixel, and hence enhances the robustness of a SLAM system towards observation noises \cite{yang2020d3vo}.

\section{Sensor Fusion}
\label{sec: sensor fusion}
Sensor fusion  {stands as a fundamental challenge in the field of robotics}. In this section, we discuss learning based sensor fusion strategy for motion tracking. We focus on visual-inertial sensor fusion as a representative example. Integrating visual and inertial data as visual-inertial odometry (VIO) is a well-defined problem.
Accurate estimation of pose heavily relies on the effective fusion of measurements from these two complementary sensors. In recent years, data-driven approaches have emerged to directly learn 6-DoF poses from visual and inertial measurements. 

\subsection{Supervised learning based VIO}
VINet \cite{Clark2017a} treats visual-inertial odometry as a sequential learning problem, and proposes an end-to-end DNN framework to solve this problem. VINet uses a ConvNet based visual encoder to extract visual features from two consecutive RGB images, and an inertial encoder to extract inertial features from a sequence of IMU data using a long short-term memory (LSTM) network. The deep sensor fusion is achieved by concatenating visual and inertial features and passing them through an LSTM module to predict relative poses. While this approach is more robust to calibration and timing offset errors, it has not addressed the issue of learning a meaningful sensor fusion strategy. To tackle the deep sensor fusion problem,  {\cite{chen2019selective,chen2022learning} propose} selective sensor fusion, that selectively learns context-dependent representations for visual inertial pose estimation. Their intuition is that the importance of features from different modalities should be considered according to the exterior (i.e., environmental) and interior (i.e., device/sensor) dynamics, by fully exploiting the complementary behaviors of two sensors.
Their approach outperforms VINet and other models without a fusion strategy, avoiding catastrophic failures.

\subsection{Self-supervised learning based VIO}
Learning visual and inertial sensor fusion can be tackled in a self-supervised manner using novel view synthesis. VIOLearner \cite{shamwell2019unsupervised} constructs motion transformations from raw inertial data and converts source images into target images with the camera matrix and depth maps. An online error correction module is introduced to correct intermediate errors, and the network parameters are optimized using a photometric loss. Similarly, DeepVIO \cite{han2019deepvio} uses an unsupervised learning framework to incorporate inertial data and stereo images, training with a dedicated loss to reconstruct trajectories on a global scale.
 {A recent unsupervised visual-inertial odometry framework, UnVIO \cite{wei2021unsupervised}, predicts per-frame depth maps and self-adaptively fuses visual-inertial motion features for pose estimation. To overcome error accumulation and scale ambiguity issues, UnVIO introduces a sliding window optimization strategy. Thanks to this strategy, UnVIO outperforms both DeepVIO \cite{han2019deepvio} and VIOLearner \cite{shamwell2019unsupervised}.}

 Overall, though current learning-based VIO models are not able to surpass the state-of-the-art classical model-based VIOs, the experiments conducted in the works discussed in the previous paragraph indicate that they offer greater robustness against real-world issues such as measurement noises and bad time synchronization. This improved performance can be attributed to the ability of DNNs to extract useful features, learn implicit multimodal fusion, and accurately model motion, providing a significant advantage over classical models.

\section{Discussions}
This survey comprehensively overviews the area of deep learning for visual localization and mapping, and provides a taxonomy to cover the relevant existing approaches from robotics, computer vision and machine learning communities. The fast development of deep learning provides an alternative to solve this problem in a data-driven way, and meanwhile paves the road towards the next-generation AI based spatial perception solution. 

Both localization and mapping modules are critical components of SLAM systems and can function independently or jointly. In instances where our discussion pertains to either localization or mapping separately, we will explicitly state "localization" or "mapping". Conversely, if the benefits or limitations apply to both localization and mapping, we will use "SLAM" or "localization and mapping". The two questions posted at the beginning of this article are visited here, and the limitations of current learning based approaches are summarized as follows.

\textbf{\textit{1) Is deep learning promising to visual localization and mapping?}}

  \begin{table*}
  		\caption{ {A summary of how deep learning can be applied to tackle localization and mapping}}
  		\label{tab: how to use}
  		\small
  		\centering
  		\begin{tabular}{l l}
  		\hline
    	 {How to apply deep learning to solve localization and mapping}	 &  {Methods}  \\ \hline
   	 {Deep learning is used as a universal approximator} &  {\cite{Wang2017,zhao2018learning,xue2019beyond,saputra2019learning,saputra2019distilling}} \\
         {Deep learning is applied to solve the association problem} &  {\cite{balntas2018relocnet,mccormac2017semanticfusion,ma2017multi,sunderhauf2015place}} \\
         {Deep learning can automatically discover relevant features}  &  {\cite{tang2019ba,chen2019selective,mirowski2016learning,zhu2017target,mirowski2018learning}} \\
         {Self-learning framework can be set up to automatically update parameters}  &  {\cite{Zhou2017,bian2019unsupervised,li2018undeepvo,Yin2018,Zhan2018,yang2018deep,zhao2018learning,almalioglu2019ganvo,li2019pose,li2019sequential,sheng2019unsupervised}}  \\
         {Deep learning can tackle some intrinsic problems of conventional algorithms}  &  {\cite{tateno2017cnn,barnes2018driven,yang2018deep,loo2019cnn,yang2020d3vo}} \\
        \hline
 		\end{tabular}
	\end{table*}

SLAM systems have progressed fast over the past decades and shown great successes in real-world deployment. Examples can be witnessed from delivery robots to mobile and wearable devices. 
Admittedly, predominant SLAM systems without embracing deep learning has already meet many needs in certain conditions by exploiting physical laws or geometry heuristics to build up models and algorithms.
Nevertheless, the final answer to the promise of deep learning for SLAM depends on application scenarios from a general view. We believe that three particular properties listed below could make deep learning a unique direction towards a general-purpose SLAM system in the future:

(1) First, deep learning offers powerful perception tools that can be integrated into the visual SLAM front-end to extract features in challenging areas  {for odometry estimation or relocalization}, and provide dense depth \cite{eigen2014depth,Zhou2017}, and semantic labelling \cite{mccormac2017semanticfusion,ma2017multi}  {for mapping}.
Deep learning has been largely embraced by the computer vision community, leading to state-of-the-art methods in a number of computer vision tasks, e.g. object detection, image recognition and semantic segmentation. Some works have already introduced learning algorithms as a 'black box' module to solve important and useful perception problems for SLAM \cite{zhou2020deeptam,yang2020d3vo}. 

(2) Second, deep learning enables high-level understanding and interaction for robots. Neural networks are known to be powerful in connecting abstract elements with human understandable terms\cite{mccormac2017semanticfusion,ma2017multi}, such as labelling scene semantics in  {a mapping or SLAM system}, which is normally hard to describe in a formal mathematical way. Deep-learning enabled scene understanding, on the other hand, is able to support high-level robotic tasks, for example, a service robot searches for an apple in kitchen room by leveraging fine-grained indoor semantics. 

(3) Third, learning methods allow SLAM systems  {or individual localization/mapping algorithms} to learn from past experience, and actively exploit new information for self-learning and adapting to new environment. 
Beyond performing in restricted areas, future SLAM systems are believed to undertake more indispensable roles in unseen scenarios, e.g. nuclear waste disposal. 
By leveraging self-supervised learning \cite{Zhou2017}, or reinforcement learning \cite{mirowski2016learning,zhu2017target,mirowski2018learning}, it would offer opportunities to self-update system (neural network) parameters, and be promising to enhancing the adaptation ability of mobile agents to unseen scenarios without human intervention. 

\textbf{\textit{2) How can deep learning be applied to solve visual localization and mapping?}}

After reviewing the existing works in above sections, predominant methods of adopting deep learning in SLAM systems can be summarized  {in Table \ref{tab: how to use}}.

(1) Deep learning is used as a universal approximator to describe certain functions of SLAM  {or individual localization/mapping algorithms}. 
For example, visual odometry can be achieved by building an end-to-end deep neural network model to directly approximate the function from images to pose \cite{Wang2017,zhao2018learning,xue2019beyond,saputra2019learning,saputra2019distilling}. The advantage here is that the learned models can be inherently incorporated and resilient to certain circumstances, e.g. featureless areas, dynamic lightning conditions and motion blur which are typically difficult to model.

(2) Deep learning is applied to solve the association problem in SLAM. Relocalization needs to connect an image with a pre-built map, and retrieves its pose\cite{balntas2018relocnet}. Semantic  {mapping or} SLAM needs to tackle the complex semantics labelling that associates pixels with its semantic meaning \cite{mccormac2017semanticfusion,ma2017multi}. Loop-closure detection requires to recognize whether observed scene is relevant to the place visited previously \cite{sunderhauf2015place}. 

(3) Deep learning is leveraged to automatically discover features relevant to the task of interest. For example, features suitable to bundle adjustment are extracted to SLAM, showing performance improvement \cite{tang2019ba}. In \cite{chen2019selective}, features relevant to sensor fusion are extracted for visual-inertial odometry. Reinforcement learning based navigation also utilizes the discovered features to constitute an implicit map for path planning and task-driven navigation \cite{mirowski2016learning,zhu2017target,mirowski2018learning}.

(4) By exploiting prior knowledge, e.g. the geometry constraints, a self-learning framework can be set up for SLAM to automatically update parameters based on input images. For instance, novel view synthesis can serve as a self-supervision signal to recover self-motion and depth from unlabelled videos \cite{Zhou2017,bian2019unsupervised,li2018undeepvo,Yin2018,Zhan2018,yang2018deep,zhao2018learning,almalioglu2019ganvo,li2019pose,li2019sequential,sheng2019unsupervised},  {thereby supporting localization tasks.}
 
(5) Deep learning can be utilized to tackle some intrinsic problems of conventional SLAM  {or localization/mapping algorithms}. For instance, the scale-ambiguity problem of monocular SLAM is mitigated by using learned depth estimates with absolute scale from deep neural networks \cite{tateno2017cnn,barnes2018driven,yang2018deep,loo2019cnn}.  Furthermore, the photometric uncertainties of scenes produced by deep neural networks can be introduced into visual odometry (VO) in order to encourage the framework to leverage features that can be trusted and thus further enhance pose estimation performance \cite{yang2020d3vo}.

\textbf{\textit{3) The limitations and Future Directions}}

It must also be pointed out that these learning techniques are reliant on massive datasets, preferably with accurate labels, to extract statistically meaningful patterns and may struggle to generalize to out-of-set environments. Futhermore, there is a lack of sufficiency interpretability of model behaviors as deep learning to date is still a black box model. Additionally, although highly parallelizable, deep learning based  {localization and mapping} %SLAM 
systems are also typically more computationally costly than simpler conventional models if model compression techniques are not used. We further discuss the limitations of existing works as follows:

(1) Real-world deployment. Deploying deep learning models in real-world environments is a systematic research problem. 
In existing research, the prediction accuracy is always chosen as the `rule of thumb' to follow, while other crucial issues are overlooked, such as the optimality of model structure and sizes. The computational and energy consumption have to be considered on resource-constrained systems, e.g., miniaturized insect robots or VR/AR/MR devices. The parallerization opportunities, such as convolutional filters or other parallel neural network modules should be further investigated in order to fully exploit the potential of GPUs. 

(2) Scalability. Deep learning based localization and mapping models have now achieved promising results on the evaluation benchmark. However, they are restricted to some scenarios. For example, odometry estimation  {or localization} is always evaluated in the city area or on the road. Whether these techniques could be applied to other environments, e.g. rural area or forest area remains as an open problem. Moreover, existing works on  {mapping or} scene reconstruction are restricted to single-objects, synthetic data or only at a room level. It is worth exploring the opportunity to scale these methods to more complex, large-scale and realistic problems.

 (3) Safety, reliability and interpretability. Safety and reliability are critical to many applications in practice, e.g. autonomous driving, surgical robots and delivery robots. In these scenarios, even a small error of pose or scene estimates will cause disasters to the entire system. Deep neural networks have been long-critisized as 'black-box', exacerbating the safety concerns for critical tasks. Some initial efforts explore the interpretability on deep models. For example, uncertainty estimation \cite{gal2016dropout,kendall2017uncertainties} offers a belief metric, representing to what extent we trust our models. In this way, unreliable predictions (with low uncertainty) should be avoided or alleviated in order to ensure system's safety and reliability.
 
 (4) Trade-Offs. Although deep learning shows impressive results in solving localization and mapping tasks, there are trade-offs came with them. 
 The metric to evaluate these learning methods is still dominated by the accuracy which can not comprehensively examine the performance of a method. Most works only focus on improving the prediction accuracy. Sometimes a larger DNN model that requires more computation and storage contributes to a higher prediction accuracy, but the hardware and system constraints in practice are rarely considered, such as model efficiency, memory storage and I/O bandwidth. This problem is increasingly pronounced with  {localization and mapping systems} becoming increasingly adopted on resource-constrained platforms, e.g., a drone or an AR headset. In addition, it is hard to choose the optimal parameters e.g. the size of hidden states or the number of layers, inside a DNN model. Another trade-off is between training a model to be overfitted in one domain or to generalize in new domains. Although a model performs well on test data, its performance could be degraded when deployed in application domains that are different from training sets. 

{\small
\bibliographystyle{ieeetr}
\bibliography{reference}

\begin{thebibliography}{100}

\bibitem{Sunderhauf2018}
N.~S{\"{u}}nderhauf, O.~Brock, W.~Scheirer, R.~Hadsell, D.~Fox, J.~Leitner,
  B.~Upcroft, P.~Abbeel, W.~Burgard, M.~Milford, and P.~Corke, ``{The Limits
  and Potentials of Deep Learning for Robotics},'' {\em International Journal
  of Robotics Research}, vol.~37, no.~4-5, 2018.

\bibitem{Forster2014}
C.~Forster, M.~Pizzoli, and D.~Scaramuzza, ``{SVO: Fast Semi-Direct Monocular
  Visual Odometry},'' in {\em The IEEE International Conference on Robotics and
  Automation (ICRA)}, pp.~15--22, 2014.

\bibitem{Qin2018}
T.~Qin, P.~Li, and S.~Shen, ``{VINS-Mono: A Robust and Versatile Monocular
  Visual-Inertial State Estimator},'' {\em IEEE Transactions on Robotics},
  vol.~34, no.~4, pp.~1004--1020, 2018.

\bibitem{sattler2011fast}
T.~Sattler, B.~Leibe, and L.~Kobbelt, ``Fast image-based localization using
  direct 2d-to-3d matching,'' in {\em The International Conference on Computer
  Vision (ICCV)}, pp.~667--674, IEEE, 2011.

\bibitem{lowry2015visual}
S.~Lowry, N.~S{\"u}nderhauf, P.~Newman, J.~J. Leonard, D.~Cox, P.~Corke, and
  M.~J. Milford, ``Visual place recognition: A survey,'' {\em IEEE Transactions
  on Robotics}, vol.~32, no.~1, pp.~1--19, 2015.

\bibitem{Montiel2015}
R.~Mur-Artal, J.~Montiel, and J.~D. Tardos, ``{ORB-SLAM : A Versatile and
  Accurate Monocular SLAM System},'' {\em IEEE Transactions on Robotics},
  vol.~31, no.~5, pp.~1147--1163, 2015.

\bibitem{thrun2005probabilistic}
S.~Thrun, W.~Burgard, and D.~Fox, {\em Probabilistic robotics}.
\newblock MIT press, 2005.

\bibitem{durrant2006simultaneous}
H.~Durrant-Whyte and T.~Bailey, ``Simultaneous localization and mapping: part
  i,'' {\em IEEE robotics \& automation magazine}, vol.~13, no.~2, pp.~99--110,
  2006.

\bibitem{grisetti2010tutorial}
G.~Grisetti, R.~Kummerle, C.~Stachniss, and W.~Burgard, ``A tutorial on
  graph-based slam,'' {\em IEEE Intelligent Transportation Systems Magazine},
  vol.~2, no.~4, pp.~31--43, 2010.

\bibitem{scaramuzza2011visual}
D.~Scaramuzza and F.~Fraundorfer, ``Visual odometry [tutorial],'' {\em IEEE
  robotics \& automation magazine}, vol.~18, no.~4, pp.~80--92, 2011.

\bibitem{cadena2016past}
C.~Cadena, L.~Carlone, H.~Carrillo, Y.~Latif, D.~Scaramuzza, J.~Neira, I.~Reid,
  and J.~J. Leonard, ``Past, present, and future of simultaneous localization
  and mapping: Toward the robust-perception age,'' {\em IEEE Transactions on
  robotics}, vol.~32, no.~6, pp.~1309--1332, 2016.

\bibitem{saputra2018visual}
M.~R.~U. Saputra, A.~Markham, and N.~Trigoni, ``Visual slam and structure from
  motion in dynamic environments: A survey,'' {\em ACM Computing Surveys
  (CSUR)}, vol.~51, no.~2, pp.~1--36, 2018.

\bibitem{tang2022perception}
Y.~Tang, C.~Zhao, J.~Wang, C.~Zhang, Q.~Sun, W.~X. Zheng, W.~Du, F.~Qian, and
  J.~Kurths, ``Perception and navigation in autonomous systems in the era of
  learning: A survey,'' {\em IEEE Transactions on Neural Networks and Learning
  Systems}, 2022.

\bibitem{bailey2006simultaneous}
T.~Bailey and H.~Durrant-Whyte, ``Simultaneous localization and mapping (slam):
  Part ii,'' {\em IEEE robotics \& automation magazine}, vol.~13, no.~3,
  pp.~108--117, 2006.

\bibitem{Wang2017}
S.~Wang, R.~Clark, H.~Wen, and N.~Trigoni, ``{DeepVO : Towards End-to-End
  Visual Odometry with Deep Recurrent Convolutional Neural Networks},'' in {\em
  The IEEE International Conference on Robotics and Automation (ICRA)}, 2017.

\bibitem{Zhou2017}
T.~Zhou, M.~Brown, N.~Snavely, and D.~G. Lowe, ``{Unsupervised Learning of
  Depth and Ego-Motion from Video},'' in {\em IEEE/CVF International Conference
  on Computer Vision and Pattern Recognition (CVPR)}, 2017.

\bibitem{yang2020d3vo}
N.~Yang, L.~von Stumberg, R.~Wang, and D.~Cremers, ``D3vo: Deep depth, deep
  pose and deep uncertainty for monocular visual odometry,'' {\em IEEE/CVF
  International Conference on Computer Vision and Pattern Recognition (CVPR)},
  2020.

\bibitem{Konda2015}
K.~Konda and R.~Memisevic, ``{Learning Visual Odometry with a Convolutional
  Network},'' in {\em International Conference on Computer Vision Theory and
  Applications}, pp.~486--490, 2015.

\bibitem{costante2015exploring}
G.~Costante, M.~Mancini, P.~Valigi, and T.~A. Ciarfuglia, ``Exploring
  representation learning with cnns for frame-to-frame ego-motion estimation,''
  {\em IEEE robotics and automation letters}, vol.~1, no.~1, pp.~18--25, 2015.

\bibitem{geiger2011stereoscan}
A.~Geiger, J.~Ziegler, and C.~Stiller, ``Stereoscan: Dense 3d reconstruction in
  real-time,'' in {\em 2011 IEEE Intelligent Vehicles Symposium (IV)},
  pp.~963--968, Ieee, 2011.

\bibitem{Fischer2015}
P.~Fischer, E.~Ilg, H.~Philip, C.~Hazırbas, P.~V.~D. Smagt, D.~Cremers, and
  T.~Brox, ``{FlowNet: Learning Optical Flow with Convolutional Networks},'' in
  {\em The International Conference on Computer Vision (ICCV)}, 2015.

\bibitem{Geiger2013}
A.~Geiger, P.~Lenz, C.~Stiller, and R.~Urtasun, ``{Vision meets robotics: The
  KITTI dataset},'' {\em The International Journal of Robotics Research},
  vol.~32, no.~11, pp.~1231--1237, 2013.

\bibitem{zhao2018learning}
C.~Zhao, L.~Sun, P.~Purkait, T.~Duckett, and R.~Stolkin, ``Learning monocular
  visual odometry with dense 3d mapping from dense 3d flow,'' in {\em The
  IEEE/RSJ International Conference on Intelligent Robots and Systems (IROS)},
  pp.~6864--6871, IEEE, 2018.

\bibitem{saputra2019learning}
M.~R.~U. Saputra, P.~P. de~Gusmao, S.~Wang, A.~Markham, and N.~Trigoni,
  ``Learning monocular visual odometry through geometry-aware curriculum
  learning,'' in {\em The IEEE International Conference on Robotics and
  Automation (ICRA)}, pp.~3549--3555, IEEE, 2019.

\bibitem{xue2019beyond}
F.~Xue, X.~Wang, S.~Li, Q.~Wang, J.~Wang, and H.~Zha, ``Beyond tracking:
  Selecting memory and refining poses for deep visual odometry,'' in {\em
  IEEE/CVF International Conference on Computer Vision and Pattern Recognition
  (CVPR)}, pp.~8575--8583, 2019.

\bibitem{saputra2019distilling}
M.~R.~U. Saputra, P.~P. de~Gusmao, Y.~Almalioglu, A.~Markham, and N.~Trigoni,
  ``Distilling knowledge from a deep pose regressor network,'' in {\em The
  International Conference on Computer Vision (ICCV)}, pp.~263--272, 2019.

\bibitem{koumis2019estimating}
A.~S. Koumis, J.~A. Preiss, and G.~S. Sukhatme, ``Estimating metric scale
  visual odometry from videos using 3d convolutional networks,'' in {\em The
  IEEE/RSJ International Conference on Intelligent Robots and Systems (IROS)},
  pp.~265--272, IEEE, 2019.

\bibitem{KuoLLLCL20}
X.~Kuo, C.~Liu, K.~Lin, E.~Luo, Y.~Chen, and C.~Lee, ``Dynamic attention-based
  visual odometry,'' in {\em The IEEE/RSJ International Conference on
  Intelligent Robots and Systems (IROS)}, pp.~5753--5760, {IEEE}, 2020.

\bibitem{li2018undeepvo}
R.~Li, S.~Wang, Z.~Long, and D.~Gu, ``Undeepvo: Monocular visual odometry
  through unsupervised deep learning,'' in {\em The IEEE International
  Conference on Robotics and Automation (ICRA)}, pp.~7286--7291, IEEE, 2018.

\bibitem{Yin2018}
Z.~Yin and J.~Shi, ``{GeoNet: Unsupervised Learning of Dense Depth, Optical
  Flow and Camera Pose},'' in {\em IEEE/CVF International Conference on
  Computer Vision and Pattern Recognition (CVPR)}, 2018.

\bibitem{Zhan2018}
H.~Zhan, R.~Garg, C.~S. Weerasekera, K.~Li, H.~Agarwal, and I.~Reid,
  ``{Unsupervised Learning of Monocular Depth Estimation and Visual Odometry
  with Deep Feature Reconstruction},'' in {\em IEEE/CVF International
  Conference on Computer Vision and Pattern Recognition (CVPR)}, pp.~340--349,
  2018.

\bibitem{casser2019depth}
V.~Casser, S.~Pirk, R.~Mahjourian, and A.~Angelova, ``Depth prediction without
  the sensors: Leveraging structure for unsupervised learning from monocular
  videos,'' in {\em The Conference on Artificial Intelligence (AAAI)}, vol.~33,
  pp.~8001--8008, 2019.

\bibitem{almalioglu2019ganvo}
Y.~Almalioglu, M.~R.~U. Saputra, P.~P. de~Gusmao, A.~Markham, and N.~Trigoni,
  ``Ganvo: Unsupervised deep monocular visual odometry and depth estimation
  with generative adversarial networks,'' in {\em The IEEE International
  Conference on Robotics and Automation (ICRA)}, pp.~5474--5480, IEEE, 2019.

\bibitem{wang2019recurrent}
R.~Wang, S.~M. Pizer, and J.-M. Frahm, ``Recurrent neural network for (un-)
  supervised learning of monocular video visual odometry and depth,'' in {\em
  IEEE/CVF International Conference on Computer Vision and Pattern Recognition
  (CVPR)}, pp.~5555--5564, 2019.

\bibitem{li2019pose}
Y.~Li, Y.~Ushiku, and T.~Harada, ``Pose graph optimization for unsupervised
  monocular visual odometry,'' in {\em The IEEE International Conference on
  Robotics and Automation (ICRA)}, pp.~5439--5445, IEEE, 2019.

\bibitem{gordon2019depth}
A.~Gordon, H.~Li, R.~Jonschkowski, and A.~Angelova, ``Depth from videos in the
  wild: Unsupervised monocular depth learning from unknown cameras,'' in {\em
  The International Conference on Computer Vision (ICCV)}, pp.~8977--8986,
  2019.

\bibitem{bian2019unsupervised}
J.~Bian, Z.~Li, N.~Wang, H.~Zhan, C.~Shen, M.-M. Cheng, and I.~Reid,
  ``Unsupervised scale-consistent depth and ego-motion learning from monocular
  video,'' in {\em Neural Information Processing Systems (NeurIPS)},
  pp.~35--45, 2019.

\bibitem{Li0CXYZ20}
S.~Li, X.~Wang, Y.~Cao, F.~Xue, Z.~Yan, and H.~Zha, ``Self-supervised deep
  visual odometry with online adaptation,'' in {\em IEEE/CVF International
  Conference on Computer Vision and Pattern Recognition (CVPR)},
  pp.~6338--6347, 2020.

\bibitem{ZouJTHC20}
Y.~Zou, P.~Ji, Q.~Tran, J.~Huang, and M.~Chandraker, ``Learning monocular
  visual odometry via self-supervised long-term modeling,'' in {\em European
  Conference on Computer Vision (ECCV)}, vol.~12359, pp.~710--727, Springer,
  2020.

\bibitem{zhao2020masked}
C.~Zhao, G.~G. Yen, Q.~Sun, C.~Zhang, and Y.~Tang, ``Masked gan for
  unsupervised depth and pose prediction with scale consistency,'' {\em IEEE
  Transactions on Neural Networks and Learning Systems}, vol.~32, no.~12,
  pp.~5392--5403, 2020.

\bibitem{ChiWHGY21}
C.~Chi, Q.~Wang, T.~Hao, P.~Guo, and X.~Yang, ``Feature-level collaboration:
  Joint unsupervised learning of optical flow, stereo depth and camera
  motion,'' in {\em IEEE/CVF International Conference on Computer Vision and
  Pattern Recognition (CVPR)}, pp.~2463--2473, 2021.

\bibitem{LiWCZ21}
S.~Li, X.~Wu, Y.~Cao, and H.~Zha, ``Generalizing to the open world: Deep visual
  odometry with online adaptation,'' in {\em IEEE/CVF International Conference
  on Computer Vision and Pattern Recognition (CVPR)}, pp.~13184--13193, 2021.

\bibitem{dai2022self}
J.~Dai, X.~Gong, Y.~Li, J.~Wang, and M.~Wei, ``Self-supervised deep visual
  odometry based on geometric attention model,'' {\em IEEE Transactions on
  Intelligent Transportation Systems}, 2022.

\bibitem{zhang2022towards}
S.~Zhang, J.~Zhang, and D.~Tao, ``Towards scale consistent monocular visual
  odometry by learning from the virtual world,'' {\em The International
  Conference on Robotics and Automation (ICRA)}, 2022.

\bibitem{haarnoja2016}
T.~Haarnoja, A.~Ajay, S.~Levine, and P.~Abbeel, ``{Backprop KF: Learning
  Discriminative Deterministic State Estimators},'' in {\em Neural Information
  Processing Systems (NeurIPS)}, 2016.

\bibitem{yin2017scale}
X.~Yin, X.~Wang, X.~Du, and Q.~Chen, ``Scale recovery for monocular visual
  odometry using depth estimated with deep convolutional neural fields,'' in
  {\em The International Conference on Computer Vision (ICCV)}, pp.~5870--5878,
  2017.

\bibitem{barnes2018driven}
D.~Barnes, W.~Maddern, G.~Pascoe, and I.~Posner, ``Driven to distraction:
  Self-supervised distractor learning for robust monocular visual odometry in
  urban environments,'' in {\em The IEEE International Conference on Robotics
  and Automation (ICRA)}, pp.~1894--1900, IEEE, 2018.

\bibitem{Jonschkowski2018}
R.~Jonschkowski, D.~Rastogi, and O.~Brock, ``{Differentiable Particle Filters:
  End-to-End Learning with Algorithmic Priors},'' in {\em Robotics: Science and
  Systems}, 2018.

\bibitem{yang2018deep}
N.~Yang, R.~Wang, J.~Stuckler, and D.~Cremers, ``Deep virtual stereo odometry:
  Leveraging deep depth prediction for monocular direct sparse odometry,'' in
  {\em The European Conference on Computer Vision (ECCV)}, pp.~817--833, 2018.

\bibitem{loo2019cnn}
S.~Y. Loo, A.~J. Amiri, S.~Mashohor, S.~H. Tang, and H.~Zhang, ``Cnn-svo:
  Improving the mapping in semi-direct visual odometry using single-image depth
  prediction,'' in {\em The IEEE International Conference on Robotics and
  Automation (ICRA)}, pp.~5218--5223, IEEE, 2019.

\bibitem{zhan2020visual}
H.~Zhan, C.~S. Weerasekera, J.~Bian, and I.~Reid, ``Visual odometry revisited:
  What should be learnt?,'' {\em The IEEE International Conference on Robotics
  and Automation (ICRA)}, 2020.

\bibitem{WagstaffPK20}
B.~Wagstaff, V.~Peretroukhin, and J.~Kelly, ``Self-supervised deep pose
  corrections for robust visual odometry,'' in {\em The IEEE International
  Conference on Robotics and Automation (ICRA)}, pp.~2331--2337, {IEEE}, 2020.

\bibitem{sun2022improving}
L.~Sun, W.~Yin, E.~Xie, Z.~Li, C.~Sun, and C.~Shen, ``Improving monocular
  visual odometry using learned depth,'' {\em IEEE Transactions on Robotics},
  vol.~38, no.~5, pp.~3173--3186, 2022.

\bibitem{li2019sequential}
S.~Li, F.~Xue, X.~Wang, Z.~Yan, and H.~Zha, ``Sequential adversarial learning
  for self-supervised deep visual odometry,'' in {\em The International
  Conference on Computer Vision (ICCV)}, pp.~2851--2860, 2019.

\bibitem{sun2021unsupervised}
Q.~Sun, Y.~Tang, C.~Zhang, C.~Zhao, F.~Qian, and J.~Kurths, ``Unsupervised
  estimation of monocular depth and vo in dynamic environments via hybrid
  masks,'' {\em IEEE Transactions on Neural Networks and Learning Systems},
  vol.~33, no.~5, pp.~2023--2033, 2021.

\bibitem{godard2017unsupervised}
C.~Godard, O.~Mac~Aodha, and G.~J. Brostow, ``Unsupervised monocular depth
  estimation with left-right consistency,'' in {\em IEEE/CVF International
  Conference on Computer Vision and Pattern Recognition (CVPR)}, pp.~270--279,
  2017.

\bibitem{engel2017direct}
J.~Engel, V.~Koltun, and D.~Cremers, ``Direct sparse odometry,'' {\em IEEE
  transactions on pattern analysis and machine intelligence}, vol.~40, no.~3,
  pp.~611--625, 2017.

\bibitem{kendall2017uncertainties}
A.~Kendall and Y.~Gal, ``What uncertainties do we need in bayesian deep
  learning for computer vision?,'' in {\em Neural Information Processing
  Systems (NeurIPS)}, pp.~5574--5584, 2017.

\bibitem{Burri2016}
M.~Burri, J.~Nikolic, P.~Gohl, T.~Schneider, J.~Rehder, S.~Omari, M.~W.
  Achtelik, and R.~Siegwart, ``{The EuRoC micro aerial vehicle datasets},''
  {\em International Journal of Robotics Research}, vol.~35, no.~10,
  pp.~1157--1163, 2016.

\bibitem{vi-dso}
L.~Von~Stumberg, V.~Usenko, and D.~Cremers, ``Direct sparse visual-inertial
  odometry using dynamic marginalization,'' in {\em The IEEE International
  Conference on Robotics and Automation (ICRA)}, pp.~2510--2517, IEEE, 2018.

\bibitem{chen2021dynanet}
C.~Chen, C.~X. Lu, B.~Wang, N.~Trigoni, and A.~Markham, ``Dynanet: Neural
  kalman dynamical model for motion estimation and prediction,'' {\em IEEE
  Transactions on Neural Networks and Learning Systems}, vol.~32, no.~12,
  pp.~5479--5491, 2021.

\bibitem{laskar2017camera}
Z.~Laskar, I.~Melekhov, S.~Kalia, and J.~Kannala, ``Camera relocalization by
  computing pairwise relative poses using convolutional neural network,'' in
  {\em The International Conference on Computer Vision (ICCV) Workshops},
  pp.~929--938, 2017.

\bibitem{wang2018dels}
P.~Wang, R.~Yang, B.~Cao, W.~Xu, and Y.~Lin, ``Dels-3d: Deep localization and
  segmentation with a 3d semantic map,'' in {\em IEEE/CVF International
  Conference on Computer Vision and Pattern Recognition (CVPR)},
  pp.~5860--5869, 2018.

\bibitem{saha2018improved}
S.~Saha, G.~Varma, and C.~Jawahar, ``Improved visual relocalization by
  discovering anchor points,'' {\em British Machine Vision Conference (BMVC)},
  2018.

\bibitem{balntas2018relocnet}
V.~Balntas, S.~Li, and V.~Prisacariu, ``Relocnet: Continuous metric learning
  relocalisation using neural nets,'' in {\em The European Conference on
  Computer Vision (ECCV)}, pp.~751--767, 2018.

\bibitem{ding2019camnet}
M.~Ding, Z.~Wang, J.~Sun, J.~Shi, and P.~Luo, ``Camnet: Coarse-to-fine
  retrieval for camera re-localization,'' in {\em The International Conference
  on Computer Vision (ICCV)}, pp.~2871--2880, 2019.

\bibitem{sarlin2021back}
P.-E. Sarlin, A.~Unagar, M.~Larsson, H.~Germain, C.~Toft, V.~Larsson,
  M.~Pollefeys, V.~Lepetit, L.~Hammarstrand, F.~Kahl, {\em et~al.}, ``Back to
  the feature: Learning robust camera localization from pixels to pose,'' in
  {\em Proceedings of the IEEE/CVF Conference on Computer Vision and Pattern
  Recognition}, pp.~3247--3257, 2021.

\bibitem{kendall2015posenet}
A.~Kendall, M.~Grimes, and R.~Cipolla, ``Posenet: A convolutional network for
  real-time 6-dof camera relocalization,'' in {\em The International Conference
  on Computer Vision (ICCV)}, pp.~2938--2946, 2015.

\bibitem{kendall2016modelling}
A.~Kendall and R.~Cipolla, ``Modelling uncertainty in deep learning for camera
  relocalization,'' in {\em The IEEE International Conference on Robotics and
  Automation (ICRA)}, pp.~4762--4769, IEEE, 2016.

\bibitem{wu2017delving}
J.~Wu, L.~Ma, and X.~Hu, ``Delving deeper into convolutional neural networks
  for camera relocalization,'' in {\em The IEEE International Conference on
  Robotics and Automation (ICRA)}, pp.~5644--5651, IEEE, 2017.

\bibitem{Clark2017}
R.~Clark, S.~Wang, A.~Markham, N.~Trigoni, and H.~Wen, ``{VidLoc: A Deep
  Spatio-Temporal Model for 6-DoF Video-Clip Relocalization},'' in {\em
  IEEE/CVF International Conference on Computer Vision and Pattern Recognition
  (CVPR)}, 2017.

\bibitem{kendall2017geometric}
A.~Kendall and R.~Cipolla, ``Geometric loss functions for camera pose
  regression with deep learning,'' in {\em IEEE/CVF International Conference on
  Computer Vision and Pattern Recognition (CVPR)}, pp.~5974--5983, 2017.

\bibitem{naseer2017deep}
T.~Naseer and W.~Burgard, ``Deep regression for monocular camera-based 6-dof
  global localization in outdoor environments,'' in {\em The IEEE/RSJ
  International Conference on Intelligent Robots and Systems (IROS)},
  pp.~1525--1530, IEEE, 2017.

\bibitem{walch2017image}
F.~Walch, C.~Hazirbas, L.~Leal-Taixe, T.~Sattler, S.~Hilsenbeck, and
  D.~Cremers, ``Image-based localization using lstms for structured feature
  correlation,'' in {\em The International Conference on Computer Vision
  (ICCV)}, pp.~627--637, 2017.

\bibitem{melekhov2017image}
I.~Melekhov, J.~Ylioinas, J.~Kannala, and E.~Rahtu, ``Image-based localization
  using hourglass networks,'' in {\em The International Conference on Computer
  Vision (ICCV) Workshops}, pp.~879--886, 2017.

\bibitem{Brahmbhatt2018}
S.~Brahmbhatt, J.~Gu, K.~Kim, J.~Hays, and J.~Kautz, ``{Geometry-Aware Learning
  of Maps for Camera Localization},'' in {\em IEEE/CVF International Conference
  on Computer Vision and Pattern Recognition (CVPR)}, pp.~2616--2625, 2018.

\bibitem{purkait2018synthetic}
P.~Purkait, C.~Zhao, and C.~Zach, ``Synthetic view generation for absolute pose
  regression and image synthesis.,'' in {\em British Machine Vision Conference
  (BMVC)}, p.~69, 2018.

\bibitem{cai2018hybrid}
M.~Cai, C.~Shen, and I.~D. Reid, ``A hybrid probabilistic model for camera
  relocalization,'' in {\em British Machine Vision Conference (BMVC)}, vol.~1,
  p.~8, 2018.

\bibitem{xue2019local}
F.~Xue, X.~Wang, Z.~Yan, Q.~Wang, J.~Wang, and H.~Zha, ``Local supports global:
  Deep camera relocalization with sequence enhancement,'' in {\em The
  International Conference on Computer Vision (ICCV)}, pp.~2841--2850, 2019.

\bibitem{huang2019prior}
Z.~Huang, Y.~Xu, J.~Shi, X.~Zhou, H.~Bao, and G.~Zhang, ``Prior guided dropout
  for robust visual localization in dynamic environments,'' in {\em The
  International Conference on Computer Vision (ICCV)}, pp.~2791--2800, 2019.

\bibitem{bui2019adversarial}
M.~Bui, C.~Baur, N.~Navab, S.~Ilic, and S.~Albarqouni, ``Adversarial networks
  for camera pose regression and refinement,'' in {\em The International
  Conference on Computer Vision (ICCV) Workshops}, pp.~0--0, 2019.

\bibitem{wang2019atloc}
B.~Wang, C.~Chen, C.~X. Lu, P.~Zhao, N.~Trigoni, and A.~Markham, ``Atloc:
  Attention guided camera localization,'' {\em The Conference on Artificial
  Intelligence (AAAI)}, 2020.

\bibitem{xue2020learning}
F.~Xue, X.~Wu, S.~Cai, and J.~Wang, ``Learning multi-view camera relocalization
  with graph neural networks,'' in {\em IEEE/CVF International Conference on
  Computer Vision and Pattern Recognition (CVPR)}, pp.~11372--11381, IEEE,
  2020.

\bibitem{shavit2021learning}
Y.~Shavit, R.~Ferens, and Y.~Keller, ``Learning multi-scene absolute pose
  regression with transformers,'' in {\em Proceedings of the IEEE/CVF
  International Conference on Computer Vision}, pp.~2733--2742, 2021.

\bibitem{shotton2013scene}
J.~Shotton, B.~Glocker, C.~Zach, S.~Izadi, A.~Criminisi, and A.~Fitzgibbon,
  ``Scene coordinate regression forests for camera relocalization in rgb-d
  images,'' in {\em IEEE/CVF International Conference on Computer Vision and
  Pattern Recognition}, pp.~2930--2937, 2013.

\bibitem{torii201524}
A.~Torii, R.~Arandjelovic, J.~Sivic, M.~Okutomi, and T.~Pajdla, ``24/7 place
  recognition by view synthesis,'' in {\em IEEE/CVF International Conference on
  Computer Vision and Pattern Recognition (CVPR)}, pp.~1808--1817, 2015.

\bibitem{ge2020self}
Y.~Ge, H.~Wang, F.~Zhu, R.~Zhao, and H.~Li, ``Self-supervising fine-grained
  region similarities for large-scale image localization,'' in {\em European
  Conference on Computer Vision}, pp.~369--386, Springer, 2020.

\bibitem{thoma2020soft}
J.~Thoma, D.~P. Paudel, and L.~V. Gool, ``Soft contrastive learning for visual
  localization,'' {\em Advances in Neural Information Processing Systems},
  vol.~33, pp.~11119--11130, 2020.

\bibitem{chen2014convolutional}
Z.~Chen, O.~Lam, A.~Jacobson, and M.~Milford, ``Convolutional neural
  network-based place recognition,'' {\em Australasian Conference on Robotics
  and Automation}, 2014.

\bibitem{sunderhauf2015performance}
N.~S{\"u}nderhauf, S.~Shirazi, F.~Dayoub, B.~Upcroft, and M.~Milford, ``On the
  performance of convnet features for place recognition,'' in {\em The IEEE/RSJ
  International Conference on Intelligent Robots and Systems (IROS)},
  pp.~4297--4304, IEEE, 2015.

\bibitem{arandjelovic2016netvlad}
R.~Arandjelovic, P.~Gronat, A.~Torii, T.~Pajdla, and J.~Sivic, ``Netvlad: Cnn
  architecture for weakly supervised place recognition,'' in {\em IEEE/CVF
  International Conference on Computer Vision and Pattern Recognition (CVPR)},
  pp.~5297--5307, 2016.

\bibitem{jegou2010aggregating}
H.~J{\'e}gou, M.~Douze, C.~Schmid, and P.~P{\'e}rez, ``Aggregating local
  descriptors into a compact image representation,'' in {\em IEEE/CVF
  International Conference on Computer Vision and Pattern Recognition (CVPR)},
  pp.~3304--3311, IEEE, 2010.

\bibitem{zhou2019learn}
Q.~Zhou, T.~Sattler, M.~Pollefeys, and L.~Leal-Taixe, ``To learn or not to
  learn: Visual localization from essential matrices,'' {\em The IEEE
  International Conference on Robotics and Automation (ICRA)}, 2019.

\bibitem{melekhov2019dgc}
I.~Melekhov, A.~Tiulpin, T.~Sattler, M.~Pollefeys, E.~Rahtu, and J.~Kannala,
  ``Dgc-net: Dense geometric correspondence network,'' in {\em IEEE Winter
  Conference on Applications of Computer Vision (WACV)}, pp.~1034--1042, IEEE,
  2019.

\bibitem{zhuang2021fusing}
B.~Zhuang and M.~Chandraker, ``Fusing the old with the new: Learning relative
  camera pose with geometry-guided uncertainty,'' in {\em Proceedings of the
  IEEE/CVF Conference on Computer Vision and Pattern Recognition}, pp.~32--42,
  2021.

\bibitem{szegedy2015going}
C.~Szegedy, W.~Liu, Y.~Jia, P.~Sermanet, S.~Reed, D.~Anguelov, D.~Erhan,
  V.~Vanhoucke, and A.~Rabinovich, ``Going deeper with convolutions,'' in {\em
  IEEE/CVF International Conference on Computer Vision and Pattern Recognition
  (CVPR)}, 2015.

\bibitem{zhu2021learning}
Y.~Zhu, R.~Gao, S.~Huang, S.-C. Zhu, and Y.~N. Wu, ``Learning neural
  representation of camera pose with matrix representation of pose shift via
  view synthesis,'' in {\em Proceedings of the IEEE/CVF Conference on Computer
  Vision and Pattern Recognition}, pp.~9959--9968, 2021.

\bibitem{bui20206d}
M.~Bui, T.~Birdal, H.~Deng, S.~Albarqouni, L.~Guibas, S.~Ilic, and N.~Navab,
  ``6d camera relocalization in ambiguous scenes via continuous multimodal
  inference,'' in {\em European Conference on Computer Vision}, pp.~139--157,
  Springer, 2020.

\bibitem{valada2018deep}
A.~Valada, N.~Radwan, and W.~Burgard, ``Deep auxiliary learning for visual
  localization and odometry,'' in {\em The IEEE International Conference on
  Robotics and Automation (ICRA)}, pp.~6939--6946, IEEE, 2018.

\bibitem{tian20203d}
M.~Tian, Q.~Nie, and H.~Shen, ``3d scene geometry-aware constraint for camera
  localization with deep learning,'' in {\em 2020 IEEE International Conference
  on Robotics and Automation (ICRA)}, pp.~4211--4217, IEEE, 2020.

\bibitem{radwan2018vlocnet++}
N.~Radwan, A.~Valada, and W.~Burgard, ``Vlocnet++: Deep multitask learning for
  semantic visual localization and odometry,'' {\em IEEE Robotics and
  Automation Letters}, vol.~3, no.~4, pp.~4407--4414, 2018.

\bibitem{li2021pogo}
X.~Li and H.~Ling, ``Pogo-net: Pose graph optimization with graph neural
  networks,'' in {\em Proceedings of the IEEE/CVF International Conference on
  Computer Vision}, pp.~5895--5905, 2021.

\bibitem{Tzeng2017}
E.~Tzeng, J.~Hoffman, K.~Saenko, and T.~Darrell, ``{Adversarial Discriminative
  Domain Adaptation},'' in {\em IEEE/CVF International Conference on Computer
  Vision and Pattern Recognition (CVPR)}, 2017.

\bibitem{li2010location}
Y.~Li, N.~Snavely, and D.~P. Huttenlocher, ``Location recognition using
  prioritized feature matching,'' in {\em The European Conference on Computer
  Vision (ECCV)}, pp.~791--804, Springer, 2010.

\bibitem{li2012worldwide}
Y.~Li, N.~Snavely, D.~Huttenlocher, and P.~Fua, ``Worldwide pose estimation
  using 3d point clouds,'' in {\em The European Conference on Computer Vision
  (ECCV)}, pp.~15--29, Springer, 2012.

\bibitem{zeisl2015camera}
B.~Zeisl, T.~Sattler, and M.~Pollefeys, ``Camera pose voting for large-scale
  image-based localization,'' in {\em The International Conference on Computer
  Vision (ICCV)}, pp.~2704--2712, 2015.

\bibitem{germain2021neural}
H.~Germain, V.~Lepetit, and G.~Bourmaud, ``Neural reprojection error: Merging
  feature learning and camera pose estimation,'' in {\em Proceedings of the
  IEEE/CVF Conference on Computer Vision and Pattern Recognition},
  pp.~414--423, 2021.

\bibitem{cavallari2017fly}
T.~Cavallari, S.~Golodetz, N.~A. Lord, J.~Valentin, L.~Di~Stefano, and P.~H.
  Torr, ``On-the-fly adaptation of regression forests for online camera
  relocalisation,'' in {\em IEEE/CVF International Conference on Computer
  Vision and Pattern Recognition}, pp.~4457--4466, 2017.

\bibitem{guzman2014multi}
A.~Guzman-Rivera, P.~Kohli, B.~Glocker, J.~Shotton, T.~Sharp, A.~Fitzgibbon,
  and S.~Izadi, ``Multi-output learning for camera relocalization,'' in {\em
  IEEE/CVF International Conference on Computer Vision and Pattern Recognition
  (CVPR)}, pp.~1114--1121, 2014.

\bibitem{massiceti2017random}
D.~Massiceti, A.~Krull, E.~Brachmann, C.~Rother, and P.~H. Torr, ``Random
  forests versus neural networks—what's best for camera localization?,'' in
  {\em The IEEE International Conference on Robotics and Automation (ICRA)},
  pp.~5118--5125, IEEE, 2017.

\bibitem{gao2003complete}
X.-S. Gao, X.-R. Hou, J.~Tang, and H.-F. Cheng, ``Complete solution
  classification for the perspective-three-point problem,'' {\em IEEE
  transactions on pattern analysis and machine intelligence}, vol.~25, no.~8,
  pp.~930--943, 2003.

\bibitem{wald2020beyond}
J.~Wald, T.~Sattler, S.~Golodetz, T.~Cavallari, and F.~Tombari, ``Beyond
  controlled environments: 3d camera re-localization in changing indoor
  scenes,'' in {\em European Conference on Computer Vision}, pp.~467--487,
  Springer, 2020.

\bibitem{noh2017large}
H.~Noh, A.~Araujo, J.~Sim, T.~Weyand, and B.~Han, ``Large-scale image retrieval
  with attentive deep local features,'' in {\em The International Conference on
  Computer Vision (ICCV)}, pp.~3456--3465, 2017.

\bibitem{taira2018inloc}
H.~Taira, M.~Okutomi, T.~Sattler, M.~Cimpoi, M.~Pollefeys, J.~Sivic, T.~Pajdla,
  and A.~Torii, ``Inloc: Indoor visual localization with dense matching and
  view synthesis,'' in {\em IEEE/CVF International Conference on Computer
  Vision and Pattern Recognition (CVPR)}, pp.~7199--7209, 2018.

\bibitem{schonberger2018semantic}
J.~L. Sch{\"o}nberger, M.~Pollefeys, A.~Geiger, and T.~Sattler, ``Semantic
  visual localization,'' in {\em IEEE/CVF International Conference on Computer
  Vision and Pattern Recognition (CVPR)}, pp.~6896--6906, 2018.

\bibitem{detone2018superpoint}
D.~DeTone, T.~Malisiewicz, and A.~Rabinovich, ``Superpoint: Self-supervised
  interest point detection and description,'' in {\em IEEE/CVF International
  Conference on Computer Vision and Pattern Recognition (CVPR) Workshops},
  pp.~224--236, 2018.

\bibitem{sarlin2018leveraging}
P.-E. Sarlin, F.~Debraine, M.~Dymczyk, R.~Siegwart, and C.~Cadena, ``Leveraging
  deep visual descriptors for hierarchical efficient localization,'' {\em The
  Annual Conference on Robot Learning (CoRL)}, 2018.

\bibitem{rocco2018neighbourhood}
I.~Rocco, M.~Cimpoi, R.~Arandjelovi{\'c}, A.~Torii, T.~Pajdla, and J.~Sivic,
  ``Neighbourhood consensus networks,'' in {\em Neural Information Processing
  Systems (NeurIPS)}, pp.~1651--1662, 2018.

\bibitem{feng20192d3d}
M.~Feng, S.~Hu, M.~H. Ang, and G.~H. Lee, ``2d3d-matchnet: learning to match
  keypoints across 2d image and 3d point cloud,'' in {\em The IEEE
  International Conference on Robotics and Automation (ICRA)}, pp.~4790--4796,
  IEEE, 2019.

\bibitem{sarlin2019coarse}
P.-E. Sarlin, C.~Cadena, R.~Siegwart, and M.~Dymczyk, ``From coarse to fine:
  Robust hierarchical localization at large scale,'' in {\em IEEE/CVF
  International Conference on Computer Vision and Pattern Recognition (CVPR)},
  pp.~12716--12725, 2019.

\bibitem{dusmanu2019d2}
M.~Dusmanu, I.~Rocco, T.~Pajdla, M.~Pollefeys, J.~Sivic, A.~Torii, and
  T.~Sattler, ``D2-net: A trainable cnn for joint description and detection of
  local features,'' in {\em IEEE/CVF International Conference on Computer
  Vision and Pattern Recognition}, pp.~8092--8101, 2019.

\bibitem{speciale2019privacy}
P.~Speciale, J.~L. Schonberger, S.~B. Kang, S.~N. Sinha, and M.~Pollefeys,
  ``Privacy preserving image-based localization,'' in {\em IEEE/CVF
  International Conference on Computer Vision and Pattern Recognition (CVPR)},
  pp.~5493--5503, 2019.

\bibitem{weinzaepfel2019visual}
P.~Weinzaepfel, G.~Csurka, Y.~Cabon, and M.~Humenberger, ``Visual localization
  by learning objects-of-interest dense match regression,'' in {\em IEEE/CVF
  International Conference on Computer Vision and Pattern Recognition (CVPR)},
  pp.~5634--5643, 2019.

\bibitem{camposeco2019hybrid}
F.~Camposeco, A.~Cohen, M.~Pollefeys, and T.~Sattler, ``Hybrid scene
  compression for visual localization,'' in {\em IEEE/CVF International
  Conference on Computer Vision and Pattern Recognition (CVPR)},
  pp.~7653--7662, 2019.

\bibitem{cheng2019cascaded}
W.~Cheng, W.~Lin, K.~Chen, and X.~Zhang, ``Cascaded parallel filtering for
  memory-efficient image-based localization,'' in {\em The International
  Conference on Computer Vision (ICCV)}, pp.~1032--1041, 2019.

\bibitem{taira2019right}
H.~Taira, I.~Rocco, J.~Sedlar, M.~Okutomi, J.~Sivic, T.~Pajdla, T.~Sattler, and
  A.~Torii, ``Is this the right place? geometric-semantic pose verification for
  indoor visual localization,'' in {\em The International Conference on
  Computer Vision (ICCV)}, pp.~4373--4383, 2019.

\bibitem{revaud2019r2d2}
J.~Revaud, P.~Weinzaepfel, C.~De~Souza, N.~Pion, G.~Csurka, Y.~Cabon, and
  M.~Humenberger, ``R2d2: Repeatable and reliable detector and descriptor,''
  {\em Neural Information Processing Systems (NeurIPS)}, 2019.

\bibitem{luo2020aslfeat}
Z.~Luo, L.~Zhou, X.~Bai, H.~Chen, J.~Zhang, Y.~Yao, S.~Li, T.~Fang, and
  L.~Quan, ``Aslfeat: Learning local features of accurate shape and
  localization,'' {\em IEEE/CVF Conference on Computer Vision and Pattern
  Recognition (CVPR)}, 2020.

\bibitem{dusmanu2021cross}
M.~Dusmanu, O.~Miksik, J.~L. Sch{\"o}nberger, and M.~Pollefeys,
  ``Cross-descriptor visual localization and mapping,'' in {\em Proceedings of
  the IEEE/CVF International Conference on Computer Vision}, pp.~6058--6067,
  2021.

\bibitem{huang2021vs}
Z.~Huang, H.~Zhou, Y.~Li, B.~Yang, Y.~Xu, X.~Zhou, H.~Bao, G.~Zhang, and H.~Li,
  ``Vs-net: Voting with segmentation for visual localization,'' in {\em
  Proceedings of the IEEE/CVF Conference on Computer Vision and Pattern
  Recognition}, pp.~6101--6111, 2021.

\bibitem{brachmann2017dsac}
E.~Brachmann, A.~Krull, S.~Nowozin, J.~Shotton, F.~Michel, S.~Gumhold, and
  C.~Rother, ``Dsac-differentiable ransac for camera localization,'' in {\em
  IEEE/CVF International Conference on Computer Vision and Pattern Recognition
  (CVPR)}, pp.~6684--6692, 2017.

\bibitem{brachmann2018learning}
E.~Brachmann and C.~Rother, ``Learning less is more-6d camera localization via
  3d surface regression,'' in {\em IEEE/CVF International Conference on
  Computer Vision and Pattern Recognition (CVPR)}, pp.~4654--4662, 2018.

\bibitem{li2018scene}
X.~Li, J.~Ylioinas, J.~Verbeek, and J.~Kannala, ``Scene coordinate regression
  with angle-based reprojection loss for camera relocalization,'' in {\em The
  European Conference on Computer Vision (ECCV)}, pp.~0--0, 2018.

\bibitem{li2018full}
X.~Li, J.~Ylioinas, and J.~Kannala, ``Full-frame scene coordinate regression
  for image-based localization,'' {\em Robotics: Science and Systems}, 2018.

\bibitem{bui2018scene}
M.~Bui, S.~Albarqouni, S.~Ilic, and N.~Navab, ``Scene coordinate and
  correspondence learning for image-based localization,'' {\em British Machine
  Vision Conference (BMVC)}, 2018.

\bibitem{brachmann2019expert}
E.~Brachmann and C.~Rother, ``Expert sample consensus applied to camera
  re-localization,'' in {\em The International Conference on Computer Vision
  (ICCV)}, pp.~7525--7534, 2019.

\bibitem{brachmann2019neural}
E.~Brachmann and C.~Rother, ``Neural-guided ransac: Learning where to sample
  model hypotheses,'' in {\em The International Conference on Computer Vision
  (ICCV)}, pp.~4322--4331, 2019.

\bibitem{yang2019sanet}
L.~Yang, Z.~Bai, C.~Tang, H.~Li, Y.~Furukawa, and P.~Tan, ``Sanet: Scene
  agnostic network for camera localization,'' in {\em The International
  Conference on Computer Vision (ICCV)}, pp.~42--51, 2019.

\bibitem{cai2019camera}
M.~Cai, H.~Zhan, C.~Saroj~Weerasekera, K.~Li, and I.~Reid, ``Camera
  relocalization by exploiting multi-view constraints for scene coordinates
  regression,'' in {\em The International Conference on Computer Vision (ICCV)
  Workshops}, pp.~0--0, 2019.

\bibitem{li2020hscnet}
X.~Li, S.~Wang, Y.~Zhao, J.~Verbeek, and J.~Kannala, ``Hierarchical scene
  coordinate classification and regression for visual localization,'' in {\em
  IEEE/CVF International Conference on Computer Vision and Pattern Recognition
  (CVPR)}, 2020.

\bibitem{zhou2020kfnet}
L.~Zhou, Z.~Luo, T.~Shen, J.~Zhang, M.~Zhen, Y.~Yao, T.~Fang, and L.~Quan,
  ``Kfnet: Learning temporal camera relocalization using kalman filtering,''
  {\em IEEE/CVF Conference on Computer Vision and Pattern Recognition (CVPR)},
  2020.

\bibitem{tang2021learning}
S.~Tang, C.~Tang, R.~Huang, S.~Zhu, and P.~Tan, ``Learning camera localization
  via dense scene matching,'' in {\em Proceedings of the IEEE/CVF Conference on
  Computer Vision and Pattern Recognition}, pp.~1831--1841, 2021.

\bibitem{mikolajczyk2004scale}
K.~Mikolajczyk and C.~Schmid, ``Scale \& affine invariant interest point
  detectors,'' {\em International journal of computer vision}, vol.~60, no.~1,
  pp.~63--86, 2004.

\bibitem{leutenegger2011brisk}
S.~Leutenegger, M.~Chli, and R.~Y. Siegwart, ``Brisk: Binary robust invariant
  scalable keypoints,'' in {\em The International Conference on Computer Vision
  (ICCV)}, pp.~2548--2555, Ieee, 2011.

\bibitem{bay2006surf}
H.~Bay, T.~Tuytelaars, and L.~Van~Gool, ``Surf: Speeded up robust features,''
  in {\em The European Conference on Computer Vision (ECCV)}, pp.~404--417,
  Springer, 2006.

\bibitem{lowe2004distinctive}
D.~G. Lowe, ``Distinctive image features from scale-invariant keypoints,'' {\em
  International journal of computer vision}, vol.~60, no.~2, pp.~91--110, 2004.

\bibitem{calonder2010brief}
M.~Calonder, V.~Lepetit, C.~Strecha, and P.~Fua, ``Brief: Binary robust
  independent elementary features,'' in {\em The European Conference on
  Computer Vision (ECCV)}, pp.~778--792, Springer, 2010.

\bibitem{rublee2011orb}
E.~Rublee, V.~Rabaud, K.~Konolige, and G.~Bradski, ``Orb: An efficient
  alternative to sift or surf,'' in {\em The International Conference on
  Computer Vision (ICCV)}, pp.~2564--2571, 2011.

\bibitem{balntas2016learning}
V.~Balntas, E.~Riba, D.~Ponsa, and K.~Mikolajczyk, ``Learning local feature
  descriptors with triplets and shallow convolutional neural networks.,'' in
  {\em British Machine Vision Conference (BMVC)}, vol.~1, p.~3, 2016.

\bibitem{simo2015discriminative}
E.~Simo-Serra, E.~Trulls, L.~Ferraz, I.~Kokkinos, P.~Fua, and F.~Moreno-Noguer,
  ``Discriminative learning of deep convolutional feature point descriptors,''
  in {\em The International Conference on Computer Vision (ICCV)},
  pp.~118--126, 2015.

\bibitem{simonyan2014learning}
K.~Simonyan, A.~Vedaldi, and A.~Zisserman, ``Learning local feature descriptors
  using convex optimisation,'' {\em IEEE Transactions on Pattern Analysis and
  Machine Intelligence}, vol.~36, no.~8, pp.~1573--1585, 2014.

\bibitem{moo2018learning}
K.~Moo~Yi, E.~Trulls, Y.~Ono, V.~Lepetit, M.~Salzmann, and P.~Fua, ``Learning
  to find good correspondences,'' in {\em IEEE/CVF International Conference on
  Computer Vision and Pattern Recognition (CVPR)}, pp.~2666--2674, 2018.

\bibitem{ebel2019beyond}
P.~Ebel, A.~Mishchuk, K.~M. Yi, P.~Fua, and E.~Trulls, ``Beyond cartesian
  representations for local descriptors,'' in {\em The International Conference
  on Computer Vision (ICCV)}, pp.~253--262, 2019.

\bibitem{larsson2019fine}
M.~Larsson, E.~Stenborg, C.~Toft, L.~Hammarstrand, T.~Sattler, and F.~Kahl,
  ``Fine-grained segmentation networks: Self-supervised segmentation for
  improved long-term visual localization,'' in {\em The International
  Conference on Computer Vision (ICCV)}, pp.~31--41, 2019.

\bibitem{pautrat2020online}
R.~Pautrat, V.~Larsson, M.~R. Oswald, and M.~Pollefeys, ``Online invariance
  selection for local feature descriptors,'' in {\em European Conference on
  Computer Vision}, pp.~707--724, Springer, 2020.

\bibitem{wang2020learning}
Q.~Wang, X.~Zhou, B.~Hariharan, and N.~Snavely, ``Learning feature descriptors
  using camera pose supervision,'' in {\em European Conference on Computer
  Vision}, pp.~757--774, Springer, 2020.

\bibitem{tian2020hynet}
Y.~Tian, A.~Barroso~Laguna, T.~Ng, V.~Balntas, and K.~Mikolajczyk, ``Hynet:
  Learning local descriptor with hybrid similarity measure and triplet loss,''
  {\em Advances in Neural Information Processing Systems}, vol.~33,
  pp.~7401--7412, 2020.

\bibitem{savinov2017quad}
N.~Savinov, A.~Seki, L.~Ladicky, T.~Sattler, and M.~Pollefeys, ``Quad-networks:
  unsupervised learning to rank for interest point detection,'' in {\em
  IEEE/CVF International Conference on Computer Vision and Pattern Recognition
  (CVPR)}, pp.~1822--1830, 2017.

\bibitem{zhang2018learning}
L.~Zhang and S.~Rusinkiewicz, ``Learning to detect features in texture
  images,'' in {\em IEEE/CVF International Conference on Computer Vision and
  Pattern Recognition (CVPR)}, pp.~6325--6333, 2018.

\bibitem{laguna2019key}
A.~B. Laguna, E.~Riba, D.~Ponsa, and K.~Mikolajczyk, ``Key. net: Keypoint
  detection by handcrafted and learned cnn filters,'' {\em The IEEE/CVF
  International Conference on Computer Vision (ICCV)}, 2019.

\bibitem{ono2018lf}
Y.~Ono, E.~Trulls, P.~Fua, and K.~M. Yi, ``Lf-net: learning local features from
  images,'' in {\em Neural Information Processing Systems (NeurIPS)},
  pp.~6234--6244, 2018.

\bibitem{yi2016lift}
K.~M. Yi, E.~Trulls, V.~Lepetit, and P.~Fua, ``Lift: Learned invariant feature
  transform,'' in {\em The European Conference on Computer Vision (ECCV)},
  pp.~467--483, Springer, 2016.

\bibitem{zhou2020da4ad}
Y.~Zhou, G.~Wan, S.~Hou, L.~Yu, G.~Wang, X.~Rui, and S.~Song, ``Da4ad:
  End-to-end deep attention-based visual localization for autonomous driving,''
  in {\em European Conference on Computer Vision}, pp.~271--289, Springer,
  2020.

\bibitem{lu2020rskdd}
F.~Lu, G.~Chen, Y.~Liu, Z.~Qu, and A.~Knoll, ``Rskdd-net: Random sample-based
  keypoint detector and descriptor,'' {\em Advances in Neural Information
  Processing Systems}, vol.~33, pp.~21297--21308, 2020.

\bibitem{harris1988combined}
C.~G. Harris, M.~Stephens, {\em et~al.}, ``A combined corner and edge
  detector.,'' in {\em Alvey vision conference}, vol.~15, pp.~10--5244,
  Citeseer, 1988.

\bibitem{wang2021p2}
B.~Wang, C.~Chen, Z.~Cui, J.~Qin, C.~X. Lu, Z.~Yu, P.~Zhao, Z.~Dong, F.~Zhu,
  N.~Trigoni, {\em et~al.}, ``P2-net: Joint description and detection of local
  features for pixel and point matching,'' {\em IEEE/CVF International
  Conference on Computer Vision (ICCV)}, 2021.

\bibitem{tian2020d2d}
Y.~Tian, V.~Balntas, T.~Ng, A.~Barroso-Laguna, Y.~Demiris, and K.~Mikolajczyk,
  ``D2d: Keypoint extraction with describe to detect approach,'' {\em The Asian
  Conference on Computer Vision (ACCV)}, 2020.

\bibitem{choy2016universal}
C.~B. Choy, J.~Gwak, S.~Savarese, and M.~Chandraker, ``Universal correspondence
  network,'' in {\em Neural Information Processing Systems (NeurIPS)},
  pp.~2414--2422, 2016.

\bibitem{fathy2018hierarchical}
M.~E. Fathy, Q.-H. Tran, M.~Zeeshan~Zia, P.~Vernaza, and M.~Chandraker,
  ``Hierarchical metric learning and matching for 2d and 3d geometric
  correspondences,'' in {\em The European Conference on Computer Vision
  (ECCV)}, pp.~803--819, 2018.

\bibitem{savinov2017matching}
N.~Savinov, L.~Ladicky, and M.~Pollefeys, ``Matching neural paths: transfer
  from recognition to correspondence search,'' in {\em Neural Information
  Processing Systems (NeurIPS)}, pp.~1205--1214, 2017.

\bibitem{hyeon2021pose}
J.~Hyeon, J.~Kim, and N.~Doh, ``Pose correction for highly accurate visual
  localization in large-scale indoor spaces,'' in {\em Proceedings of the
  IEEE/CVF International Conference on Computer Vision}, pp.~15974--15983,
  2021.

\bibitem{berton2021viewpoint}
G.~Berton, C.~Masone, V.~Paolicelli, and B.~Caputo, ``Viewpoint invariant dense
  matching for visual geolocalization,'' in {\em Proceedings of the IEEE/CVF
  International Conference on Computer Vision}, pp.~12169--12178, 2021.

\bibitem{sattler2018benchmarking}
T.~Sattler, W.~Maddern, C.~Toft, A.~Torii, L.~Hammarstrand, E.~Stenborg,
  D.~Safari, M.~Okutomi, M.~Pollefeys, J.~Sivic, {\em et~al.}, ``Benchmarking
  6dof outdoor visual localization in changing conditions,'' in {\em IEEE/CVF
  International Conference on Computer Vision and Pattern Recognition (CVPR)},
  pp.~8601--8610, 2018.

\bibitem{dong2021robust}
S.~Dong, Q.~Fan, H.~Wang, J.~Shi, L.~Yi, T.~Funkhouser, B.~Chen, and L.~J.
  Guibas, ``Robust neural routing through space partitions for camera
  relocalization in dynamic indoor environments,'' in {\em Proceedings of the
  IEEE/CVF Conference on Computer Vision and Pattern Recognition},
  pp.~8544--8554, 2021.

\bibitem{brachmann2020visual}
E.~Brachmann and C.~Rother, ``Visual camera re-localization from rgb and rgb-d
  images using dsac,'' {\em IEEE Transactions on Pattern Analysis and Machine
  Intelligence}, 2021.

\bibitem{wang2021continual}
S.~Wang, Z.~Laskar, I.~Melekhov, X.~Li, and J.~Kannala, ``Continual learning
  for image-based camera localization,'' in {\em Proceedings of the IEEE/CVF
  International Conference on Computer Vision}, pp.~3252--3262, 2021.

\bibitem{Newcombe2011}
R.~A. Newcombe, S.~J. Lovegrove, and A.~J. Davison, ``{DTAM : Dense Tracking
  and Mapping in Real-Time},'' in {\em The International Conference on Computer
  Vision (ICCV)}, pp.~2320--2327, 2011.

\bibitem{kerl2013dense}
C.~Kerl, J.~Sturm, and D.~Cremers, ``Dense visual slam for rgb-d cameras,'' in
  {\em The IEEE/RSJ International Conference on Intelligent Robots and Systems
  (IROS)}, pp.~2100--2106, IEEE, 2013.

\bibitem{whelan2015real}
T.~Whelan, M.~Kaess, H.~Johannsson, M.~Fallon, J.~J. Leonard, and J.~McDonald,
  ``Real-time large-scale dense rgb-d slam with volumetric fusion,'' {\em The
  International Journal of Robotics Research}, vol.~34, no.~4-5, pp.~598--626,
  2015.

\bibitem{eigen2014depth}
D.~Eigen, C.~Puhrsch, and R.~Fergus, ``Depth map prediction from a single image
  using a multi-scale deep network,'' in {\em Neural Information Processing
  Systems (NeurIPS)}, pp.~2366--2374, 2014.

\bibitem{ummenhofer2017demon}
B.~Ummenhofer, H.~Zhou, J.~Uhrig, N.~Mayer, E.~Ilg, A.~Dosovitskiy, and
  T.~Brox, ``Demon: Depth and motion network for learning monocular stereo,''
  in {\em IEEE/CVF International Conference on Computer Vision and Pattern
  Recognition}, pp.~5038--5047, 2017.

\bibitem{liu2015learning}
F.~Liu, C.~Shen, G.~Lin, and I.~Reid, ``Learning depth from single monocular
  images using deep convolutional neural fields,'' {\em IEEE transactions on
  pattern analysis and machine intelligence}, vol.~38, no.~10, pp.~2024--2039,
  2015.

\bibitem{karsch2014depth}
K.~Karsch, C.~Liu, and S.~B. Kang, ``Depth transfer: Depth extraction from
  video using non-parametric sampling,'' {\em IEEE transactions on pattern
  analysis and machine intelligence}, vol.~36, no.~11, pp.~2144--2158, 2014.

\bibitem{garg2016unsupervised}
R.~Garg, V.~K. BG, G.~Carneiro, and I.~Reid, ``Unsupervised cnn for single view
  depth estimation: Geometry to the rescue,'' in {\em The European Conference
  on Computer Vision (ECCV)}, pp.~740--756, Springer, 2016.

\bibitem{tateno2017cnn}
K.~Tateno, F.~Tombari, I.~Laina, and N.~Navab, ``Cnn-slam: Real-time dense
  monocular slam with learned depth prediction,'' in {\em IEEE/CVF
  International Conference on Computer Vision and Pattern Recognition (CVPR)},
  pp.~6243--6252, 2017.

\bibitem{engel2014lsd}
J.~Engel, T.~Sch{\"o}ps, and D.~Cremers, ``Lsd-slam: Large-scale direct
  monocular slam,'' in {\em The European Conference on Computer Vision (ECCV)},
  pp.~834--849, Springer, 2014.

\bibitem{qi2017pointnet}
C.~R. Qi, H.~Su, K.~Mo, and L.~J. Guibas, ``Pointnet: Deep learning on point
  sets for 3d classification and segmentation,'' in {\em IEEE/CVF International
  Conference on Computer Vision and Pattern Recognition (CVPR)}, pp.~652--660,
  2017.

\bibitem{fan2017point}
H.~Fan, H.~Su, and L.~J. Guibas, ``A point set generation network for 3d object
  reconstruction from a single image,'' in {\em IEEE/CVF International
  Conference on Computer Vision and Pattern Recognition (CVPR)}, pp.~605--613,
  2017.

\bibitem{groueix2018papier}
T.~Groueix, M.~Fisher, V.~G. Kim, B.~C. Russell, and M.~Aubry, ``A
  papier-m{\^a}ch{\'e} approach to learning 3d surface generation,'' in {\em
  IEEE/CVF International Conference on Computer Vision and Pattern Recognition
  (CVPR)}, pp.~216--224, 2018.

\bibitem{wang2018pixel2mesh}
N.~Wang, Y.~Zhang, Z.~Li, Y.~Fu, W.~Liu, and Y.-G. Jiang, ``Pixel2mesh:
  Generating 3d mesh models from single rgb images,'' in {\em The European
  Conference on Computer Vision (ECCV)}, pp.~52--67, 2018.

\bibitem{ladicky2017point}
L.~Ladicky, O.~Saurer, S.~Jeong, F.~Maninchedda, and M.~Pollefeys, ``From point
  clouds to mesh using regression,'' in {\em The International Conference on
  Computer Vision (ICCV)}, pp.~3893--3902, 2017.

\bibitem{dai2019scan2mesh}
A.~Dai and M.~Nie{\ss}ner, ``Scan2mesh: From unstructured range scans to 3d
  meshes,'' in {\em IEEE/CVF International Conference on Computer Vision and
  Pattern Recognition (CVPR)}, pp.~5574--5583, 2019.

\bibitem{peng2021shape}
S.~Peng, C.~Jiang, Y.~Liao, M.~Niemeyer, M.~Pollefeys, and A.~Geiger, ``Shape
  as points: A differentiable poisson solver,'' {\em Advances in Neural
  Information Processing Systems}, vol.~34, 2021.

\bibitem{mukasa20173d}
T.~Mukasa, J.~Xu, and B.~Stenger, ``3d scene mesh from cnn depth predictions
  and sparse monocular slam,'' in {\em The International Conference on Computer
  Vision (ICCV) Workshops}, pp.~921--928, 2017.

\bibitem{bloesch2019learning}
M.~Bloesch, T.~Laidlow, R.~Clark, S.~Leutenegger, and A.~J. Davison, ``Learning
  meshes for dense visual slam,'' in {\em IEEE/CVF International Conference on
  Computer Vision and Pattern Recognition (CVPR)}, pp.~5855--5864, 2019.

\bibitem{park2019deepsdf}
J.~J. Park, P.~Florence, J.~Straub, R.~Newcombe, and S.~Lovegrove, ``Deepsdf:
  Learning continuous signed distance functions for shape representation,'' in
  {\em IEEE/CVF International Conference on Computer Vision and Pattern
  Recognition (CVPR)}, pp.~165--174, 2019.

\bibitem{mescheder2019occupancy}
L.~Mescheder, M.~Oechsle, M.~Niemeyer, S.~Nowozin, and A.~Geiger, ``Occupancy
  networks: Learning 3d reconstruction in function space,'' in {\em IEEE/CVF
  International Conference on Computer Vision and Pattern Recognition (CVPR)},
  pp.~4460--4470, 2019.

\bibitem{peng2020convolutional}
S.~Peng, M.~Niemeyer, L.~Mescheder, M.~Pollefeys, and A.~Geiger,
  ``Convolutional occupancy networks,'' in {\em European Conference on Computer
  Vision}, pp.~523--540, Springer, 2020.

\bibitem{mildenhall2020nerf}
B.~Mildenhall, P.~P. Srinivasan, M.~Tancik, J.~T. Barron, R.~Ramamoorthi, and
  R.~Ng, ``Nerf: Representing scenes as neural radiance fields for view
  synthesis,'' in {\em The European Conference on Computer Vision (ECCV)},
  pp.~405--421, Springer, 2020.

\bibitem{chibane2020neural}
J.~Chibane, G.~Pons-Moll, {\em et~al.}, ``Neural unsigned distance fields for
  implicit function learning,'' {\em Advances in Neural Information Processing
  Systems}, vol.~33, pp.~21638--21652, 2020.

\bibitem{ji2017surfacenet}
M.~Ji, J.~Gall, H.~Zheng, Y.~Liu, and L.~Fang, ``Surfacenet: An end-to-end 3d
  neural network for multiview stereopsis,'' in {\em The International
  Conference on Computer Vision (ICCV)}, pp.~2307--2315, 2017.

\bibitem{paschalidou2018raynet}
D.~Paschalidou, O.~Ulusoy, C.~Schmitt, L.~Van~Gool, and A.~Geiger, ``Raynet:
  Learning volumetric 3d reconstruction with ray potentials,'' in {\em IEEE/CVF
  International Conference on Computer Vision and Pattern Recognition (CVPR)},
  pp.~3897--3906, 2018.

\bibitem{kar2017learning}
A.~Kar, C.~H{\"a}ne, and J.~Malik, ``Learning a multi-view stereo machine,'' in
  {\em Neural Information Processing Systems}, pp.~365--376, 2017.

\bibitem{hane2017hierarchical}
C.~H{\"a}ne, S.~Tulsiani, and J.~Malik, ``Hierarchical surface prediction for
  3d object reconstruction,'' in {\em International Conference on 3D Vision
  (3DV)}, pp.~412--420, 2017.

\bibitem{tatarchenko2017octree}
M.~Tatarchenko, A.~Dosovitskiy, and T.~Brox, ``Octree generating networks:
  Efficient convolutional architectures for high-resolution 3d outputs,'' in
  {\em The International Conference on Computer Vision (ICCV)}, pp.~2088--2096,
  2017.

\bibitem{dai2017shape}
A.~Dai, C.~Ruizhongtai~Qi, and M.~Nie{\ss}ner, ``Shape completion using
  3d-encoder-predictor cnns and shape synthesis,'' in {\em IEEE/CVF
  International Conference on Computer Vision and Pattern Recognition (CVPR)},
  pp.~5868--5877, 2017.

\bibitem{riegler2017octnetfusion}
G.~Riegler, A.~O. Ulusoy, H.~Bischof, and A.~Geiger, ``Octnetfusion: Learning
  depth fusion from data,'' in {\em 2017 International Conference on 3D Vision
  (3DV)}, pp.~57--66, IEEE, 2017.

\bibitem{kirillov2019panoptic}
A.~Kirillov, K.~He, R.~Girshick, C.~Rother, and P.~Doll{\'a}r, ``Panoptic
  segmentation,'' in {\em IEEE/CVF International Conference on Computer Vision
  and Pattern Recognition (CVPR)}, pp.~9404--9413, 2019.

\bibitem{mccormac2017semanticfusion}
J.~McCormac, A.~Handa, A.~Davison, and S.~Leutenegger, ``Semanticfusion: Dense
  3d semantic mapping with convolutional neural networks,'' in {\em The IEEE
  International Conference on Robotics and Automation (ICRA)}, pp.~4628--4635,
  IEEE, 2017.

\bibitem{ma2017multi}
L.~Ma, J.~St{\"u}ckler, C.~Kerl, and D.~Cremers, ``Multi-view deep learning for
  consistent semantic mapping with rgb-d cameras,'' in {\em The IEEE/RSJ
  International Conference on Intelligent Robots and Systems (IROS)},
  pp.~598--605, IEEE, 2017.

\bibitem{xiang2017rnn}
Y.~Xiang and D.~Fox, ``Da-rnn: Semantic mapping with data associated recurrent
  neural networks,'' {\em Robotics: Science and Systems}, 2017.

\bibitem{qin2021light}
T.~Qin, Y.~Zheng, T.~Chen, Y.~Chen, and Q.~Su, ``A light-weight semantic map
  for visual localization towards autonomous driving,'' in {\em 2021 IEEE
  International Conference on Robotics and Automation (ICRA)},
  pp.~11248--11254, IEEE, 2021.

\bibitem{sunderhauf2017meaningful}
N.~S{\"u}nderhauf, T.~T. Pham, Y.~Latif, M.~Milford, and I.~Reid, ``Meaningful
  maps with object-oriented semantic mapping,'' in {\em The IEEE/RSJ
  International Conference on Intelligent Robots and Systems (IROS)},
  pp.~5079--5085, IEEE, 2017.

\bibitem{grinvald2019volumetric}
M.~Grinvald, F.~Furrer, T.~Novkovic, J.~J. Chung, C.~Cadena, R.~Siegwart, and
  J.~Nieto, ``Volumetric instance-aware semantic mapping and 3d object
  discovery,'' {\em IEEE Robotics and Automation Letters}, vol.~4, no.~3,
  pp.~3037--3044, 2019.

\bibitem{mccormac2018fusion++}
J.~McCormac, R.~Clark, M.~Bloesch, A.~Davison, and S.~Leutenegger, ``Fusion++:
  Volumetric object-level slam,'' in {\em International Conference on 3D Vision
  (3DV)}, pp.~32--41, IEEE, 2018.

\bibitem{doherty2019multimodal}
K.~Doherty, D.~Fourie, and J.~Leonard, ``Multimodal semantic slam with
  probabilistic data association,'' in {\em 2019 international conference on
  robotics and automation (ICRA)}, pp.~2419--2425, IEEE, 2019.

\bibitem{narita2019panopticfusion}
G.~Narita, T.~Seno, T.~Ishikawa, and Y.~Kaji, ``Panopticfusion: Online
  volumetric semantic mapping at the level of stuff and things,'' {\em The
  IEEE/RSJ International Conference on Intelligent Robots and Systems (IROS)},
  2019.

\bibitem{Bloesch2018}
M.~Bloesch, J.~Czarnowski, R.~Clark, S.~Leutenegger, and A.~J. Davison,
  ``{CodeSLAM — Learning a Compact, Optimisable Representation for Dense
  Visual SLAM},'' in {\em IEEE/CVF International Conference on Computer Vision
  and Pattern Recognition (CVPR)}, 2018.

\bibitem{eslami2018neural}
S.~A. Eslami, D.~J. Rezende, F.~Besse, F.~Viola, A.~S. Morcos, M.~Garnelo,
  A.~Ruderman, A.~A. Rusu, I.~Danihelka, K.~Gregor, {\em et~al.}, ``Neural
  scene representation and rendering,'' {\em Science}, vol.~360, no.~6394,
  pp.~1204--1210, 2018.

\bibitem{tobin2019geometry}
J.~Tobin, W.~Zaremba, and P.~Abbeel, ``Geometry-aware neural rendering,'' in
  {\em Neural Information Processing System (NeurIPS)}, pp.~11555--11565, 2019.

\bibitem{lim2019neural}
J.~H. Lim, P.~O. Pinheiro, N.~Rostamzadeh, C.~Pal, and S.~Ahn, ``Neural
  multisensory scene inference,'' in {\em Neural Information Processing Systems
  (NeurIPS)}, pp.~8994--9004, 2019.

\bibitem{trevithick2020grf}
A.~Trevithick and B.~Yang, ``Grf: Learning a general radiance field for 3d
  scene representation and rendering,'' 2020.

\bibitem{schwarz2020graf}
K.~Schwarz, Y.~Liao, M.~Niemeyer, and A.~Geiger, ``Graf: Generative radiance
  fields for 3d-aware image synthesis,'' {\em Neural Information Processing
  Systems (NeurIPS)}, 2020.

\bibitem{yu2021plenoctrees}
A.~Yu, R.~Li, M.~Tancik, H.~Li, R.~Ng, and A.~Kanazawa, ``Plenoctrees for
  real-time rendering of neural radiance fields,'' {\em IEEE/CVF International
  Conference on Computer Vision (ICCV)}, 2021.

\bibitem{Lindell20arxiv_AutoInt}
D.~Lindell, J.~Martel, and G.~Wetzstein, ``{AutoInt}: Automatic integration for
  fast neural volume rendering,'' {\em IEEE/CVF Conference on Computer Vision
  and Pattern Recognition (CVPR)}, 2020.

\bibitem{neff2021donerf}
T.~Neff, P.~Stadlbauer, M.~Parger, A.~Kurz, J.~H. Mueller, C.~R.~A. Chaitanya,
  A.~S. Kaplanyan, and M.~Steinberger, ``{DONeRF: Towards Real-Time Rendering
  of Compact Neural Radiance Fields using Depth Oracle Networks},'' {\em
  Computer Graphics Forum}, vol.~40, no.~4, 2021.

\bibitem{Wang2021d}
P.~Wang, L.~Liu, Y.~Liu, C.~Theobalt, T.~Komura, and W.~Wang, ``{NeuS: Learning
  Neural Implicit Surfaces by Volume Rendering for Multi-view
  Reconstruction},'' {\em Neural Information Processing Systems (NeurIPS)},
  2021.

\bibitem{zhi2021place}
S.~Zhi, T.~Laidlow, S.~Leutenegger, and A.~J. Davison, ``In-place scene
  labelling and understanding with implicit scene representation,'' in {\em
  Proceedings of the IEEE/CVF International Conference on Computer Vision},
  pp.~15838--15847, 2021.

\bibitem{sucar2021imap}
E.~Sucar, S.~Liu, J.~Ortiz, and A.~J. Davison, ``imap: Implicit mapping and
  positioning in real-time,'' in {\em IEEE/CVF International Conference on
  Computer Vision (ICCV)}, pp.~6229--6238, 2021.

\bibitem{zhu2022nice}
Z.~Zhu, S.~Peng, V.~Larsson, W.~Xu, H.~Bao, Z.~Cui, M.~R. Oswald, and
  M.~Pollefeys, ``Nice-slam: Neural implicit scalable encoding for slam,'' in
  {\em Proceedings of the IEEE/CVF Conference on Computer Vision and Pattern
  Recognition}, pp.~12786--12796, 2022.

\bibitem{mirowski2016learning}
P.~Mirowski, R.~Pascanu, F.~Viola, H.~Soyer, A.~J. Ballard, A.~Banino,
  M.~Denil, R.~Goroshin, L.~Sifre, K.~Kavukcuoglu, {\em et~al.}, ``Learning to
  navigate in complex environments,'' {\em International Conference on Learning
  Representations (ICLR)}, 2017.

\bibitem{zhu2017target}
Y.~Zhu, R.~Mottaghi, E.~Kolve, J.~J. Lim, A.~Gupta, L.~Fei-Fei, and A.~Farhadi,
  ``Target-driven visual navigation in indoor scenes using deep reinforcement
  learning,'' in {\em The IEEE International Conference on Robotics and
  Automation (ICRA)}, pp.~3357--3364, IEEE, 2017.

\bibitem{mirowski2018learning}
P.~Mirowski, M.~Grimes, M.~Malinowski, K.~M. Hermann, K.~Anderson,
  D.~Teplyashin, K.~Simonyan, A.~Zisserman, R.~Hadsell, {\em et~al.},
  ``Learning to navigate in cities without a map,'' in {\em Neural Information
  Processing Systems (NeurIPS)}, pp.~2419--2430, 2018.

\bibitem{li2019deep}
H.~Li, Q.~Zhang, and D.~Zhao, ``Deep reinforcement learning-based automatic
  exploration for navigation in unknown environment,'' {\em IEEE transactions
  on neural networks and learning systems}, vol.~31, no.~6, pp.~2064--2076,
  2019.

\bibitem{banino2018vector}
A.~Banino, C.~Barry, B.~Uria, C.~Blundell, T.~Lillicrap, P.~Mirowski,
  A.~Pritzel, M.~J. Chadwick, T.~Degris, J.~Modayil, {\em et~al.},
  ``Vector-based navigation using grid-like representations in artificial
  agents,'' {\em Nature}, vol.~557, no.~7705, pp.~429--433, 2018.

\bibitem{sunderhauf2015place}
N.~S{\"u}nderhauf, S.~Shirazi, A.~Jacobson, F.~Dayoub, E.~Pepperell,
  B.~Upcroft, and M.~Milford, ``Place recognition with convnet landmarks:
  Viewpoint-robust, condition-robust, training-free,'' {\em Proceedings of
  Robotics: Science and Systems XII}, 2015.

\bibitem{gao2017unsupervised}
X.~Gao and T.~Zhang, ``Unsupervised learning to detect loops using deep neural
  networks for visual slam system,'' {\em Autonomous robots}, vol.~41, no.~1,
  pp.~1--18, 2017.

\bibitem{merrill2018lightweight}
N.~Merrill and G.~Huang, ``Lightweight unsupervised deep loop closure,'' {\em
  Robotics: Science and Systems}, 2018.

\bibitem{memon2020loop}
A.~R. Memon, H.~Wang, and A.~Hussain, ``Loop closure detection using supervised
  and unsupervised deep neural networks for monocular slam systems,'' {\em
  Robotics and Autonomous Systems}, vol.~126, p.~103470, 2020.

\bibitem{triggs1999bundle}
B.~Triggs, P.~F. McLauchlan, R.~I. Hartley, and A.~W. Fitzgibbon, ``Bundle
  adjustment—a modern synthesis,'' in {\em International workshop on vision
  algorithms}, pp.~298--372, Springer, 1999.

\bibitem{nocedal2006numerical}
J.~Nocedal and S.~Wright, {\em Numerical optimization}.
\newblock Springer Science \& Business Media, 2006.

\bibitem{clark2018learning}
R.~Clark, M.~Bloesch, J.~Czarnowski, S.~Leutenegger, and A.~J. Davison,
  ``Learning to solve nonlinear least squares for monocular stereo,'' in {\em
  The European Conference on Computer Vision (ECCV)}, pp.~284--299, 2018.

\bibitem{tang2019ba}
C.~Tang and P.~Tan, ``Ba-net: Dense bundle adjustment network,'' {\em
  International Conference on Learning Representations}, 2019.

\bibitem{zhou2020deeptam}
H.~Zhou, B.~Ummenhofer, and T.~Brox, ``Deeptam: Deep tracking and mapping with
  convolutional neural networks,'' {\em International Journal of Computer
  Vision}, vol.~128, no.~3, pp.~756--769, 2020.

\bibitem{czarnowski2020deepfactors}
J.~Czarnowski, T.~Laidlow, R.~Clark, and A.~J. Davison, ``Deepfactors:
  Real-time probabilistic dense monocular slam,'' {\em IEEE Robotics and
  Automation Letters}, 2020.

\bibitem{gal2016dropout}
Y.~Gal and Z.~Ghahramani, ``Dropout as a bayesian approximation: Representing
  model uncertainty in deep learning,'' in {\em international conference on
  machine learning}, pp.~1050--1059, 2016.

\bibitem{wang2018end}
S.~Wang, R.~Clark, H.~Wen, and N.~Trigoni, ``End-to-end, sequence-to-sequence
  probabilistic visual odometry through deep neural networks,'' {\em The
  International Journal of Robotics Research}, vol.~37, no.~4-5, pp.~513--542,
  2018.

\bibitem{kendall2017bayesian}
A.~Kendall, V.~Badrinarayanan, and R.~Cipolla, ``Bayesian segnet: Model
  uncertainty in deep convolutional encoder-decoder architectures for scene
  understanding,'' in {\em British Machine Vision Conference (BMVC)}, 2017.

\bibitem{klodt2018supervising}
M.~Klodt and A.~Vedaldi, ``Supervising the new with the old: learning sfm from
  sfm,'' in {\em The European Conference on Computer Vision (ECCV)},
  pp.~698--713, 2018.

\bibitem{Clark2017a}
R.~Clark, S.~Wang, H.~Wen, A.~Markham, and N.~Trigoni, ``{VINet :
  Visual-Inertial Odometry as a Sequence-to-Sequence Learning Problem},'' in
  {\em The Conference on Artificial Intelligence (AAAI)}, pp.~3995--4001, 2017.

\bibitem{chen2019selective}
C.~Chen, S.~Rosa, Y.~Miao, C.~X. Lu, W.~Wu, A.~Markham, and N.~Trigoni,
  ``Selective sensor fusion for neural visual-inertial odometry,'' in {\em
  IEEE/CVF International Conference on Computer Vision and Pattern Recognition
  (CVPR)}, pp.~10542--10551, 2019.

\bibitem{chen2022learning}
C.~Chen, S.~Rosa, C.~X. Lu, B.~Wang, N.~Trigoni, and A.~Markham, ``Learning
  selective sensor fusion for state estimation,'' {\em IEEE Transactions on
  Neural Networks and Learning Systems}, 2022.

\bibitem{shamwell2019unsupervised}
E.~J. Shamwell, K.~Lindgren, S.~Leung, and W.~D. Nothwang, ``Unsupervised deep
  visual-inertial odometry with online error correction for rgb-d imagery,''
  {\em IEEE transactions on pattern analysis and machine intelligence}, 2019.

\bibitem{han2019deepvio}
L.~Han, Y.~Lin, G.~Du, and S.~Lian, ``Deepvio: Self-supervised deep learning of
  monocular visual inertial odometry using 3d geometric constraints,'' {\em The
  IEEE/RSJ International Conference on Intelligent Robots and Systems (IROS)},
  2019.

\bibitem{wei2021unsupervised}
P.~Wei, G.~Hua, W.~Huang, F.~Meng, and H.~Liu, ``Unsupervised monocular
  visual-inertial odometry network,'' in {\em Proceedings of the Twenty-Ninth
  International Conference on International Joint Conferences on Artificial
  Intelligence}, pp.~2347--2354, 2021.

\bibitem{sheng2019unsupervised}
L.~Sheng, D.~Xu, W.~Ouyang, and X.~Wang, ``Unsupervised collaborative learning
  of keyframe detection and visual odometry towards monocular deep slam,'' in
  {\em The International Conference on Computer Vision (ICCV)}, pp.~4302--4311,
  2019.

\end{thebibliography}
}

\begin{IEEEbiography}
[{\includegraphics[width=1in,height=1.25in,clip,keepaspectratio]{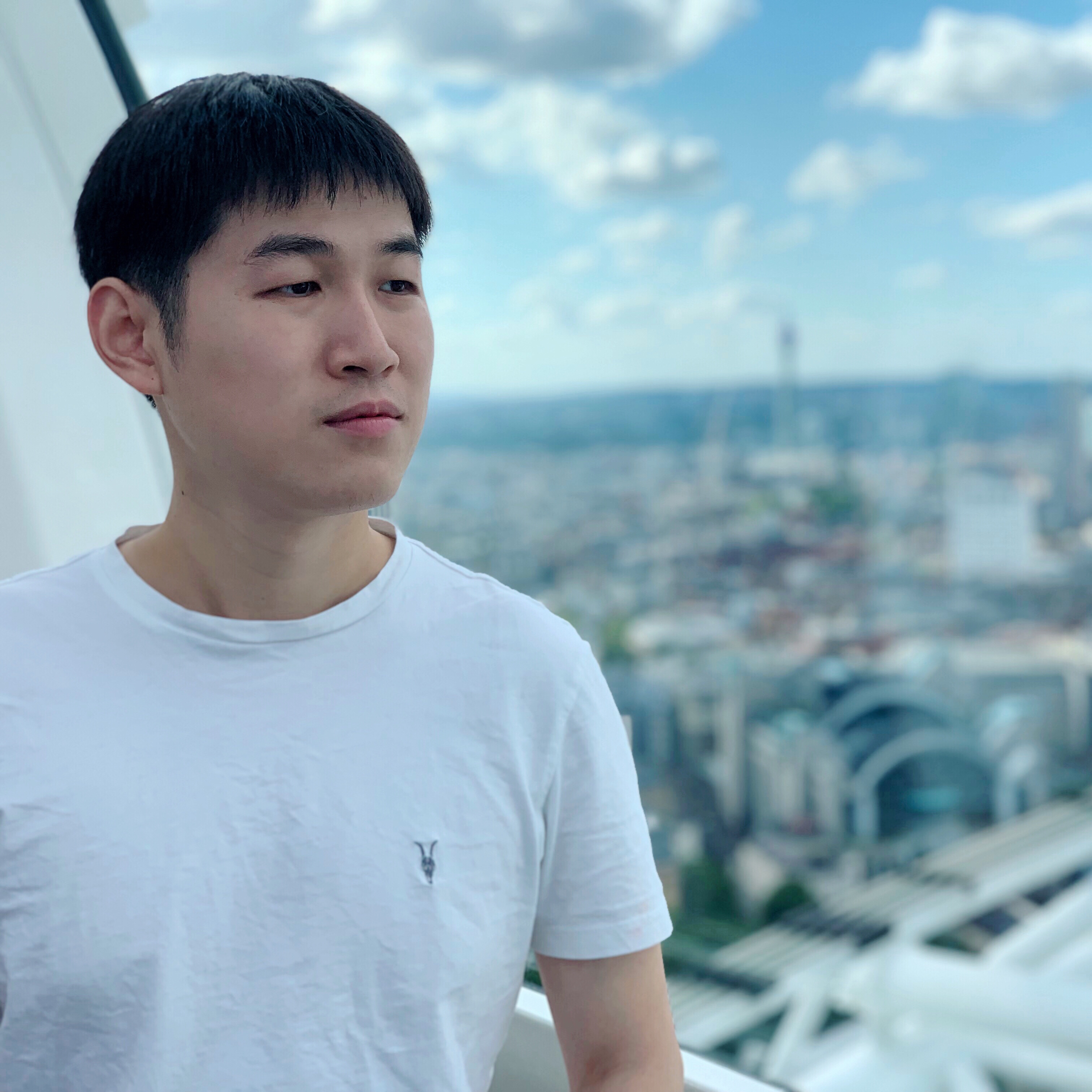}}]{Changhao Chen}
is a Lecturer at College of Intelligence Science and Technology, National University of Defense Technology. Before that, he obtained his Ph.D. degree at University of Oxford (UK), M.Eng. degree at National University of Defense Technology (China), and B.Eng. degree at Tongji University (China). His research interest lies in robotics, computer vision and cyberphysical systems. 
\end{IEEEbiography}

\begin{IEEEbiography}
[{\includegraphics[width=1in,height=1.25in,clip,keepaspectratio]{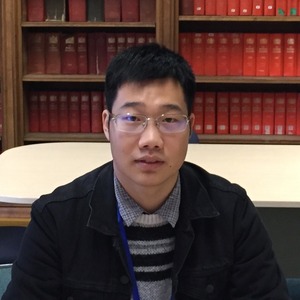}}]{Bing Wang} 
is an Assistant Professor in the Department of Aeronautical and Aviation Engineering, The Hong Kong Polytechnic University. Before that, he obtained his Ph.D. degree from University of Oxford. His research interests broadly lie in the design of intelligent perception solutions for autonomous systems, and the development of reliable 3D scene understanding algorithms on mobile robotics operating in the real world.
\end{IEEEbiography}

\begin{IEEEbiography}
[{\includegraphics[width=1in,height=1.25in,clip,keepaspectratio]{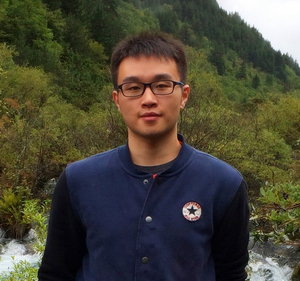}}]{Chris Xiaoxuan Lu}
is an Assistant Professor at School of Informatics, University of Edinburgh. Before that, he obtained his Ph.D degree at University of Oxford, and MEng degree at Nanyang Technology University,
Singapore. His research interest lies in Cyber Physical Systems, which use networked smart devices to sense and interact with the physical world.
\end{IEEEbiography}

\begin{IEEEbiography}
[{\includegraphics[width=1in,height=1.25in,clip,keepaspectratio]{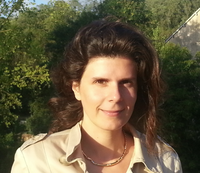}}]{Niki Trigoni}
is a Professor at the Department of Computer Science, University of Oxford. She is currently the director of the EPSRC Centre for Doctoral Training on Autonomous Intelligent
Machines and Systems, and leads the Cyber Physical Systems Group. Her research interests lie in intelligent and autonomous sensor systems with applications in positioning, healthcare, environmental monitoring and smart cities.
\end{IEEEbiography}

\begin{IEEEbiography}
[{\includegraphics[width=1in,height=1.25in,clip,keepaspectratio]{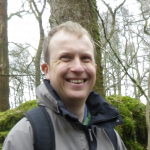}}]{Andrew Markham}
is a Professor at the Department of Computer Science, University of Oxford. He obtained his BSc (2004)
and PhD (2008) degrees from the University of Cape Town, South Africa. He is the Director of the MSc in Software Engineering. He works on resource-constrained systems, positioning systems,
in particular magneto-inductive positioning and machine intelligence.
\end{IEEEbiography}

\end{document}